\documentclass[12pt, titletoc, page]{thesis-umich}

\usepackage{booktabs}
\usepackage[english]{babel}

\RequirePackage[dvipsnames]{xcolor}  
\usepackage[dvipsnames]{xcolor}

\usepackage{graphicx}
\usepackage{subcaption}
\hypersetup{colorlinks=true, allcolors=blue}
\usepackage{hyperref}   
\usepackage{cleveref}   
\usepackage{url}            
\usepackage{booktabs}       
\usepackage{amsfonts}       
\usepackage{nicefrac}       
\usepackage{microtype}      
\usepackage{xcolor}         
\usepackage{microtype}
\usepackage{graphicx}
\usepackage{booktabs} 
\usepackage{tikz}
\usetikzlibrary{arrows}
\usetikzlibrary{decorations.markings}
\usetikzlibrary{calc,intersections}
\usepackage{bbding}
\newcommand{\xmark}{\ding{55}}
\usepackage{pifont}
\usepackage{amssymb}
\usepackage{sidecap}
\usepackage{svg}
\usepackage{caption}
\usepackage{textgreek}
\usepackage{amssymb}
\usepackage{multirow}
\usepackage{multicol}
\usepackage{amsmath}

\usepackage{arydshln}
\usepackage{natbib}
\usepackage{microtype}
\usepackage{graphicx}
\usepackage{booktabs} 
\usepackage{caption}
\usepackage{svg}
\usepackage{multirow}
\usepackage{verbatim}
\usepackage{algpseudocode}
\usepackage{algorithm}
\usepackage{algpseudocode}

\usepackage[dvipsnames]{xcolor}  

\usepackage{amsmath,amsfonts,bm}

\def\eqref#1{equation~\ref{#1}}

\def\1{\bm{1}}

\DeclareMathAlphabet{\mathsfit}{\encodingdefault}{\sfdefault}{m}{sl}
\SetMathAlphabet{\mathsfit}{bold}{\encodingdefault}{\sfdefault}{bx}{n}

\newcommand{\ourmethod}{HyperFields}

\title{Acquiring and Adapting Priors for Novel Tasks  via \texorpdfstring{\\}{ } Neural Meta-Architectures}
\date{}
\author{Sudarshan Babu}

\department{Computer Science}

\email{sudarshan@ttic.edu}

\committee{ %
	Professor Michael Maire, Co-Chair \\
	Professor Greg Shakhnarovich, Co-Chair \\
	Professor Aly Azeem Khan \\
	Professor David McAllester \\
	
} 

\cochair{Co-chair One \& Co-chair Two}

\acknowledgments{
I am deeply grateful to the communities at the Toyota Technological Institute at Chicago (TTIC) and the University of Chicago for fostering an environment that encouraged learning, growth, and exploration. The intellectually rich and collaborative culture across both institutions has been instrumental in shaping my academic journey.

I would like to thank the administrative team at TTIC—Amy Minick, Chrissy Coleman, Mary Marre, Adam Bohlander, Jessica Jacobson, Erica Cocom, and Derre Kobets—for their patience and support in helping me navigate countless forms and deadlines, despite my chronic tardiness. Your reminders and willingness to assist were invaluable. A special thanks to Mary for always letting me know when there was leftover food in the fridge waiting to be rescued—those timely updates were much appreciated. I’m also grateful to Adam for maintaining the TTIC cluster and for always responding—often within a minute—to my many cluster-related questions, no matter how silly they may have been.

The faculty at TTIC have been incredibly kind to me. A heartfelt thanks to Jinbo Xu for first taking me in as an intern without his support, I wouldn’t have had the opportunity to study at TTIC. During my time here, I benefited immensely from the courses I took, and I am especially grateful to Greg Shakhnarovich (for awarding me the first `A' I ever received in all of college), Mesrob Ohannessian, Karen Livescu, Kevin Gimpel, Avrim Blum, and Lek-Heng Lim for the various courses they taught me. Speaking of coursework, I owe much to my incoming batch of PhD students—Igor Vasiljevic, Pedro Savarese, Ankita Pasad, Lingyu Gao, Shengjie Lin, Omar Montasser, Davis Yosida, and Kevin Stangl—whom I regularly pestered for help and who always responded generously.

I’m especially thankful for the friendships I formed early on at TTIC, which were instrumental in helping me build the foundation I needed—both technically and personally—to navigate graduate school. Dan Hendrycks was generous with his time and advice, helping me make sense of coursework in those early months. Areej Khoury not only taught me a great deal of math, but also took it upon herself to “civilize” me—most notably in the fine arts of cooking and shopping. Mesrob Ohannessian, in addition to offering thoughtful guidance on academic life, introduced me to analysis, which gave me some much-needed mathematical grounding.

As I progressed in the program, the qualifying exams marked a significant milestone. I am grateful to my qual committee—Nati Srebro, Yury Makarychev, Avrim Blum, Julia Chuzhoy, and Matt Walter (yes, it was a large committee—that’s what happens when you take quals twice :P). Beyond the committee, I’d like to acknowledge Tushant Mittal, Rachit Nimavat, and my brother Arvindakshan Babu for patiently explaining several bespoke (then and now) math concepts over multiple days. Special thanks to Aravindan Vijayaraghavan, who generously met me on a weekend to help me with my qual paper (he was one of the authors). I'm also indebted to Greg Shakhnarovich, Michael Maire, and Igor Vasiljevic for running me through mock exams.

Learning from peers has been one of the most rewarding aspects of my time at TTIC. I want to thank all my group members—Pedro Savarese, Avery Zhou, Richard Liu, Xiao Zhang, Xin Yuan, Tri Huynh, Xiaodan Du, Jiahao Li, Haochen Wang, Igor Vasiljevic, Dale Decatur, David Yunis, and Phillip Lo—for countless technical discussions that shaped my research. They also provided much-needed social balance, often over shared meals. Especially Igor Vasiljevic for matching my energy in exchanging gossip—truly an underrated graduate school survival skill.

Beyond research, I’m grateful for the many friendships I formed with students outside those I worked with. In particular, I would like to thank Shashank, Rachit, David, Xiao, Roxi, Kayva, Anirudhh, Ahona, Arvind, Mrinal, Max, Nik, Nirmit, Pushkar, Igor, Naren, Anmol, Richard, Matt, Chris, Gowtham, Jenna, and Takuma for making graduate school a fun and memorable experience. Your friendship brought laughter and levity to many long days. During my time as an intern at TTIC, I appreciated the warm welcome and fun conversations with Arvind, Tianming, and Vijay that helped me feel at home right from the start of my journey in the U.S.

I can never thank my 4 AM friend group enough—Akilesh, Bruno, Kshitij, Mantas, Santosh, Shubham, and Tushant—for being a consistent and unwavering source of support, encouragement, and advice on everything under the sun. Your presence made the hardest moments manageable and the best ones even more memorable.

Despite skipping more classes than I should have as an undergrad, I did manage to make a few lifelong friends. I’m especially grateful to Ganesh for his steady support and for introducing me to TTI. I am thankful to Vikaasa for not only being a great friend but also my first co-author. While I have received support in various forms from friends, Sockalingam supported me like no other, by failing almost in each of the 25 courses I failed, there by giving me great support in retaking failed courses during my undergrad. I’d also like to acknowledge Shreya for her friendship and help with coursework.

I also want to thank some of the faculty from my undergraduate days who showed me kindness despite my academic antics. Susan Elias ma'am, Uma Habiba ma'am, Gomatheeswari Preethika ma'am, and Srividhya ma'am—all of you were generous with your encouragement. A very special thanks goes to Professor Venkateswaran, who introduced me to research during my undergrad days.

Moving on to friends from high school, I’m thankful to Bragadeesh, Deepak, Niranjan, Sajan, and Vishok for indulging my addiction to cricket and for making those years so much fun. A special thanks to Solaiappan for being an amazing friend since kindergarten.

I’m grateful to my cousin Niranjan Damera for his help in putting together my graduate school applications—his support and encouragement at that stage is something I truly appreciate. I’d also like to thank Prakash uncle, Jayanthi aunty, and my cousin Anand for thoughtfully shipping snacks and other goodies during the COVID lockdowns, when I was barely stepping out of my apartment. It was such a thoughtful gesture, and I truly appreciate it. I am thankful to my Aunt Jyothi and her family for their well wishes and for always wanting to see me everytime I go back to India.

Circling back to Faculty, I’d like to thank my committee members—Michael Maire, Greg Shakhnarovich, Aly Khan, and David McAllester—for their valuable time and feedback. A special thanks to Aly for continuing to work with me and helping me explore various interesting problems in computational biology.

I am thankful to Madhur Tulsiani for his kindness and generosity with his time. I still remember, as an intern, ambushing him with questions about the TTI admissions process. Despite the impromptu nature of the conversation, he patiently took the time to explain how the process worked and what faculty typically look for in applicants. It was not only an incredibly helpful discussion, but also a remarkably kind gesture toward someone he had just met. Since then, I’ve always enjoyed and looked forward to our interactions.

I greatly benefited from working with Rana Hanocka, who introduced me to a whole host of problems in 3D vision and graphics. I’m thankful for her experimental suggestions, which were crucial in showcasing our model's capabilities. Beyond the technical advice, I deeply appreciated her enthusiasm throughout the very long review process.

I am extremely fortunate to have Greg Shakhnarovich as my co-advisor. Over the years, his insights across several projects have consistently shaped my research. In particular, his suggestion to apply meta-learning tools from earlier projects to 3D generation has been central to my work—even now in my current position, I continue to explore that thread. Greg has also shown great personal kindness, especially during a transitional period when I was between advisors. Even when I wasn’t working with him directly, he invited me to join his group meetings—making me feel welcome and included, which at that time, when I was searching for an advisor, made all the difference to me.

Moving to my family members, I don’t think anyone would be as happy as my grandmother to know that I’ve completed my PhD. I’m always grateful for her unwavering desire to see me do well. My brother, Arvindakshan, has offered invaluable advice over the years, and I’m especially grateful for what stands out as the single best piece of advice I’ve ever received in my life—encouraging me to study machine learning. I’m also grateful to him for equipping me with many of the tools I needed to navigate graduate school. I’m thankful to my sister-in-law, Kabila, and my very cute nephews, Shivam and Vedanth, for bringing so much joy and happiness into my life.

I’m deeply thankful to my father for exemplifying a strong and tireless work ethic well into his 60s—something I hope to emulate. As for my mother, she somehow managed to spark my interest in science, despite the fact that I was a kid who refused to sit still with a book for more than five minutes—a true testament to her patience and persistence. I can never thank her enough for the countless sacrifices she has made for me. 

I want to close by thanking my advisor, Michael Maire—being mentored by him has been one of the best things to happen to me. I can never thank Michael enough for introducing me to a wide range of interesting problems and for giving me the freedom to explore them in my own way. The patience and kindness with which he supported me—especially in the face of my many shortcomings—is something I truly cherish and can only hope to emulate in my own life. For your unwavering support over all these years—thank you, Michael.
}

\abstract{
The ability to transfer knowledge from prior experiences to novel tasks stands as a pivotal capability of intelligent agents, including both humans and computational models. This principle forms the basis of transfer learning, where large pre-trained neural networks are fine-tuned to adapt to downstream tasks. Transfer learning has demonstrated tremendous success, both in terms of task adaptation speed and performance. However there are several domains where, due to lack of data, training such large pre-trained models or foundational models is not a possibility – computational chemistry, computational immunology, and medical imaging are examples. To address these challenges, our work focuses on designing architectures to enable efficient acquisition of priors when large amounts of data are unavailable. In particular, we demonstrate that we can use neural memory to enable adaptation on non-stationary distributions with only a few samples. Then we demonstrate that our hypernetwork designs (a network that generates another network)  can acquire more generalizable priors than standard networks when trained with Model Agnostic Meta-Learning (MAML). Subsequently, we apply hypernetworks to 3D scene generation, demonstrating that they can acquire priors efficiently on just a handful of training scenes, thereby leading to faster text-to-3D generation. We then extend our hypernetwork framework to perform 3D segmentation on novel scenes with limited data by efficiently transferring priors from earlier viewed scenes. Finally, we repurpose an existing molecular generative method as a pre-training framework that facilitates improved molecular property prediction, addressing critical challenges in computational immunology.
}

\begin{document}


\chapter{Introduction}

Acquiring and learning from prior experience is essential for any intelligent agent to adapt to novel tasks. This ability enables humans and other intelligent agents to leverage previously observed patterns for use in novel tasks, ultimately enabling them to solve these problems faster than when starting from scratch \cite{kyllonen2014role,singley1989transfer,johnson1987human}. Typically in neural networks this acquisition of priors is done by training on a larger dataset and the transfer is done by fine-tuning these pre-trained models to downstream tasks \cite{devlin2018bert,howard2018universal,jia2018transfer,zoph2020rethinking}. This process of fine-tuning a pre-trained model to solve a novel task is essentially transfer learning \cite{weiss2016survey,pan2009survey,yosinski2014transferable,DBLP:journals/corr/RussakovskyDSKSMHKKBBF14}.

Pre-trained models have been pivotal in driving advancements across multiple AI modalities, as evidenced by their transformative impact \cite{devlin2018bert,amodei2016deep,senior2020improved}. Specifically, the adoption of these models as foundational backbones has ushered in enhanced generalization, bolstered robustness, and expedited convergence for a plethora of downstream vision tasks. This includes, but is not limited to, semantic segmentation, instance segmentation, object detection, depth estimation, normal estimation, and pose estimation \cite{long2015fully,Mostajabi_2015_CVPR,he2017mask,girshick2015fast,bansal2017pixelnet,newell2016stacked}. Beyond their role in conventional vision tasks, pre-trained models have been instrumental in grounding visual data within other modalities, notably language \cite{lu2019vilbert,radford2021learning,tan2019lxmert}. This has catalyzed a series of innovative vision-language integrations, encompassing text-to-image synthesis, text-driven visual retrieval, visual question answering, and text-to-3D model generation \cite{ramesh2021zero,alayrac2022flamingo,grinols1996development}.

The success of large pre-trained models demonstrates that transfer learning is greatly improved by simply increasing the volume of the pre-training data. These pre-trained models often contain billions of parameters that are trained on significantly large volumes of data. The scale of both the model and data enables learning of highly informative feature representations, which is central to enabling significantly improved performance on a wide range of downstream tasks \cite{devlin2018bert, radford2019language, he2022masked}. In addition to scaling model and data, there are many works directed towards improving transfer learning. Several methods involve learning adaptor networks that transform either fully or partially the weights learned during the pre-training phase for the purpose of the downstream task \cite{rozantsev2018residual,guo2019spottune}. Self-supervised learning leverages large volumes of unlabeled data to learn rich visual representations, which can be effectively transferred to downstream tasks \cite{DBLP:journals/corr/abs-1911-05722,chen2020improved,caron2020unsupervised}. Several works propose augmenting pre-trained networks with task-specific parameters, and these methods vary only in the parameterization of the additional task-based parameters \cite{he2021effectiveness,perez2018film,mudrakarta2018k}. 

While transfer learning has made possible significant advances in several applications across modalities, there are several challenges in enabling learning systems to adapt faster for a novel task. In medical imaging, large pre-trained models require significant fine-tuning to adapt the network to X-rays or MRI data as the distribution shift is significant when compared to internet images \citep{wang2023medfmc}. In 3D vision applications, obtaining annotated 3D data is challenging. Therefore, long fine-tuning becomes vital for both generative and discriminative 3D tasks \cite{poole2022dreamfusion,kobayashi2022decomposing}.

We utilize meta-learning as a key mechanism for enabling rapid task adaptation. Gradient-based meta-learning methods optimize the initialization of gradient descent across a distribution of tasks, facilitating more efficient adaptation \cite{finn2017model, rusu2018meta,andrychowicz2016learning,ravi2016optimization,park2019meta,li2017meta}. These methods are indeed effective when dealing with only a handful of samples to learn novel tasks. Prior works parameterize gradient descent by learning an initialization that is amenable to rapid task adaptation \cite{finn2017model,rusu2018meta}. Additionally, methods learn pre-conditioning transforms on gradients that enable efficient transfer of priors or faster fine-tuning to downstream tasks \cite{ravi2016optimization,park2019meta,li2017meta}. 

Meta-learning methods generally employ a base learner, trained on a diverse set of tasks, which either provides the adaptation signal or directly parameterizes the task network responsible for novel downstream tasks. The base learner is typically pre-trained on a bunch of tasks for it to draw its experience from while generating the adaptation signal or network to solve the novel task. For example, consider developing a system capable of handling tasks drawn from a non-stationary distribution (a distribution that changes over time as the process evolves). To develop such a system, we train the meta-learner to solve many such non-stationary tasks. This enables the model to learn high-level strategies required to solve a novel task where input data is sampled from a non-stationary distribution.

In our work, we make architectural contributions to improve the state of meta-learning under challenging conditions to enable rapid adaptation in tasks wherein simply scaling pre-training data is either not helpful or not possible. We demonstrate that hypernetworks \cite{ha2016hypernetworks} -- networks trained to generate the weights of another network -- are efficient means of transferring priors across multiple tasks as they enable a form of weight sharing across tasks. We improve on the design of hypernetworks in order to facilitate rapid adaptation to novel tasks with limited data across both 2D (images) and 3D (object geometry) modalities. We demonstrate that, in the context of non-stationary distributions, neural memory can be leveraged to develop update strategies that mitigate the recency bias commonly observed in gradient descent optimization. We now outline our contributions in more detail.

In our work~\cite{babu2021online}, we propose a model augmented with memory, wherein memory is solely responsible for adaptation on (a) rapidly changing task objectives in an online stream of data, (b) a non-stationary online stream of data. We demonstrate that memory-augmented models outperform gradient-based fine-tuning in conditions where task objectives change rapidly. Our pre-training design strategy for the memory model involves providing inputs and labels to the memory model, enabling it to learn adaptation strategies with different inductive biases compared to gradient-based learning. The network’s learned adaptation strategies surpass gradient-based approaches on unseen online tasks with dynamic changes in task objectives. Additionally, we showcase the usefulness of this approach on various online tasks, including those with limited data or non-stationary distributions.

In our work~\cite{hypermaml}, we use hypernetworks to facilitate rapid task adaptation when only limited samples are available for learning new tasks (few-shot tasks). To this end, we train our hypernetwork across many tasks with only a few samples per class. We synthetically craft these few-shot tasks by segmenting larger datasets like ImageNet \cite{5206848}. The key idea is that, even if an individual few-shot task contains only a handful of samples, training the hypernetwork on multiple few-shot tasks prevents overfitting and enhances generalization to novel tasks, analogous to how training on diverse images improves generalization in image classification. Our findings reveal that once trained, the hypernetwork adeptly generates specialized networks to tackle novel few-shot tasks of interest. In fact, we show that our hypernetwork generates few-shot models that outperform the then state-of-the-art few-shot models when there is a significant distribution gap between the training tasks and the new few-shot task at hand. 

In our work~\cite{babu2023hyperfields}, we address the technical challenges of text-to-3D synthesis, an area gaining traction due to its potential applications in diverse fields such as gaming, virtual/augmented reality. The current paradigm relies on training individual Neural Radiance Fields (NeRFs) for each text prompt, resulting in significant optimization time and storage issues, as each unique scene requires a separate NeRF model. Our contribution, HyperFields, detailed in \cite{babu2023hyperfields}, overcomes this overhead by introducing a hypernetwork capable of generating NeRF weights for any given text prompt, leveraging learned information from previous scenes to facilitate rapid and efficient novel scene generation. This approach not only accelerates the creation of 3D scenes from hours to minutes but also alleviates the need to store individual NeRFs for each scene, thereby offering a scalable and transferable solution to the problem of text-to-3D synthesis.

In addition to our contributions in data-scarce 2D and 3D modalities, we extend our contributions to the molecular domain, where learning robust molecular representations under data scarcity is essential. Specifically, we use a pre-trained MiDi diffusion model to extract geometry-aware molecular features from molecular conformers \cite{vignac2023midi}. These features, combined with sequence models, enable effective predictions for downstream tasks such as drug-target interaction modeling. Our results show that diffusion-derived features complement text-based molecular representations, enhancing performance and capturing a diverse set of candidate molecules for experimental validation.

Our work demonstrates the efficacy of meta-learning architectures in efficiently transferring priors, especially when the target distribution differs significantly from the source distribution. The application of these meta-learning architectures to various problems highlights the flexibility and task-agnostic nature of our architectural contributions. The ability to effectively transfer priors, coupled with the flexibility of meta-learning architectures, makes a strong case for their adoption in the training of foundational models that are used ubiquitously across various machine learning applications.

\chapter{Online Adaptation via Distributed Neural Memory}

\section{Introduction}

In this chapter, we investigate the meta-learning capabilities of memory-based architectures and examine the organization of neural memory for efficient meta-learning. Motivating this focus is the generality and flexibility of memory-based approaches. By leveraging memory for adaptation, meta-learning can be formulated as a standard learning problem using a straightforward loss function that treats entire episodes as examples and applies standard optimization techniques. The actual burden of task adaptation becomes an implicit responsibility of the memory subsystem: the network must learn to use its persistent memory in a manner that facilitates task adaptation. This contrasts with explicit adaptation mechanisms such as stored prototypes.

In this implicit adaptation setting, memory architecture plays a crucial role in determining what kind of adaptation can be learned. We experimentally evaluate the effectiveness of alternative neural memory architectures for meta-learning and observe particular advantages to distributing memory throughout a network. More specifically, we view the generic LSTM equations, $Wx + W^{'}h_{-1}$, as adaptation induced by hidden states in activation space (see Figure~\ref{fig:adaptation}). By distributing LSTM memory cells across the depth of the network, each layer is tasked with generating hidden states that are useful for adaptation. Such a memory organization is compatible with many standard networks, including CNNs, and can be achieved by merely swapping LSTM memory cells in place of existing filters.

Our straightforward approach also contrasts with several existing memory-based meta-learning methods used in both generative and classification tasks \cite{santoro2016one,bartunov2019meta,wu2018kanerva,wu2018learning,pritzel2017neural}. These methods view memory as a means to store and retrieve useful inductive biases for task adaptation and hence focus on designing better read and write protocols. Unlike traditional methods that employ a feature extractor feeding into a memory network for adaptation, our architecture integrates memory directly into the network layers, removing this distinction.

\begin{figure}[b]

\captionsetup{font=small,labelfont=small}

{\centerline{

  \resizebox{7.3cm}{!}{
    \begin{tikzpicture}[
      scale=0.4,
      level/.style={thick},
      virtual/.style={thick,densely dashed},
      trans/.style={thick,<->,shorten >=2pt,shorten <=2pt,>=stealth},
      classical/.style={thin,double,<->,shorten >=4pt,shorten <=4pt,>=stealth}
    ]
    \node at (3.85,5.8) (nodeA) {\tiny $F^{*}$};
    \node  at (0.1,6.2) (nodeB) {\tiny F};
    \node at (4.8,5.8) (nodeC) {  \tiny{ \color{red}{$u_{1}$ }}};
    \node at (6.4,6.9) (nodeD) {  \tiny{ \color{red}{$u_{n}$}}};
    \node at (4,4.6) (nodeC) {  \tiny{ \color{blue}{$z_{1}$}}};
    \node at (7.7,4.5) (nodeD) {  \tiny{ \color{blue}{$z_{n}$}}};
    \node at (-1,5) (nodeC) {  \tiny{ {$u_{i} = F^{*}(x^{(1)}_{i}, \color{red} {h^{(1)}_{i}}\color{black}{)} $ }}};
    \node at (-1,4) (nodeC) {  \tiny{ {$z_{i} = F^{*}(x^{(2)}_{i}, \color{blue}{h^{(2)}_{i}}\color{black}{)} $ }}};
    \node [rotate=355 ] at (2,6.2) (node z) {\tiny Training};
    \draw[->,black,thick] (0.2,6) to (1,5.92)  ;
    \draw[->,black,thick] (1,5.92) to (2,5.80);
    \draw[->,black,thick] (2,5.80) to (3,5.63);
    \draw[->,black,thick] (3,5.63) to (4,5.40);
    \draw[->,red,thick] (4,5.40) to (4.3, 6.2); 
    \draw[->,red,thick] (4.3, 6.2) to (4.5, 6.5);
    \draw[->,red,thick] (4.5,6.5) to (4.8, 6.8);
    \draw[->,red,thick] (4.8,6.8) to (5.2, 7);
    \draw[->,red,thick] (5.2,7) to (5.5, 7.1);
    \draw[->,red,thick] (5.5,7.1) to (6.2, 7.2);
    \draw[->,blue,thick] (4,5.40) to (4.5, 4.7); 
    \draw[->,blue,thick] (4.5,4.7) to (5.3, 4.40); 
    \draw[->,blue,thick] (5.3, 4.40) to (5.9, 4.40);
    \draw[->,blue,thick] (5.9, 4.40) to (6.7, 4.40);
    \draw[->,blue,thick] (6.7, 4.40) to (7.3, 4.45);
    \end{tikzpicture}
  }
}
}
{\caption{%
\textbf{Adaptation in activation space.}
A trained memory-based model $F^{*}$ adapting to two different tasks (red and blue path) using the corresponding persistent states $h_{i}^{1}$ and $h_{i}^{2}$ at the $i^{th}$ time step of both tasks. $x_{i}^{(1)}$ and  $x_{i}^{(2)}$ are samples of task 1 (red) and 2 (blue) at time step $i$.%
}}%
\addtocounter{figure}{-1}  
\refstepcounter{figure}
\label{fig:adaptation}

\end{figure}
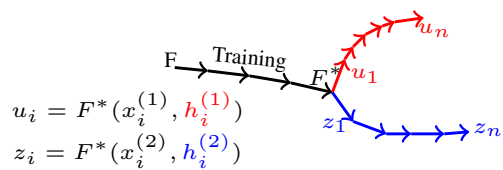

We test the efficacy of network architectures with distributed memory cells on online few-shot and continual learning tasks as in \cite{santoro2016one,ren2020wandering} and \cite{javed2019meta}. The online setting is challenging for two reasons: 1) It is empirically observed that networks are not well suited for training/adaptation with a batch size of one \cite{bengio2012practical}; 2) In this setting, the model has to adapt to one image at a time step, requiring it to handle a prolonged adaptation phase. Thus, these tasks provide a rigorous testbed for evaluating how effectively the network's hidden states contribute to adaptation.

We empirically observe that our method outperforms strong gradient-based and prototypical baselines, delineating the efficacy of the local adaptation rule learned by each layer. Particularly important is the distributed nature of our memory, which allows every network layer to adapt when provided with label information; restricting adaptability to only later network layers significantly hampers performance. These results suggest that co-design of memory architecture and meta-training strategies should be a primary consideration in the ongoing development of memory-based meta-learning. We further test our model in a harder online few-shot learning scenario, wherein the corresponding label to a sample arrives after a long delay \cite{mesterharm2005line}. Without requiring any modifications to the model, our method adapts seamlessly, whereas other adaptation strategies struggle under these conditions. These results highlight promising directions for advancing and simplifying meta-learning by relying upon distributed memory for adaptation.

\section{Related Work}
Early work on meta-learning introduces several important concepts. Schmidhuber \cite{schmidhuber1992learning} proposes the use of task-specific weights, referred to as fast weights, and weights that adapt across tasks, known as slow weights. Another approach, presented in \cite{bengio1992optimization}, updates the network via a learning rule parameterized by another neural network. In a lifelong learning scenario, \cite{thrun1998lifelong} demonstrates how an algorithm can accrue information from past experiences to adapt effectively to the task at hand. Hochreiter et al.~\cite{hochreiter2001learning} train a memory network to learn its own adaptation rule through its recurrent states. These high-level concepts are reflected in more recent methods. We group current meta-learning methods based on the nature of their adaptation strategies and discuss them below.

\textbf{Gradient-based Adaptation Methods.}
Methods that adapt via gradients constitute a prominent class of meta-learning algorithms, as outlined in \cite{hospedales2020meta}. Model-agnostic meta-learning (MAML)  exemplifies this approach by learning an initialization that can be efficiently adapted via gradient descent for new tasks \cite{finn2017model}. Another approach, presented in \cite{finn2019online}, focuses on learning a network capable of using experience from previously seen tasks for current task adaptation. This is achieved by utilizing a network pre-trained with MAML on samples from prior tasks. Online adaptation under non-stationary distributions is explored in \cite{nagabandi2018deep,caccia2020online}, where either a mixture model is used, or a MAML pre-trained network is spawned when the input distribution changes. Additionally, \cite{javed2019meta,beaulieu2020learning} employ a bi-level optimization routine similar to MAML, with the outer loop loss addressing catastrophic forgetting. This approach enables the learning of representations that are robust to forgetting and accelerates future learning during online updates.

\textbf{Memory and Gradient-based Adaptation.}
The works by \cite{andrychowicz2016learning,ravi2016optimization} demonstrate learning an update rule for network weights by transforming gradients through an LSTM, resulting in performance that surpasses human-designed and fixed SGD update rules. Another work, described in \cite{munkhdalai2017meta}, involves learning a transform that maps gradients to fast (task-specific) weights, which are stored and retrieved via attention during evaluation. In this method, the slow weights (across-task weights) are updated at the end of each task.

\textbf{Prototypical Methods.}
These methods learn an encoder which projects training data to a metric space, and obtain class-wise prototypes via averaging representations within the same class. Following this, test data is mapped to the same metric space, wherein classification is achieved via a simple rule (\emph{e.g.,}~nearest neighbor prototype based on either Euclidean distance or cosine similarity) \cite{snell2017prototypical, oreshkin2018tadam,pan2019transferrable}. These methods are naturally amenable for online learning as class-wise prototypes can be updated in an online manner as shown by \cite{ren2020wandering}.

\textbf{Memory-based Adaptation.}
Efficient read and write protocols for a Neural Turing Machine \cite{NTM} have been designed by \cite{santoro2016one} for the purposes of online few-shot learning. Sparse read and write operations, which are scalable in both time and space, are introduced by \cite{rae2016scaling}. Ramalho and Garnelo~\cite{ramalho2019adaptive} use logits generated by the model to decide if a certain sample should be written to neural memory. Mishra et al.~\cite{mishra2017simple} employ an attention-based mechanism for adaptation, utilizing a CNN to generate features for the attention mechanism. However, their model requires storing samples across all time steps explicitly, thereby violating the online learning assumption of accessing each sample only once.

These methods mainly focus on designing better memory modules, either through more recent attention mechanisms or by improving read and write rules for neural memory. Typically, these approaches use a CNN that is not adapted to the current task. Our approach differs by studying the efficient organization of memory for both online few-shot learning and meta-learning more generally, demonstrating that our distributed memory organization enables the entire network to adapt effectively when provided with relevant feedback.

An interesting form of weight sharing is introduced by \cite{kirsch2020meta}, where LSTM cells with tied weights are distributed across the width and depth of the network, with each position maintaining its own hidden state. These authors also include backward connections from later layers to earlier layers, allowing the network to implement its own learning algorithm or clone a human-designed one, such as backpropagation. Both their model and ours implement an adaptation strategy purely using recurrent states. However, the key difference lies in the nature of the adaptation strategy: their backward connections propagate error from the last layer to earlier layers, similar to conventional learning algorithms, while our architecture presents the feedback signal as another input, propagating it from the first layer to the last.

Memory-based meta-learning approaches have also been applied beyond classification settings. For instance, \cite{guez2019investigation} employ memory-based meta-learning to perform adaptation in reinforcement learning tasks, demonstrating the generality of using memory as a means for adaptation.

\textbf{Few-shot Semantic Segmentation.}
Few-shot segmentation methods commonly rely on using prototypes \cite{shaban2017one,rakelly2018conditional}, although recent approaches have included gradient-based methods analogous to MAML \cite{banerjee2020meta}. Methods employing neural memory typically integrate it in the final network stages to fuse features of different formats for efficient segmentation. For instance, \cite{li2018referring} use ConvLSTMs \cite{shi2015convolutional} to fuse features from different stages of the network, while \cite{valipour2017recurrent} apply ConvLSTMs to fuse spatio-temporal features during video segmentation. Hu et al.~\cite{hu2019attention} utilize a ConvLSTM to fuse features of the query with features from the support set, and \cite{azad2021texture} implement a bidirectional ConvLSTM to fuse segmentation derived from multiple scale-space representations.

Our approach differs from these works in the organization, use, and information provided to the memory module: (1) Memory is distributed across the network as the sole driver of adaptation; (2) Label information is provided to assist with adaptation.

\textbf{Meta-learning Benchmarks.} The benchmarks presented by \cite{caccia2020online} measure the ability of a model to adapt to a new task using the inductive biases it has acquired from solving previously seen tasks. More specifically, the benchmark introduces an online non-stationary stream of tasks, and the model's ability to adapt to a new task at each time step is evaluated. It is important to note that this benchmark does not assess the model's ability to remember earlier tasks; the focus is solely on the model's capacity to adapt well to a newly presented task.

Benchmarks for continual few-shot learning are introduced by \cite{antoniou2020defining}. In this setup, the network is presented with a number of few-shot tasks, one after the other, and is then expected to generalize even to the previously seen tasks. This setup is both challenging and interesting because the network must demonstrate robustness to catastrophic forgetting while learning from limited data. However, our interest lies in evaluating the online adaptation ability of models, whereas \cite{antoniou2020defining} feed data in a batch setting. We follow the experimental setup described by \cite{javed2019meta}, where the model is required to remember inductive biases acquired over a longer time frame compared to the experimental setup used by \cite{antoniou2020defining}.

\section{Method}
\subsection{Problem: Online Few-shot Learning}
\label{sec:task}
This setting combines facets of online and few-shot learning: the model is expected to make predictions on a stream of input samples, while it sees only a few samples per class in the given input stream.  In particular, we use a task protocol similar to \cite{santoro2016one}.  At time step $i$, an image $x_{i}$ is presented to the model and it makes a prediction for $x_{i}$.  In the following time step, the correct label $y_{i}$ is revealed to the model.  The model's performance depends on the correctness of its prediction at each time step. The following ordered set constitutes a task: $\mathcal{T} = \big((x_{1}, null), (x_{2},y_{1}), \cdots (x_{t}, y_{t  -1}) \big)$.  Here $null$ indicates that no label is passed at the first time step, and $t$ is the total number of time steps (length) of the task. For a k-way N-shot task $t = k \times N$.  The entire duration of the task is considered as the adaptation phase, as with every time step the model gets a new sample and must adapt on it to improve its understanding of class concepts.

\subsection{Memory as Adaptation in Activation Space}
\label{mem_adapt}
Consider modulating the output of a network $F$ for input $x$ with a persistent state $h$: $u = F(x,h)$.  Now, if adding $h$ aids in realizing a better representation $u$ than otherwise ($F(x)$), we could view this as adaptation in activation space.  In Figure~\ref{fig:adaptation}, model $F^{*}$ adapts to tasks using its persistent states $h$. Specifically let us consider the generic LSTM equations $Wx + W^{'}h_{-1}$, we could view $Wx$ as the original response and  $W^{'}h_{-1}$ as modulation by a persistent state (memory) in the activation space.  So, for the online learning task at hand, we seek to train a LSTM which learns to generate hidden state $h_{i}$ at each time step $i$, such that it could enable better adaptation in ensuing time steps.  We note that adaptation in activation space has been discussed in earlier works.  We use this perspective to organize memory better and to enable effective layer-wise adaptation across the network.

\begin{figure*}
\centerline{
  \resizebox{13cm}{!}{
    \begin{tikzpicture}[
      scale=1,
      level/.style={thick},
      virtual/.style={thick,densely dashed},
      trans/.style={thick,<->,shorten >=2pt,shorten <=2pt,>=stealth},
      classical/.style={thin,double,<->,shorten >=4pt,shorten <=4pt,>=stealth}
    ]
    \draw[->,line width =1mm] (0,0)--(-0,16) node[midway, above, sloped, scale =3]{$Time$};
    \foreach \x [count=\xi] in {0,...,3}
        \draw [draw=black,fill=lightgray] (8+\x*8,1) rectangle (10+\x*8,6) node[pos=.5, rotate=90, scale =2.0] {ConvLSTM-$\xi$};
    \foreach \x [count=\xi]  in {0,...,3}   
        \draw [draw=black,fill=cyan] (-1+8+\x*8,8) rectangle (1+10+\x*8,10) node[pos=.5, rotate=0, scale =2.3] {($c^{(\xi)}_{i}, h^{(\xi)}_{i}$)};
    \foreach \x in {0,...,4}   
        \draw [draw,-triangle 90,fill=blue, line width =1mm]  (9+\x*8,10) -- (9+\x*8,12);
    \foreach \x in {0,...,4}   
        \draw [draw,-triangle 90,fill=blue, line width =1mm] (9+\x*8,6) -- (9+\x*8,8) ;
    \foreach \x [count=\xi]  in {0,...,3}
        \draw [draw=black,fill=lightgray] (8+\x*8,12) rectangle (10+\x*8,17) node[pos=.5, rotate=90, scale =2.0] {ConvLSTM-$\xi$};
    \foreach \x in {0,...,4}
        \draw [draw,-triangle 90,fill=blue, line width =1mm] (10+\x*8,14.5) -- (12+\x*8,14.5);
    \foreach \x in {0,...,4}
        \draw [draw,-triangle 90,fill=blue, line width =1mm] (14+\x*8,14.5) -- (16+\x*8,14.5) ;
    \foreach \x [count=\xi]  in {0,...,3}   
        \draw [draw=black,fill=cyan] (12+\x*8,13.5) rectangle (14+\x*8,15.5) node[pos=.5 ,rotate=0, scale =2.8] {$h^{(\xi)}_{i+1}$};
    \foreach \x [count=\xi]  in {0,...,3}   
        \draw  [draw=black,fill=cyan] (12+\x*8,2.5) rectangle (14+\x*8,4.5) node[pos=.5, rotate=0, scale =2.8] {$h^{(\xi)}_{i}$};
    \foreach \x in {0,...,4}
        \draw [draw,-triangle 90,  line width =1mm,fill=blue] (10+\x*8,3.5) -- (12+\x*8,3.5);
    \foreach \x in {0,...,4}
        \draw[ -triangle 90, line width =1mm ] (14+\x*8,3.5) -- (16+\x*8,3.5) ; 
    \draw [draw=black,fill=lightgray] (34-2+4+4,1) rectangle (36-2+4+4,6) node[pos=.5, rotate=90, scale =2.0] {LSTM};    
    \draw [draw=black,fill=lightgray] (34-2+4+4,12) rectangle (36-2+4+4,17) node[pos=.5, rotate=90, scale =2.0] {LSTM};  
    \draw [draw = black, fill = green] (3,2.5) rectangle (5,4.5) node[pos=.5, rotate = 0,scale =3] {$x_{i}$} ;
    \draw [draw = black, fill = orange] (3,-1.5) rectangle (5,.5) node[pos=.5, rotate = 0,scale =3] {$y_{i-1}$} ;
    \draw [draw,red, line width =1mm] (5,-1.2) -- (41.,-1.2) ;
    \draw [draw,violet, line width =1mm] (5,-.7) -- (33,-.7) ;
    \draw [draw,pink, line width =1mm] (5,-.3) -- (8.3,-.3) ;
    \draw [draw,-triangle 90,red, line width =1mm] (41,-1.2) -- (41.,-1.2+2+.2) ;
    \draw [draw,-triangle 90,violet, line width =1mm] (33,-.7) -- (33,-1.2+2+.2) ;
    \draw [draw,-triangle 90,violet, line width =1mm] (25,-.7) -- (25,-1.2+2+.2) ;
    \draw [draw,-triangle 90,violet, line width =1mm] (17,-.7) -- (17,-1.2+2+.2) ;
    \draw [draw,-triangle 90,violet, line width =1mm] (9.5,-.7) -- (9.5,-1.2+2+.2) ;
    \draw [draw,-triangle 90,pink, line width =1mm] (8.3,-.3) -- (8.3,-1.2+2+.2) ;
    \draw [draw,-triangle 90,fill=blue, line width =1mm] (5,3.5) -- (8,3.5) ;
    \draw [draw = black, fill = green] (3,13.5) rectangle (5,15.5) node[pos=.5, rotate = 0,scale =3] {$x_{i+1}$} ;
    \draw [draw,-triangle 90,fill=blue, line width =1mm] (5,14.5) -- (8,14.5) ;
    \draw [draw=black,fill=cyan] (8+32-1,8) rectangle (10+32+1,10) node[pos=.5, rotate=0, scale =3] {$(c_{i},h_{i})$};
    \draw [draw=black,fill=cyan](12+32,2.5) rectangle (14+32,4.5) node[pos=.5, rotate=0, scale =3] {$h_{i}$};
    \draw [draw=black,fill=cyan] (12+32,13.5) rectangle (14+32,15.5) node[pos=.5, rotate=0, scale =3] {$h_{i+1}$};
    \draw [draw=black,fill=lightgray] (34-2+4+4+6+2,1) rectangle (36-2+4+4+6+2,6) node[pos=.5, rotate=90, scale =2.0] {Classifier};
    \draw [draw=black,fill=lightgray] (34-2+4+4+6+2,12) rectangle (36-2+4+4+6+2,17) node[pos=.5, rotate=90, scale =2.0] {Classifier};
    \end{tikzpicture}
  }
}
\caption{%
\textbf{Methodology.}
An example distributed memory architecture consists of four layers of convolutional LSTMs (gray), followed by an LSTM (gray), and a classifier (gray).  At the $i_{th}$ time step, the sample $x_{i}$ (green) and the previous sample's label $y_{-1}$ are presented to network.  Three different label injection modes are shown.
\textcolor{pink}{Pink}: label is fed to the first CL layer only;
\textcolor{violet}{Violet}: label is fed to each CL layer;
\textcolor{red}{Red}: label is fed to the LSTM layer.
Label information is provided at every time step; for sake of clarity we avoid showing it at time $i+1$ here.  $h^{(j)}_{i}$, $c^{(j)}_{i}$ are the hidden and cell state of the $j_{th}$ layer at the $i_{th}$ time step.  We use cyan, gray, orange, and green to denote persistent states, network parameters, label information, and input sample, respectively.  Best viewed in color.}
\label{fig:method}
\end{figure*}
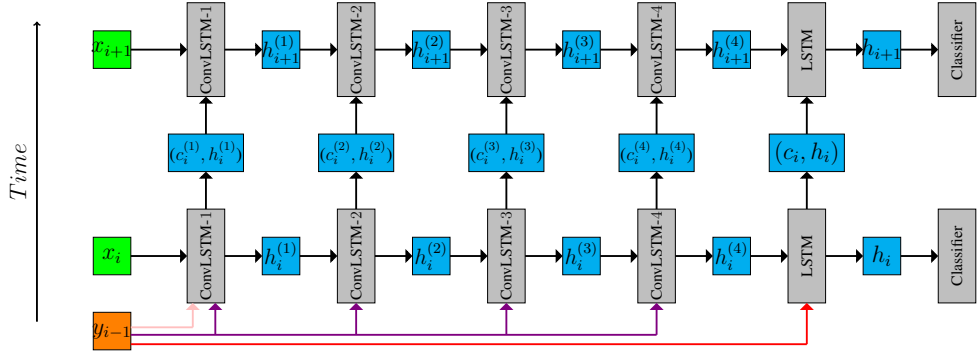

\subsection{Model}
\label{sec:arch1}
\textbf{Architecture.}
We distribute memory across the layers of the network, in order to enable the layers to learn local layer-wise adaptation rules.  In particular, we use a model in which each layer of the feature extractor is a ConvLSTM (CL) \cite{shi2015convolutional} followed by a LSTM \cite{hochreiter1997long} and a classifier, as shown in Figure~\ref{fig:method}.

Similar to the LSTM, each ConvLSTM (CL) layer consists of its own input, forget, and output gates.  The key difference is that convolution operations (denoted by $*$) replace matrix-vector multiplication. In this setup, we view the addition by $W_{hi}*h_{t-1}$ as adaptation in the $i_{th}$ time step within the input gate.  The same view could be extended to other gates as well.  The cell and hidden state generation are likewise similar to LSTM, but use convolution operations:
  \begin{align}
    i_{t} &= \sigma(W_{ii}*x_{t} + W_{hi}*h_{t-1}) \label{eqn:1} \\
    f_{t} &= \sigma(W_{if}*x_{t} + W_{hf}*h_{t-1}) \label{eqn:2} \\
    o_{t} &= \sigma(W_{io}*x_{t} + W_{ho}*h_{t-1}) \label{eqn:3} \\
    c_{t} &= f_{t} \odot c_{t-1} + i_{t} \odot \tanh(W_{ig}*x_{t} + W_{hg}*h_{t-1}) \label{eqn:4} \\
    h_{t} &= o_{t} \odot \tanh(c_{t}) \label{eqn:5}
\end{align}
In initial experiments, we observe that for tasks with 50 time steps these models did not train well. We hypothesize that this could be due to the same network being repeated 50 times, thereby inducing an effectively very deep network. We resolve this issue by adding skip connections between the second layer and the fourth layer (omitted in Figure~\ref{fig:method}). Further discussion on this is in Appendix ~\ref{app:opt}.

\textbf{Label Encoding.}
As label information is essential for learning an adaptation rule, we inject labels offset by one time step to the ConvLSTM feature extractor and the LSTM.  This provides the opportunity for each layer to learn an adaptation rule.  For a k-way classification problem involving images of spatial resolution $s$, we feed the label information as a $k \times s^{2}$ matrix with all ones in the $c^{th}$ row if $c$ is indeed the true label.  We reshape this matrix as a $k \times s \times s$ tensor and concatenate it along the channel dimension of image at the next time step.  To the LSTM layer, we feed the label in its one-hot form by concatenating it with the flattened activations from the previous layer.

\subsection{Training and Evaluation}
Following \cite{santoro2016one}, we perform episodic training by exposing the model to a variety of tasks from the training distribution $\mathcal{P}(\mathcal{T}_{train})$.  For a given task, the model incurs a loss $\mathcal{L}_{i}$ at every time step of the task; we sum these losses and backpropagate through the sum at the end of the task.    We evaluate the model using a partition of the dataset that is class-wise disjoint from the training partition. The model makes a prediction at every time step and adapts to the sequence by using its own hidden states, thereby not requiring any gradient information for adaptation.  

\section{Results}
\subsection{Online Few-Shot Learning}
We use CIFAR-FS \cite{bertinetto2018meta} and Omniglot \cite{blake} datasets for our few-shot learning tasks; see Appendix ~\ref{app:of} for details. We adopt the following methods to serve as baselines for comparison.

\subsection{Baselines}
Santoro et al.~\cite{santoro2016one} use a LSTM and a NTM~\cite{NTM} with read and write protocols for the task of online few-shot learning. Both aim to meta-learn tasks by employing a neural memory. 

\textbf{Adaptive Posterior Learning (APL).}
Ramalho and Garnelo ~\cite{ramalho2019adaptive} propose a memory-augmented model that stores data point embeddings based on a measure of \emph{surprise}, which is computed by the loss incurred by each sample. During inference, they retrieve a fixed number of nearest-neighbor data embeddings, which are then fed to a classifier alongside the current sample.

\textbf{Online Prototypical Networks (OPN).}
Ren et al.~\cite{ren2020wandering} extend prototypical networks to the online case, where they sequentially update the current class-wise prototypes using weighted averaging. 

\textbf{Contextual Prototypical Memory (CPM).}
Ren et al.~\cite{ren2020wandering} improve on OPN by learning a representation space that is conditioned on the current task. Furthermore, weights used to update prototypes are determined by a newly-introduced gating mechanism.

Table~\ref{tab:onc} shows that our model outperforms the baselines in most settings. These results suggest that the adaptation rules emergent from our design are more efficient than adaptation via prototypes, and adaptation via other memory-based architectures. In the CIFAR-FS experiments, the prototypical methods outperform our method only in the 1-shot scenario.  As the 5-shot and 8-shot scenarios have a longer fine-tuning or adaptation phase, this shows that our method is more adept at handling tasks with longer adaptation phases.  One reason could be that the stored prototypes which form the persistent state of OPN and CPM are more rigid than the persistent state of our method.  The rigidity stems from the predetermined representation size of each prototype, which thereby prevents allocation of representation size depending upon classification difficulty.  In our architecture, the network has the freedom to allocate representation size for each class as it deems fit.  Consequently, this may help the network learn more efficient adaptation strategies that improve with time. 

\begin{table*}[h]
\centering
\scalebox{1}{
\begin{tabular}[t]{ccccccc}
\toprule
\multicolumn{1}{c}{\multirow{2}{*}{Model}}&  \multicolumn{3}{c}{Omniglot}  & \multicolumn{3}{c}{CIFAR-FS}  \\
\multicolumn{1}{c}{} &  \multicolumn{1}{c}{1-shot} & \multicolumn{1}{c}{5-shot}& \multicolumn{1}{c}{8-shot}& \multicolumn{1}{c}{1-shot} &  \multicolumn{1}{c}{5-shot}& \multicolumn{1}{c}{8-shot} \\
\midrule
LSTM$^\dagger$  & 85.3 (0.2) & 94.4 (0.1) & 95.8 (0.6) & - & - & - \\ 
NTM$^\dagger$   & 88.7 (0.5) & 96.8 (0.1) & 97.3 (0.1) & - & - & -\\
APL  & 89.1 (0.0) & 94.9 (0.0) & 95.7 (0.1) & 37.6 (0.6) & 45.8 (0.4) & 46.9 (0.7) \\ 
OPN  & 91.2 (1.1)  & 94.6 (1.1)    & 95.8 (0.5) & 49.9 (0.2) & 54.9 (0.4) & 56.8 (0.1)   \\
CPM  & 94.5 (0.1) & 97.0 (0.1) & 97.4 (0.4) & \textbf{50.2} (0.2) & 55.8 (0.4) & 57.8 (0.4) \\
CL+LSTM & \textbf{96.8} (0.5) & \textbf{99.4} (0.2) & \textbf{99.7} (0.2) & 47.6 (1.0) & \textbf{56.7} (1.6) & \textbf{61.0} (1.3) \\
\end{tabular}}
\caption{
\textbf{Omniglot and CIFAR-FS results for 5-way online few-shot learning}.
Shown are average and standard deviation across 3 runs. Methods with hand-designed memory mechanisms, like CPM and APL, benefit less from increased number of samples. Distributed memory in CL+LSTM comfortably outperforms other adaptation methods.  $^\dagger$ These methods fail to train on CIFAR-FS.}
\label{tab:onc}
\end{table*}
\subsection{Online Continual Learning}
\label{subsec:ocl}
We address the problem of continual learning in the online setting. In this setup, the model sees a stream of samples from a non-stationary task distribution, and the model is expected to generalize well even while encountering samples from a previously seen task distribution. Concretely, for a single continual learning task we construct $n$ subtasks from an underlying dataset and first present to the model samples from the first subtask, then the second subtask, so on until the $n^{th}$ subtask in that order. Once the model is trained on all $n$ subtasks sequentially, it is expected to classify images from any of the subtasks, thereby demonstrating robustness catastrophic forgetting \cite{kirkpatrick2017overcoming}. 

\begin{figure}
\centering
\begin{minipage}{.40\textwidth}
  \centering
  \includegraphics[width=.8\linewidth, height = 3cm]{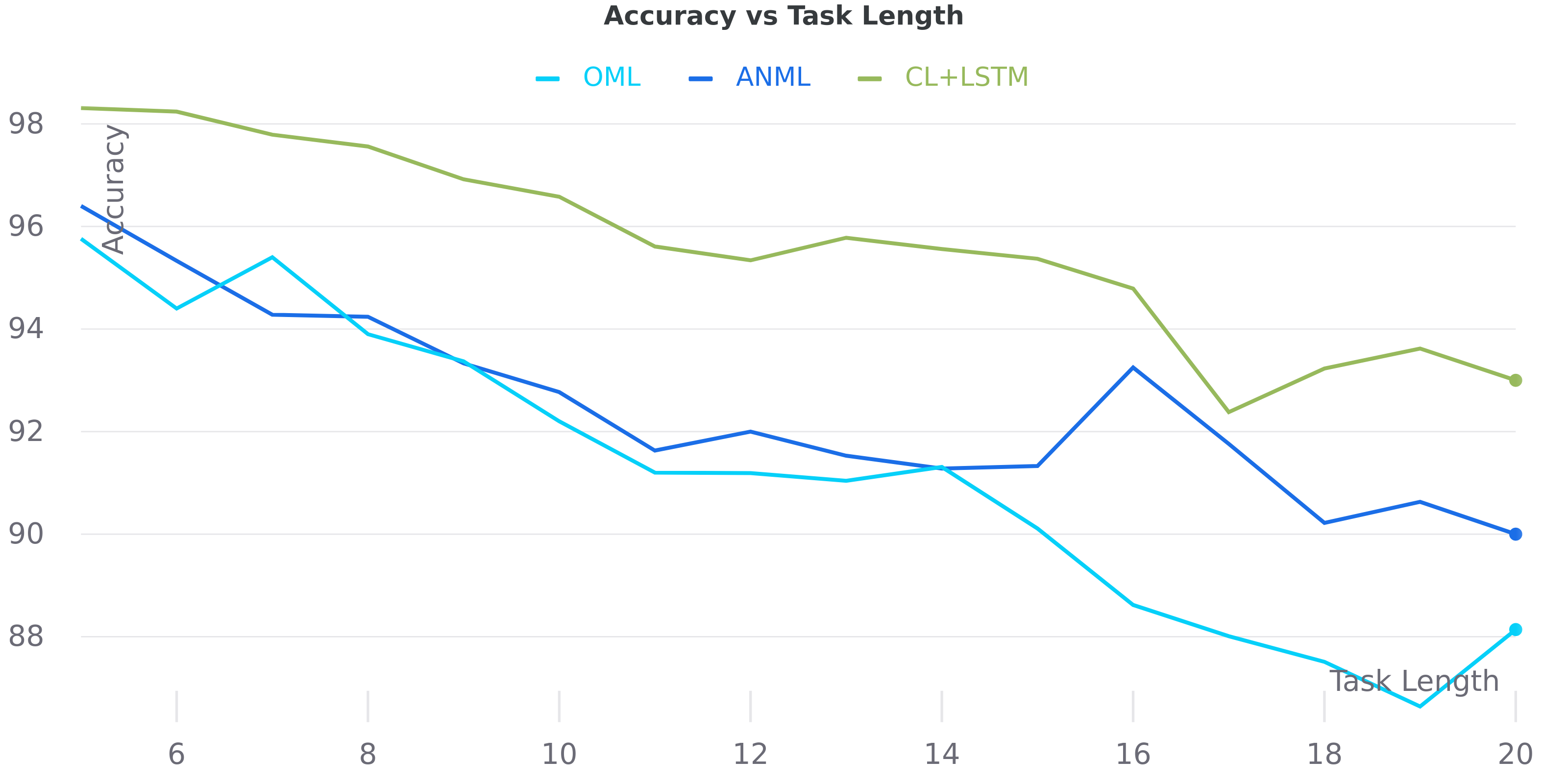}
\captionof{figure}{%
\textbf{Online few-shot continual learning.}
Accuracy vs task length on Omniglot. CL+LSTM model outperforms the baselines across all task lengths, strongly suggesting that the CL+LSTM model is adept at storing inductive biases required to solve the subtasks within a given continual learning task.}
  \label{fig:cont_res}
\end{minipage}%
\hfill
\begin{minipage}{.58\textwidth}
  \centering
  \includegraphics[width=.7\linewidth, height = 3cm]{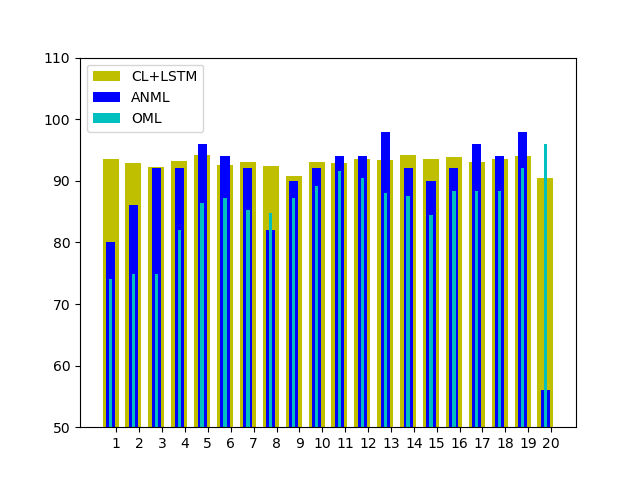}
  \captionof{figure}{%
\textbf{Accuracy of subtasks within a 20 task continual learning task.}
Typically in earlier tasks (first 10), CL+LSTM seems to have a higher accuracy than ANML and OML, suggesting that it is more immune to forgetting. ANML performs poorly in the $20^{th}$ task, about 56 \% we impute that to momentum issues in training. Note: Excluding the $20^{th}$ task, the average ANML performance in the first 19 tasks is $91.78$,  while CL+LSTM averages $93.15$ in the first 19 tasks.}
  \label{fig:forget2}
\end{minipage}
\end{figure}

\textbf{Task Details.} We use the Omniglot dataset for our experiments. Following \cite{javed2019meta}, we define each subtask as learning a single class concept. So in this protocol a single online 5-way 5-shot continual learning task is defined as the following ordered set: $\mathcal{T} = \big(\mathcal{T}_{1},\mathcal{T}_{2},\mathcal{T}_{3},\mathcal{T}_{4},\mathcal{T}_{5}\big)$. Here, subtask $\mathcal{T}_{i}$ contains 5 samples from 1 particular Omniglot class. After adaptation is done on these 5 subtasks (25 samples) we expect the model to classify samples from a query set consisting of samples from all of the subtasks. The performance of the model is the prediction accuracy on the query set. We experiment by varying the total number of subtasks from 5 to 20 as in Figure~\ref{fig:cont_res}.

\textbf{Training Details.} We perform episodic training by exposing our model to a variety of continual learning tasks from the training partition. At the end of each continual learning task, the model incurs a loss on the query set. We update our model by backpropagating through this query set loss. Note that during evaluation on the query set, we freeze the persistent states of our model in order to prevent any information leak across the query set. Since propagating gradients across long time steps renders training difficult, we train our model using a simple curriculum of increasing task length every 5K episodes. This improves generalization and convergence. Appendix~\ref{app:ocl} presents more details. Further, we shuffle the labels across tasks in order to prevent the model from memorizing the training classes. During evaluation, we sample tasks from classes the model has not encountered. The model adapts to the subtasks using just the hidden states and then acquires the ability to predict on the query set, which contains samples from all of the subtasks. We use the same class wise disjoint train/test split as in \cite{blake}.

\textbf{Baseline: Online Meta Learning (OML).}
Javed et al.~\cite{javed2019meta} adopt a meta-training strategy similar to MAML. They adapt deeper layers in the inner loop for the current task, while updating the entire network in the outer loop, based on a loss measuring forgetting. For our OML experiments we use a 4-layer CNN followed by two fully connected layers. 

\textbf{Baseline: A Neuromodulated Meta-Learning Algorithm (ANML).} Beaulieu et al.~\cite{beaulieu2020learning} use a hypernetwork to modulate the output of the trunk network. In the inner loop, the trunk network is adapted via gradient descent. In the outer loop, they update both the hypernetwork and the trunk network on a loss measuring forgetting. For our ANML experiments, we use a 4-layer CNN followed by a linear layer as the trunk network, with a 3-layer hypernetwork modulating the activations of the CNN. They use 3 times as many parameters as our CL+LSTM model. 

\textbf{Results.} Figure~\ref{fig:cont_res} plots average accuracy on increasing the length of the continual learning task. Task length is the number of subtasks within each continual learning task, which ranges from 5 to 20 subtasks in our experiments. As expected, we observe that the average accuracy generally decreases with increased task length for all models. However, the CL+LSTM model's performance degrades slower than the baselines, suggesting that the model has learnt an efficient way of storing inductive biases required to solve each of the subtasks effectively.

From Figure~\ref{fig:forget2}, we see that CL+LSTM is robust against forgetting, as the variance on performance across subtasks is low. This suggests that the CL+LSTM model learns adaptation rules that minimally interfere with other tasks.

\textbf{Analysis of Computational Cost.} During inference, our model does not require any gradient computation and fully relies on hidden states to perform adaptation. Consequently, it has lower computational requirements compared to gradient-based models -- assuming adaptation is required at every time step. For a comparative case study, let us consider three models and their corresponding GFLOPs per forward pass: OML baseline (1.46 GFLOPs); CL+LSTM (0.40 GFLOPs); 4-layer CNN (0.30 GFLOPs) with parameter count similar to CL+LSTM.  Here, we employ standard methodology for estimating of compute cost \citep{AIandCompute}, with a forward and backward pass together incurring three times the operations in a forward pass alone.

We can extend these estimates to compute GFLOPs for the entire adaptation phase. Suppose we are adapting/updating our network on a task of length $t$ iterations.  The OML baseline and the 4-layer CNN (adapting on gradient descent) would consume 4.38$t$ GFLOPs and 0.9$t$ GFLOPs, respectively.  Our CL+LSTM model would consume only 0.40$t$ GFLOPs; here, we drop the factor of three while computing GFLOPs for the CL+LSTM model, since we do not require any gradient computation for adaptation.  During training, we lose this advantage since we perform backpropagation through time, making the computational cost similar to computing meta-gradients.
\subsection{Online few-shot Semantic Segmentation}
\label{sec:semseg}
\begin{figure}[t]
\centering
\includegraphics[width=\linewidth]{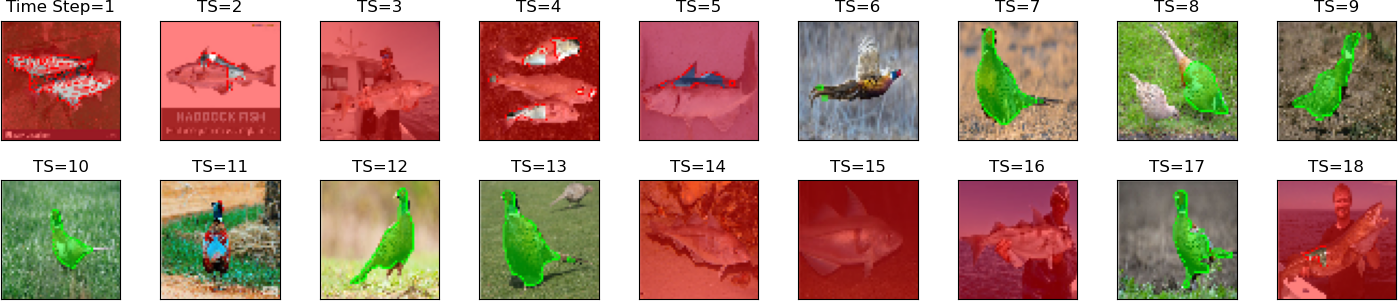}
\caption{\textbf{Sample online few-shot segmentation task with distractors.} At each time step, the model gets one image, and its corresponding ground truth in the subsequent time step. The model is tasked with either segmenting or masking, depending on whether or not the image is a distractor. Here, the fish are distractors and the ducks are objects we want to segment. The portion highlighted in green is the predicted segmentation, and the red portion is the predicted mask. These are results from our 10-ConvLSTM model with label injection at the first layer. \vspace{-0.3cm}}
\label{fig:semseg}
\end{figure}

These experiments investigate the efficacy and applicability of adaptation via persistent states to a challenging segmentation task and analyze the effectiveness of label injection for segmentation.

\textbf{Task Details.} We consider a binary segmentation task: we present the model a sequence of images, one at each time step (as in Figure~\ref{fig:semseg}), and the model must either segment or mask out the image based on whether it is a distractor. Similar to the classification tasks, we augment the ground truth segmentation information along the channel dimension. The ground truth is offset by 1 time step, so at the first time step we concatenate to the channel dimension an all -1 matrix as a null label, at the next time step we concatenate to the channel dimension the actual ground truth of the image at time step 1.  If it is an image to be segmented, we concatenate the ground truth binary mask of the object and the background in the form of a binary matrix. If it is a distractor image, we concatenate to the channel dimension an all zeros matrix indicating that the entire image should be masked out. $k$-shot scores for segmentation is the IoU of the predicted segmentation on the ${k+1}^{th}$ time the model sees the object we want to segment out. For $k$-shot masking scores, we compute the fraction of the object that has been masked, when model sees the distractor image for the ${k+1}^{th}$ time step. We sample our episodes from the dataset FSS1000 \cite{li2020fss}; more dataset details are in Appendix~\ref{app:seg}.

The construction of this task avoids zero-shot transfer of inductive biases required for segmentation and forces the model to rely on the task data to learn which objects are to be segmented.

\textbf{Training Details.} We augment a 10 layer U-Net \cite{ronneberger2015u} like CNN with memory cells in each layer, by converting each convolution into a convolutional LSTM--referred to as CL U-Net (architecture details in Appendix~\ref{app:seg}). We utilize episodic training, where each episode is an online few-shot segmentation task, as in Figure~\ref{fig:semseg} with 18 time steps in total (9 segmentation images and 9 distractors). We follow a simple training curriculum to train: the first 100k episodes we train without any distractors; in the next 100k episodes we train with distractors as in Figure~\ref{fig:semseg}. Further training details are in Appendix~\ref{app:seg}. The episodes presented during evaluation contain novel classes.

\textbf{Baselines.} We use a 10 layer U-Net like CNN pre-trained with MAML for segmentation without any distractors (architecture details in Appendix~\ref{app:seg}).  We use this model as our fine-tuning CNN baseline, in that we fine-tune the model on the online stream of images using gradient descent at each time step. From Table~\ref{tab:seg}, we see that the model fails to mask out the distractors, indicating its inability to ability to adapt to the online feed.

From Table~\ref{tab:seg}, we see that CL U-Net variants are capable of effective online adaptation; both models are capable of segmenting and masking images. However we observe that providing label information at the first layer significantly boosts our performance, thereby bolstering our claim that effective task adaptation can be achieved by providing relevant feedback to a network containing distributed memory.
\begin{table*}[t]
\centering
\scalebox{1}{
\begin{tabular}[t]{cccccccc}
\toprule
\multicolumn{1}{c}{\multirow{2}{*}{Model}}& \multicolumn{1}{c}{\multirow{2}{*}{Label Injection}} & \multicolumn{2}{c}{Segmentation}  & \multicolumn{2}{c}{Masking}  \\
\multicolumn{1}{c}{} &  \multicolumn{1}{c}{} &  \multicolumn{1}{c}{1-shot} & \multicolumn{1}{c}{6-shot}& \multicolumn{1}{c}{1-shot} &  \multicolumn{1}{c}{6-shot} \\
\midrule
Fine-tuned CNN & No  & 65.3 & 64.0 & 23.1 & 23.6  \\ 
CL U-Net& 9th layer & 38.7& 45.7&73.8& 77.8  \\
CL U-Net& 1st layer & 49.2& 56.0& 84.6& 91.7\\
\end{tabular}}
\caption{\textbf{Segmentation and Masking results on CL U-Net.} Fine-tuning a CNN fails to adapt to distractors, while CL U-Net variants demonstrate adept capacity for adaptation. Further label injection at the first layer outperforms label injection at the $9^{th}$ layer, suggesting that label injection enables network wide adaptation.}
\label{tab:seg}
\end{table*}
\section{Discussion}
Our results highlight distributed memory architectures as a promising technical approach to recasting the problem of meta-learning as simply learning with memory-augmented models.  This view has potential to eliminate the need for ad-hoc design of mechanisms or optimization procedures for task adaptation, replacing them with generic and general-purpose memory modules. Our ablation studies show the effectiveness of distributing memory throughout a deep neural network (resulting in an increased capacity for adaptation), rather than limiting it to a single layer or final classification stage.

We demonstrate that standard LSTM cells, when provided with relevant feedback, can act as a basic building block of a network designed for meta-learning.  On a wide variety of tasks, a distributed memory architecture can learn adaptation strategies that outperform existing methods. The applicability of a purely memory-based network to online semantic segmentation points to the untapped versatility and efficacy of adaptation enabled by distributed persistent states.

\clearpage
\chapter{HyperNetwork Designs for Improved Classification and Robust Few-Shot Learning}

\section{Introduction}

In this chapter, we propose a set of design and training schemes that enhance the capabilities of hypernetworks and extend their practical impact to settings requiring rapid task adaptation. Hypernetworks are networks that generate weights for a target network to solve a given task \cite{ha2016hypernetworks}. We are motivated to explore and improve hypernetworks because of their ability to acquire priors from multiple tasks by generating target networks for each task.

This ability to acquire priors is critical in several applied fields such as drug discovery and computational chemistry, where data is sparse or scarcely available \cite{hudson2023can,dou}. We demonstrate that with our design and training schema, hypernetworks are efficient at acquiring priors and can be used in scenarios with limited pre-training data.

We enhance the hypernetwork's ability to efficiently acquire priors and adapt quickly by training it via Model-Agnostic Meta-learning (MAML) \cite{finn2017model}. The MAML training routine segments multiple few-shot tasks (tasks with few training samples) from a large dataset and trains the hypernetwork to generate target networks that generalize to these crafted few-shot tasks. The multi-task training setup helps the hypernetwork acquire priors, while the training objective of requiring target networks that generalize well to few-shot tasks enables hypernetworks to adapt rapidly to any novel task with just a few examples. We experimentally show that hypernetworks adapt to novel tasks significantly better than standard networks.

Furthermore, the training of hypernetworks introduces a host of optimization issues: incorrect scale of gradients, lack of effective regularization, and insufficient momentum control. To address these challenges, we propose solutions that stabilize the training process and enable hypernetworks to generate robust target ResNet~\cite{he2015deep} weights. We demonstrate via experiments that we improve significantly over naive hypernetwork design and even surpass conventionally trained ResNets on standard vision tasks.

In the ensuing sections, we discuss our design and training strategies for hypernetworks and then present experiments to demonstrate their efficacy in various settings.

\section{Related Work}

The idea of one neural network generating the weights of another neural network was introduced in \cite{schmidhuber1992learning}. This setup was used to handle temporal sequences, where the first network provided context-dependent weights for the second network. Further applications of this setup were later explored in \cite{gomez2005evolving}.

A two-layer fully connected network is used to generate weights for a target ResNet, with the goal of reducing parameters by sharing the hypernetwork across layers of the target network \cite{ha2016hypernetworks}. Each layer in the target network has a low-dimensional embedding, which is mapped by the hypernetwork to ambient weight space. Although this model provides significant parameter reduction, it suffers from performance degradation and slower convergence when compared to standard ResNets.

Issues related to the incorrect scale of hypernetwork-generated weights, which make these models harder to optimize, are identified by \cite{chang2019principled}. These issues result in slower learning and convergence. They address this problem by appropriately scaling the weights of the hypernetwork at initialization, leading to reduced training loss as well as better convergence. However, their initialization scheme {does not handle ResNets as the target networks, and they provide results only on hypernetworks generating weights for CNNs and do not extend to ResNets. As we will see, our hypernetwork strategies generalize to ResNets as target networks while also providing faster convergence.

A simple linear combination hypernetwork approach is suggested in \cite{savarese2019learning}. Each layer has a trainable vector that linearly combines a set of trainable parameters shared across layers. Our approach enables greater flexibility by allowing the hypernetwork to generate task-specific weights without such constraints.

\subsection{Meta-learning}

An approach that learns a learning rule through a genetic algorithm was introduced in \cite{schmidhuber1987evolutionary}. Since then, various strategies for meta-learning (i.e., `learning to learn') have been proposed. In recent years, meta-learning via gradient-based optimization to learn an update rule for the model has been actively studied \cite{hochreiter2001learning,andrychowicz2016learning}.

One of the most prominent meta-learning algorithms is Model Agnostic Meta-Learning (MAML) \cite{finn2017model}. MAML is a routine used to learn an initialization with the goal of quickly converging to a good solution for any given task. It consists of two nested loops: the outer loop finds a meta-initialization, while the inner loop rapidly adapts the meta-initialization for the task at hand. A limitation of MAML is its difficulty in scaling to deeper networks, as demonstrated in its original experiments that were limited to 4-layer architectures.

A variant of MAML that leverages a network pre-trained on a large dataset as initialization, thereby enabling MAML training of deeper networks, is proposed by \cite{sun2019meta}. Once pre-trained, the network is frozen, and during training with MAML, it learns only a small set of parameters that scale and shift the frozen pre-trained weights.

A method that learns an encoder to map training data into a small embedding vector, which is then mapped to model parameters via a decoder, is proposed in \cite{rusu2018meta}. The encoder and decoder parameters are learned in the outer loop, while the embedding vector is updated during the inner loop to enable adaptation for the current task.

It is discovered that MAML-based algorithms perform worse in few-shot learning tasks when training and testing tasks are sampled from different datasets \cite{chen2019closer}.

A fundamental question about MAML is asked by \cite{raghu2019rapid}: Is its success due to its ability to find initializations that generalize to new tasks, or is it that the class-wise features learned from the training set are incidentally suitable for classes in the test set? Their analysis concludes that it is primarily the latter. This discovery helps explain the degradation in performance of MAML models under shift in distributions. Once the distribution of test tasks changes, the features learned from the training tasks no longer generalize, causing a drop in performance.

The ability to adapt to new tasks, or rapid task adaptation, is essential for generalizing to new tasks drawn from a distribution different from the training distribution. As we will see later, HyperResNets trained with MAML are better suited for rapid task adaptation compared to ResNets. Although \cite{li2018learning} modifies MAML to improve task adaptation, their approach is primarily algorithmic, whereas our work focuses on architectural improvements.

\section{Improving Hypernetworks}

We describe several shortcomings in both the design and training of existing hypernetwork-based architectures, along with our proposed remedies.

\begin{figure}[!b]
    \centering
    \includegraphics[width=0.85\linewidth]{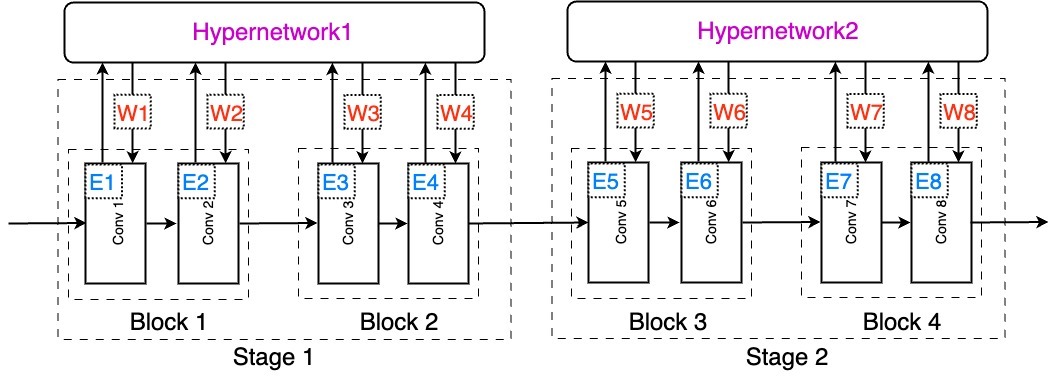}
    \caption{\textbf{Shared hypernetwork.} Each hypernetwork is shared across all layers within a stage of the ResNet. Each layer has a trainable embedding (blue), which is mapped to weights (red) by the hypernetwork. }
    \label{fig:hypnet1}
\end{figure}

\textbf{Hypernetwork Sharing Scheme:}
Typically, the hypernetwork is shared across all layers of the target ResNet \cite{ha2016hypernetworks}. In such a setup, the hypernetwork receives gradient contributions from every layer, potentially making optimization more difficult because gradients from different layers often have differing magnitudes. As a result, Adam \cite{kingma2014adam} is commonly used to optimize the hypernetwork parameters since it adaptively scales the gradients.

{\par}
To address this, we propose two new sharing schemes: (1) sharing the hypernetwork across all layers within each stage of the ResNet (Figure~\ref{fig:hypnet1}), and (2) an unshared scheme, where each layer has its own independent hypernetwork (Figure~\ref{fig:hypnet2}). Both approaches distribute the optimization burden more evenly, helping keep gradient scales in check. This allows for the use of SGD instead of Adam, which has been shown to improve generalization performance \cite{wilson2017marginal}.

\textbf{Hypernetwork and Embedding Design:} In \cite{ha2016hypernetworks}, a fully connected hypernetwork maps an embedding vector to the weights of a target network. This approach, however, imposes strong constraints on the embedding size due to the rapid growth of hypernetwork parameters as the embedding dimension increases. A small embedding restricts the generated weights to have fewer degrees of freedom. For instance, when using a 64-dimensional embedding to generate a convolutional layer with dimensions 16×16×3×3, the number of degrees of freedom is limited to just 64. In contrast, a conventionally trained layer of the same shape would have 2304 independently learned parameters, highlighting the restrictive nature of the embedding-based approach.

{\par}

To overcome this limitation, we adopt a one-layer convolutional hypernetwork. Convolutional layers tend to be more parameter-efficient than fully connected layers, partly because their parameter count does not grow with spatial resolution. This design choice allows us to use larger embeddings without dramatically increasing the hypernetwork’s total parameter count.

{\par}
Specifically, to generate weights for a layer $L_i$, we assign an embedding of shape $C_{\text{out}} \times d \times k \times k$ to $L_i$, and the hypernetwork itself is a convolutional layer with filters of shape $C_{\text{in}} \times d \times k \times k$. The hypernetwork’s output thus has shape $C_{\text{out}} \times C_{\text{in}} \times k \times k$. Here, $C_{\text{in}}$ and $C_{\text{out}}$ are the input and output channels of $L_i$, and $k$ is the kernel size. We typically set $d = C_{\text{in}}$, though this choice can be adjusted as needed. This approach effectively removes the strict degree-of-freedom limitation on the generated weights.

\begin{figure}[ht]
    \centering
    \includegraphics[width=0.65\linewidth]{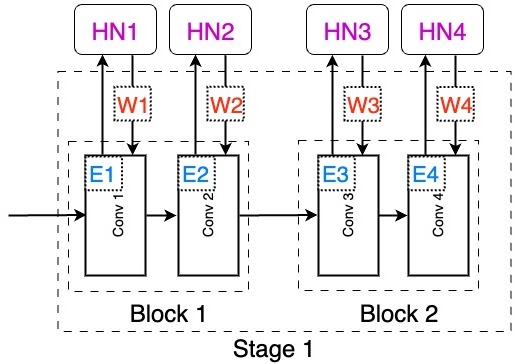}
    \caption{\textbf{Unshared hypernetwork.} There is an independent hypernetwork (HN) associated with each layer in the model.}
    \label{fig:hypnet2}
\end{figure}

\begin{figure*}
     \centering
     \begin{subfigure}[b]{0.3\textwidth}
         \centering
         \includegraphics[width=\textwidth]{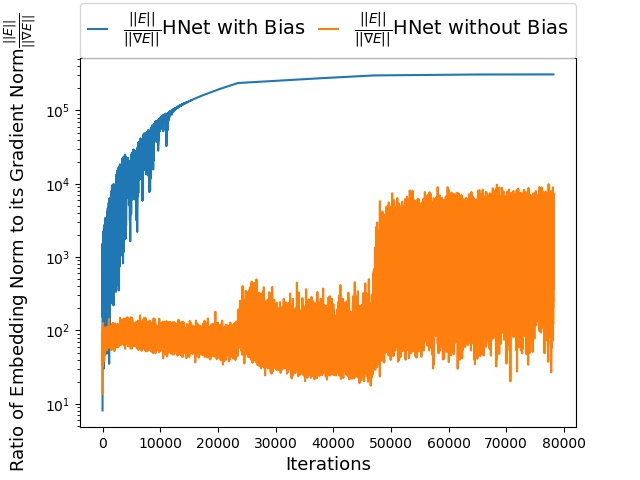}
         \caption{$E$ is the embedding belonging to the second convolution in stage 1 block 1 of the HyperResNet.}
        
     \end{subfigure}
     \hfill
     \begin{subfigure}[b]{0.3\textwidth}
         \centering
         \includegraphics[width=\textwidth]{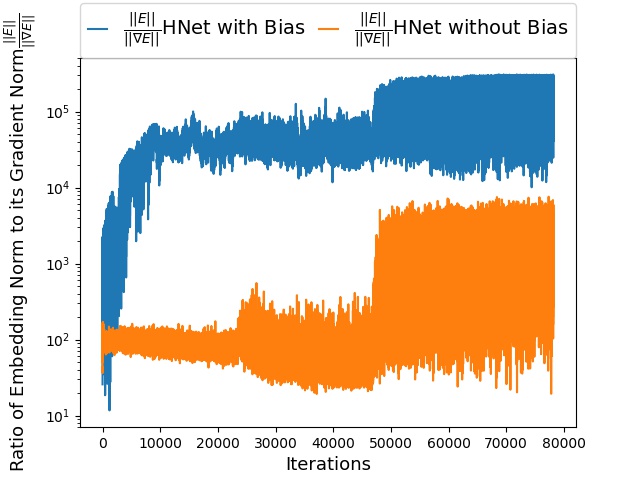}
         \caption{$E$ is the embedding belonging to second convolution in stage 2 block 1 of the HyperResNet.}
      
     \end{subfigure}
     \hfill
     \begin{subfigure}[b]{0.3\textwidth}
         \centering
         \includegraphics[width=\textwidth]{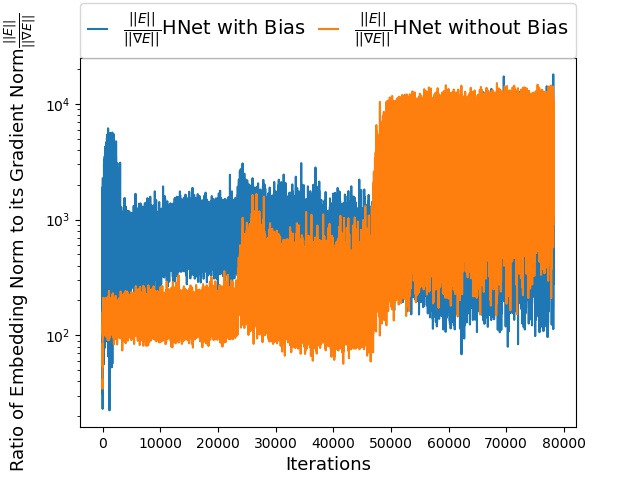}
         \caption{$E$ is the embedding belonging to second convolution in stage 3 block 1 of the HyperResNet.}
   
     \end{subfigure}
        \caption{Plots showing ratio of embedding norm to its gradient norm for embeddings in different stages of the hyperresnet. The higher ratio of  $\frac{\|E\|}{\|\nabla E\|}$ suggests slow convergence with bias term in the hypernetwork.}
        \label{fig:gradient_scale}
\end{figure*}

\textbf{Correcting the Scale of Hypernetwork's Gradients:}
To understand the slow convergence of naive hypernetworks, we measure the ratio of a embedding parameters' norm to its gradient norm, $\frac{\|E\|}{\|\nabla E\|}$. As shown in Figure~\ref{fig:gradient_scale}, these ratios are notably higher than those typically reported for convolutional layers in networks like AlexNet \cite{krizhevsky2012imagenet,you2017large}, indicating limited parameter movement from initialization. Furthermore, the ratio remains high even in the early stages of training, potentially explaining the observed poor convergence and generalization.

{\par}
We find that removing the bias term from the hypernetwork’s convolutional layer mitigates this issue, effectively restricting the layer to purely affine transformations. As shown in Figure~\ref{fig:gradient_scale}, removing the bias reduces $\frac{\|E\|}{\|\nabla E\|}$ early in training, improving parameter updates. As training progresses, the ratio naturally increases due to learning rate decay, which aligns with the expected behavior of convergence.

\textbf{Regularizing Hypernetworks:}
Weight decay on model parameters is the standard regularization choice for typical networks, but in our experiments, applying weight decay directly to hypernetwork parameters did not significantly improve generalization. Instead, we regularize the $\ell_2$-norm of the generated weights (i.e., the hypernetwork’s output), analogous to regularizing the weights in a conventional ResNet. This method enhances classification performance for hypernetworks. We also do not apply weight decay to the embeddings or the hypernetwork’s own parameters.

\textbf{Momentum Control:}
Many networks benefit from Stochastic Weight Averaging (SWA), in which the current iterate is influenced by both the present gradient and an exponential moving average of past iterates \cite{izmailov2018averaging}. This average serves as a momentum term to reduce the effect of noise in the gradients.

{\par}
Because the hypernetwork’s predicted weights depend on both the embedding and the hypernetwork parameters, the effects of noisy gradients can be amplified. To counteract this, we apply SWA to both the embeddings and hypernetwork parameters, observing a stronger benefit compared to standard ResNets.

\subsection{Background: Model-Agnostic Meta-Learning for Few-Shot Learning}
MAML aims to learn a meta-initialization that can quickly adapt to new tasks with minimal updates \cite{finn2017model}. In principle, this allows a network to efficiently adapt to a novel task, achieving high performance with only a few updates.

{\par}
In the few-shot setting, MAML samples a batch of tasks $\{T_{1}, \ldots, T_{n}\}$ from a task distribution $\mathcal{P}(T)$. This batch is known as a meta-batch. Each task $T_{i}$ is divided into a training subset $T^{\text{tr}}_{i}$ and a testing subset $T^{\text{te}}_{i}$. The model is first updated using $T^{\text{tr}}_{i}$ and subsequently evaluated on $T^{\text{te}}_{i}$ to measure its adaptability.

{\par}
\textbf{MAML Stage 1}:
In the first stage, each task $T_{i}$ in the meta-batch is used to create a temporary model $\theta_{i}$ by performing a gradient update on the loss computed over $T^{\text{tr}}_{i}$ (Eq.~\ref{maml:equ1}). This yields a set of adapted models $ \{\theta_{1}, \theta_{2}, \ldots, \theta_{n}\}$. The inner-loop loss $L_{T^{\text{tr}}_{i}}(f_{\theta})$ is typically a standard loss (e.g., cross-entropy).

\begin{equation}
\theta_{i}= \theta -\alpha \nabla_{\theta} L_{T^{tr}_{i}}(f_{\theta})
\label{maml:equ1}
\end{equation}

{\par}
\textbf{MAML Stage 2}:
Next, each adapted model $\theta_{i}$ is evaluated on its testing split $T^{\text{te}}_{i}$, yielding a loss $L_{T^{\text{te}}_{i}}(f_{\theta_{i}})$. The outer loop then updates the original parameters $\theta$ to $\theta_{next}$ by minimizing the sum of these testing losses (Eq.~\ref{maml:equ2}).

\begin{equation}
\theta_{next}= \theta -\beta \nabla_{\theta} \sum_{i=1}^{n} L_{T^{te}_{i}}(f_{\theta_{i}})
\label{maml:equ2}
\end{equation}

Here, $\beta$ is the learning rate in the outer loop. The process repeats for subsequent meta-batches, ultimately producing an initialization $\theta$ that quickly adapts to new tasks. The MAML routine is summarized in Algorithm~\ref{alg:maml}.

\begin{algorithm}
\caption{MAML Routine}
\begin{algorithmic}[1]
\State \textbf{Require:} Collection of tasks $T_1, T_2, \ldots, T_n$.
\State \textbf{Require:} Hyperparameters $\alpha$, $\beta$.
\State Initialize network parameters $\theta$.
\State Initialize preconditioner $\mathcal{P}$ \hfill
\While{not done}
    \State Sample a batch of tasks $\mathcal{B} = \{T_1, T_2, \ldots, T_k\}$.
    \For{$T_i \in \mathcal{B}$}
        \State Compute $\mathcal{L}_{T_i^{tr}}(f_{\theta})$.
        \State Update $\theta_i = \theta - \alpha\, \mathcal{P}\,\nabla_{\theta} \mathcal{L}_{T_i^{tr}}(f_{\theta})$.
    \EndFor
    \State Update $\theta \leftarrow \theta - \beta\, \nabla_{\theta} \sum_{T_i^{te} \in \mathcal{B}} \mathcal{L}_{T_i^{te}}(f_{\theta_i})$.
    \State Update $\mathcal{P} \leftarrow \mathcal{P} - \beta\, \nabla_{\mathcal{P}} \sum_{T_i^{te} \in \mathcal{B}} \mathcal{L}_{T_i^{te}}(f_{\theta_i})$.
\EndWhile
\State \Return $\theta$.
\end{algorithmic}
\label{alg:maml}
\end{algorithm}

\begin{algorithm}
\caption{Training HyperResNets via MAML Routine}\label{hypmaml}
\begin{algorithmic}[1]
\Require Collection of tasks $T_{1}, T_{2} \cdots T_{n}$ 
\Require Pre-trained network  $ \theta = \{\theta_{E}, \theta_{H}, \theta_{C} \}$. Where $\theta_{E}, \theta_{H},\theta_{C}$ are embedding, hypernetwork, and classifier parameters.
\Require Hyperparameters $\alpha$ and $\beta$.
    \While {not done}
    \State Sample a batch of tasks $\mathcal{B}$ = \{$T_{1}, T_{2}, \cdots, T_{k}$\}
    \ForAll{$T_{i}\in B$}
    \State Compute $\mathcal{L}_{T^{tr}_{i}}(f_{\theta})$
    \State Update $\theta^{i}_{E} = \theta_{E} - \alpha \nabla_{\theta_{E}}\mathcal{L}_{T^{tr}_{i}}(f_{\theta})$
    \State Update $\theta^{i}_{C} = \theta_{C} - \alpha \nabla_{\theta_{C}}\mathcal{L}_{T^{tr}_{i}}(f_{\theta})$
    \State $\theta^{i} = \{\theta^{i}_{E}, \theta_{H}, \theta^{i}_{C}\}$
    \EndFor
    \State Update $\theta_{H^{'}} = \theta_{H} - \beta \nabla_{\theta_{H}}\sum_{T^{te}_{i}}\mathcal{L}_{T^{te}_{i}}(f_{\theta^{i}})$
    \State Update $\theta_{E^{'}} = \theta_{E} - \beta \nabla_{\theta_{E}}\sum_{T^{te}_{i}}\mathcal{L}_{T^{te}_{i}}(f_{\theta^{i}})$
   
    \State Update $\theta_{C^{'}} = \theta_{C} -\beta \nabla_{\theta_{C}}\sum_{T^{te}_{i}}\mathcal{L}_{T^{te}_{i}}(f_{\theta^{i}})$
    \State Update $\theta = \{\theta_{E^{'}}, \theta_{H^{'}}, \theta_{C^{'}}\}$
    \EndWhile
    \State \Return $\theta$ 
\end{algorithmic}
\label{alg:hypermaml}
\end{algorithm}

\subsection{MAML-Based Training for Hypernetworks}
\textbf{Pre-training Phase:}
Using MAML to train deep networks from scratch often yields near-chance accuracy. A common practice to circumvent this is to pre-train on a large dataset \cite{sun2019meta}. Hence, we first pre-train our hypernetwork on a large dataset, then replace the final classification layer with a randomly initialized linear layer before initiating meta-training.

\textbf{Meta-training Phase:}
Algorithm~\ref{alg:hypermaml} details how to train a hypernetwork in the shared setting via MAML. The procedure is similar to MAML for a standard ResNet, except that the embedding parameters are updated in the inner loop, while the hypernetwork parameters are updated in the outer loop. The final classification layer is updated in both loops.

{\par}
By updating the hypernetwork parameters using the outer-loop gradients (accumulated across multiple tasks), the hypernetwork learns to produce weights that generalize well across tasks, acting as a form of regularization. In shared-hypernetwork training, the embedding often has more parameters than the hypernetwork itself, making it beneficial to focus updates on the embedding in the inner loop, thereby enabling task-specific adaptation.

\section{Results}
\subsection{Image Classification on CIFAR}

\textbf{CIFAR Dataset:} 
CIFAR-10 and CIFAR-100 are standard benchmarks for image classification \cite{krizhevsky2009learning}. CIFAR-10 has 10 classes, each with 5000 training images and 1000 testing images, while CIFAR-100 has 100 classes with 500 training and 100 testing images per class. All images are 32\(\times\)32 pixels.
{\par}

\begin{table}[ht]
\centering
\scalebox{1}{
\begin{tabular}{lcc}
\toprule
\textbf{Model} & \textbf{CIFAR-10} & \textbf{CIFAR-100}\\
\midrule
HyperWRN-40-2 (unshared variant) W Bias  & 91.89 & 68.15\\
HyperWRN-40-2 (unshared variant) W/O Bias & \textbf{93.99} & \textbf{73.42}\\
\midrule
HyperWRN-28-4 (unshared variant) W Bias  & 92.96 & 70.73\\
HyperWRN-28-4 (unshared variant) W/O Bias & \textbf{94.72} & \textbf{76.03}\\
\bottomrule
\end{tabular}
}
\caption{CIFAR classification accuracies. Removing the bias in the hypernetwork (unshared variant) significantly improves generalization.}
\label{Tab:bad-bias}
\end{table}

\begin{figure}[bt!]
     \centering
     \includegraphics[width=.60\textwidth]{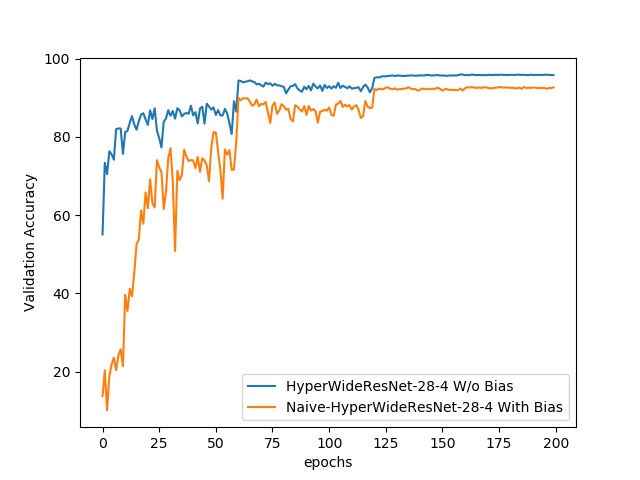}
     \caption{Validation curves on CIFAR-10. HyperWRN-28-4 converges faster and achieves higher accuracy when bias is removed.}
     \label{bd-bias}
\end{figure}

\textbf{Ablations:}
From Table \ref{Tab:bad-bias}, we see that removing the bias in the hypernetwork leads to substantial accuracy gains for all architectures tested. Figure~\ref{bd-bias} similarly shows faster convergence when the hypernetwork bias is removed. In Table \ref{Tab:good-reg}, we observe additional performance boosts by regularizing the hypernetwork’s generated weights. Together, these results highlight the effectiveness of our proposed design and training strategies.
{\par}
\textbf{Comparing hypernetwork to standard networks:}
Tables \ref{wrn} and \ref{rn} show that the unshared hypernetwork variant consistently outperforms standard ResNets and WideResNets (WRNs) on CIFAR-10 and CIFAR-100 across different depths and widths, while the shared variant remains competitive. On CIFAR-100, the unshared approach beats the baseline by roughly 0.60\% in absolute terms for most architectures. Although gains on CIFAR-10 are more modest, they become meaningful when viewed against deeper baselines. For instance, a HyperResNet-18 surpasses a standard ResNet-34 on CIFAR-10, and a HyperWRN-28-4 performs comparably to a standard WRN-100-4. These findings suggest that, at least for CIFAR, dedicating parameters to the hypernetwork is more beneficial than simply increasing the depth of the target network.

{\par}
In Table \ref{swa}, we see that on CIFAR-10, a HyperWRN-28-4 extends its lead over a standard WRN-28-4 from 0.25\% to 0.40\% when trained with SWA. On CIFAR-100, the margin grows from 0.60\% to 0.85\%.

{\par}
\textbf{Experimental details:} All models are trained with the same hyperparameters as \cite{zagoruyko2016wide}, using SGD for 200 epochs with momentum 0.9 and an initial learning rate of 0.1, which is reduced fivefold at epochs 60, 120, and 160. Standard WRNs and ResNets apply weight decay of \(5 \times 10^{-4}\), while their hypernetwork counterparts use an \(\ell_2\) penalty of \(6.25 \times 10^{-5}\) on generated outputs. Each experiment is run three times, and we report mean validation accuracies. The hyperparameters for WRN follows \cite{zagoruyko2016wide}, and ResNet follows \cite{devries2017improved}.

\begin{table}[ht]
\centering

\scalebox{1}{
\begin{tabular}{lcc}
\toprule
\textbf{Model} & \textbf{CIFAR-10} & \textbf{CIFAR-100}\\
\midrule
HyperWRN-40-2 (unshared variant) W/O Regularization & 93.99 & 73.42\\
HyperWRN-40-2 (unshared variant) W Regularization   & \textbf{95.20} & \textbf{76.77}\\
\midrule
HyperWRN-28-4 (unshared variant) W/O Regularization & 94.72 & 76.03\\
HyperWRN-28-4 (unshared variant) W Regularization   & \textbf{95.90} & \textbf{79.20}\\
\bottomrule
\end{tabular}
}
\caption{CIFAR classification accuracies showing improved generalization after regularizing the hypernetwork’s output.}
\label{Tab:good-reg}
\end{table}

\begin{table}[ht]
\centering
\scalebox{0.9}{
\begin{tabular}{lcccc}
\toprule
\textbf{Model} & \textbf{CIFAR-10} & \textbf{CIFAR-100} & \textbf{Train Params} & \textbf{Test Params}\\
\midrule
WRN-40-2 & 94.80 & 76.06 & 2.2M & 2.2M\\
HyperWRN-40-2 (shared variant) & 95.00 & 75.70 & 2.4M & 2.2M\\
HyperWRN-40-2 (unshared variant) & \textbf{95.20} & \textbf{76.77} & 4.17M & 2.2M\\
\midrule
WRN-28-4 & 95.65 & 78.62 & 5.8M & 5.8M\\
HyperWRN-28-4 (shared variant) & 95.65 & 78.49 & 6.6M & 5.8M\\
HyperWRN-28-4 (unshared variant) & \textbf{95.90} & \textbf{79.20} & 10.3M & 5.8M\\
\midrule
WRN-100-4 & 95.90 & 79.23 & 24.4M & 24.4M\\
\bottomrule
\end{tabular}
}
\caption{CIFAR classification accuracies. HyperWRN unshared variant surpasses standard WRNs, while the shared variant is competitive. After training, the hypernetwork can be reparameterized into a standard WRN with no extra overhead at inference.}
\label{wrn}
\end{table}

\begin{table}[ht]
\centering
\scalebox{0.85}{
\begin{tabular}{lccccc}
\toprule
\textbf{Model} & \textbf{CIFAR-10} & \textbf{CIFAR-100} & \textbf{ImageNet} & \textbf{Train Params} & \textbf{Test Params}\\
\midrule
ResNet-18 & 95.27 & 77.92 & 70.05 & 11.1M & 11.1M\\
HyperResNet-18 (unshared variant) & \textbf{95.52} & \textbf{78.81} & 70.01 & 21.4M & 11.1M\\
\midrule
ResNet-34 & 95.34 & 79.00 & 73.58 & 21.3M & 21.3M\\
HyperResNet-34 (unshared variant) & \textbf{95.62} & \textbf{79.30} & \textbf{73.87} & 41.6M & 21.3M\\
\bottomrule
\end{tabular}
}
\caption{Analogous results for ResNet-based architectures. HyperResNet unshared variant shows consistent improvements with minimal overhead at inference.}
\label{rn}
\end{table}

\begin{table}[ht]
\centering
\scalebox{0.9}{
\begin{tabular}{lcccc}
\toprule
\textbf{Model} & \textbf{CIFAR-10} & \textbf{CIFAR-100} & \textbf{Train Params} & \textbf{Test Params}\\
\midrule
WRN-28-4 + SWA & 95.86 & 80.76 & 5.8M & 5.8M\\
HyperWRN-28-4 (unshared variant) + SWA & \textbf{96.35} & \textbf{81.59} & 10.3M & 5.8M\\
\bottomrule
\end{tabular}
}
\caption{CIFAR accuracies with 300 epochs of SWA training. HyperWRN-28-4 unshared variant benefits more from SWA than the standard WRN-28-4.}
\label{swa}
\end{table}

\subsection{Image Classification on ImageNet}

ImageNet is a large-scale dataset with 1.2 million training images and 50,000 validation images, covering 1,000 classes \cite{deng2009imagenet}.

{\par}
From Table \ref{rn}, we observe modest gains on ResNet-34 and essentially parity on ResNet-18. Notably, HyperResNet models converge faster—for example, at around 70\% of the total training budget, HyperResNet-34 matches the final accuracy of a standard ResNet-34. A similar trend holds for ResNet-18.
{\par}
\textbf{Experimental Details:}
We follow \cite{savarese2019learning} for ImageNet training: 100 epochs of SGD with momentum 0.9, an initial learning rate of 0.1, and a 10x decay every 30 epochs. ResNets use weight decay of \(10^{-4}\), whereas hypernetwork outputs are regularized with \(6.25 \times 10^{-5}\). The ResNet architecture follows \cite{he2016deep}.

\subsection{Meta-Learning Experiments}

We design meta-learning experiments to evaluate how well models adapt to new tasks under distribution shifts between training and testing data. We measure performance using few-shot learning tasks.

\textbf{MiniImageNet.}
MiniImageNet is a subset of ImageNet with 100 classes: 64 for training, 16 for validation, and 20 for testing. Each class has 600 color images at 84\(\times\)84 resolution \cite{vinyals2016matching}.

\textbf{FewShot-CIFAR-100.} FewShot-CIFAR-100 is a widely used few-shot benchmark derived from CIFAR-100. We use its training split of 60 classes in our experiments \cite{oreshkin2018tadam}.

\begin{figure*}[bt]
    \centering
    \begin{minipage}[b]{0.3\textwidth}
        \centering
        \includegraphics[width=\textwidth]{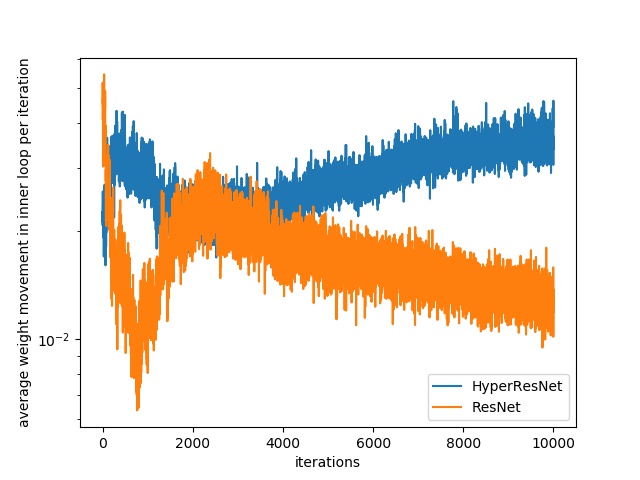}
    \end{minipage}
    \hfill
    \begin{minipage}[b]{0.3\textwidth}
        \centering
        \includegraphics[width=\textwidth]{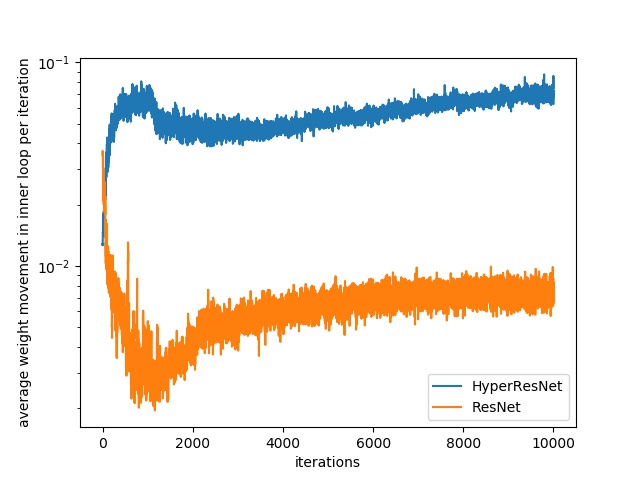}
    \end{minipage}
    \hfill
    \begin{minipage}[b]{0.3\textwidth}
        \centering
        \includegraphics[width=\textwidth]{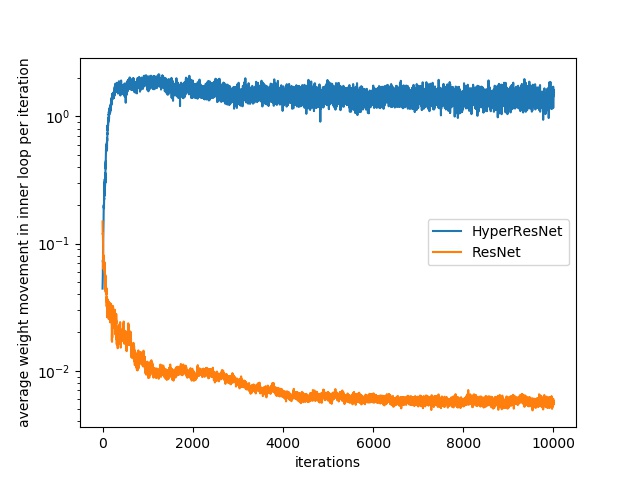}
    \end{minipage}
    \caption{Plots of average inner-loop parameter movement for hypernetwork-predicted weights (blue) versus standard ResNet weights (orange) at three different network depths. Hypernetwork-generated weights exhibit significantly larger movement, suggesting a stronger adaptation to the task during the inner loop.}
    \label{fig:wt_movement}
\end{figure*}

\textbf{Experimental Details:}
We test models on 1-shot and 5-shot, 5-way classification tasks. For 1-shot, each class provides just one training example. The inner loop uses gradient descent with a 0.01 learning rate, while the outer loop is optimized by Adam starting at 0.001 \cite{sun2019meta} and decayed by half every 1000 iterations. We use a meta-batch size of 4, and each task has 15 test samples per class.

\textbf{Experiments on the Standard Setting:}
In the canonical few-shot setup, we train on the 64-class training split of MiniImageNet and test on its 20-class test split. As shown in Table \ref{canon}, hypernetworks are slightly better on 1-shot tasks, while standard ResNets perform better on 5-shot tasks. Since training and test splits are closely related, this scenario introduces little distribution shift, making it a less sensitive test of rapid adaptation.

\begin{table}[ht]
\centering
\begin{tabular}{lcc}
\toprule
\textbf{Model} & \textbf{1-shot 5-way} & \textbf{5-shot 5-way}\\
\midrule
ResNet-12 & 57.65 & \textbf{74.33}\\
HyperResNet-12 & \textbf{58.00} & 73.00\\
\bottomrule
\end{tabular}
\caption{1-shot and 5-shot accuracies on 5-way MiniImageNet tasks with training, validation, and testing all on MiniImageNet. HyperResNet-12 and ResNet-12 are similar in performance when there is lesser distribution shift between training and testing tasks.}
\label{canon}
\end{table}

\textbf{Realistic and Robust Setting:}
To impose a larger distribution shift, we train our meta-learner on few-shot tasks sampled from the FewShot-CIFAR-100 train split, then evaluate on the MiniImageNet test split. This reflects a more realistic scenario where the training distribution may differ considerably from the test distribution, which is especially relevant in few-shot problems with limited data.

As shown in Table \ref{cifar2mini}, HyperWRNs outperform standard WRNs by a notable margin on both 1-shot and 5-shot tasks, demonstrating better rapid adaptation. Deeper WRNs often degrade in performance \cite{chen2019closer}, but the decline is more pronounced in standard WRNs than in HyperWRNs. Moreover, Figure~\ref{conv3} shows that hypernetworks converge better than ResNets when evaluated on cross-domain tasks.

\begin{table}[ht]
\centering
\begin{tabular}{lcc}
\toprule
\textbf{Model} & \textbf{1-shot 5-way} & \textbf{5-shot 5-way}\\
\midrule
WRN-16-4 & 40.75 & 53.20\\
HyperWRN-16-4 & \textbf{43.17} & \textbf{58.66}\\
\midrule
WRN-28-4 & 39.65 & 49.80\\
HyperWRN-28-4 & \textbf{43.56} & \textbf{58.30}\\
\bottomrule
\end{tabular}
\caption{1-shot and 5-shot accuracies (5-way tasks) when pre-training and meta-training on FewShot-CIFAR-100, then meta-testing on MiniImageNet. HyperWRNs significantly surpass standard WRNs.}
\label{cifar2mini}
\end{table}

\begin{figure}[bt!]
\vspace{-0.2in}
\centering
\includegraphics[width=.35\textwidth]{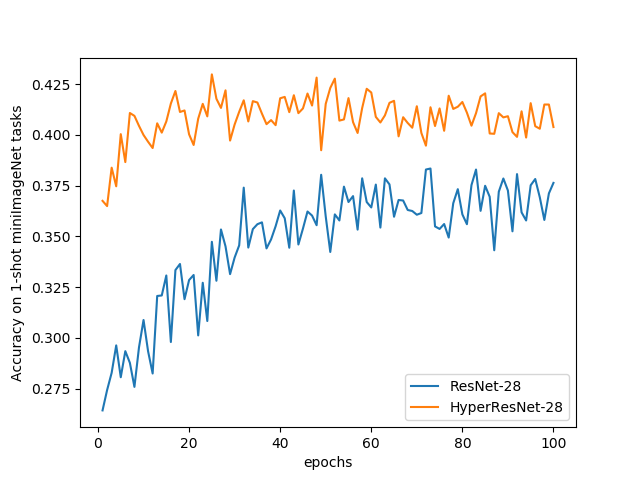}
\caption{Validation curves for 1-shot and 5-shot tasks (same setup as Table \ref{cifar2mini}). A hypernetwork-generated WRN-28-4 trained via MAML converges more effectively than the standard WRN-28-4.}
\label{conv3}
\end{figure}

\section{Domain Generalization Experiments}

The PACS dataset is a popular benchmark for evaluating domain generalization methods \cite{li2017deeper}. It contains four distinct visual domains: Photo, Art Painting, Cartoon, and Sketch, with images drawn from seven categories shared across all domains. The domain generalization task on PACS involves training a model on three of the four domains and evaluating its zero-shot performance on the held-out domain\textemdash one that the model has never seen during training. This task is designed to test a model’s ability to generalize to new distributions without  training on them.

Our baseline method follows a standard approach where we use an ImageNet pre-trained ResNet-18 and fine-tune it on the PACS dataset using the Representation Self-Challenging (RSC) technique \cite{huang2020self}. RSC is a domain generalization method that improves model robustness by masking dominant feature representations during training, forcing the network to learn more generalizable features. 

In our method, we replace the standard ImageNet pre-trained ResNet-18 with a hypernetwork pre-trained on ImageNet. We then train this hypernetwork on PACS using the same RSC technique. As shown in Table \ref{tab:DG}, the HyperResNet-18 consistently achieves superior performance compared to the baseline ResNet-18 across all PACS domains. Notably, the largest improvements occur in the high-domain-shift categories, such as Sketch and Cartoon.

The results demonstrate the effectiveness of hypernetworks for zero-shot out-of-distribution (OOD) generalization. The observed performance gains suggest that hypernetworks facilitate the learning of more transferable and robust feature representations compared to the baseline ResNet-18. This experiment further underscores the potential of hypernetworks as a tool for domain generalization, particularly when paired with effective regularization techniques like RSC.

\begin{table}
\centering
 \begin{tabular}[t]{lcccccc} 
 \toprule
 Method &Photo & Cartoon & Sketch& Art painting  & Mean \\ 
  \midrule
ResNet 18 (code)& 93.13  & 76.92 & 79.19 & 79.19 &  82.10\\
HyperResNet-18& \textbf{94.63} & \textbf{79.05} & \textbf{81.10} & \textbf{81.00} &\textbf{83.94}\\
\hdashline
R-18 (paper) &  95.99  & 80.31 & 80.85 & 83.43 & 85.15 \\ 

\end{tabular}
\caption{HyperResNet reparameterizations leveraging pre-trained embeddings consistently outperform their ResNet counterparts under the Representation Self Challenging (RSC) framework.}
\label{tab:DG}
\end{table}

\section{Discussion}
The early times of deep learning were marked with numerous obstacles that hindered the training of deep networks. Design and modeling choices such as rectifier activations, batch normalization, and skip connections, have since then enabled the training of powerful neural models. In a similar vein, we believe hypernetworks are a design choice towards neural models that can generalize better to OOD samples. This belief is based on our improved classification and robust meta-learning results. However, akin to the earlier days of deep learning, complex hypernetworks cannot be properly trained yet. Here, we make useful contributions by proposing strategies for successfully training a one convolutional hypernetwork. With this work, we hope to increase interest in hypernetworks as a fruitful object of study for OOD generalization.

\clearpage

\chapter{HyperFields: Towards Zero-Shot Generation of NeRFs from Text}
\label{chap:hf}

\section{Introduction}

\begin{figure}[t]
    \centering
    \includegraphics[width=\textwidth]{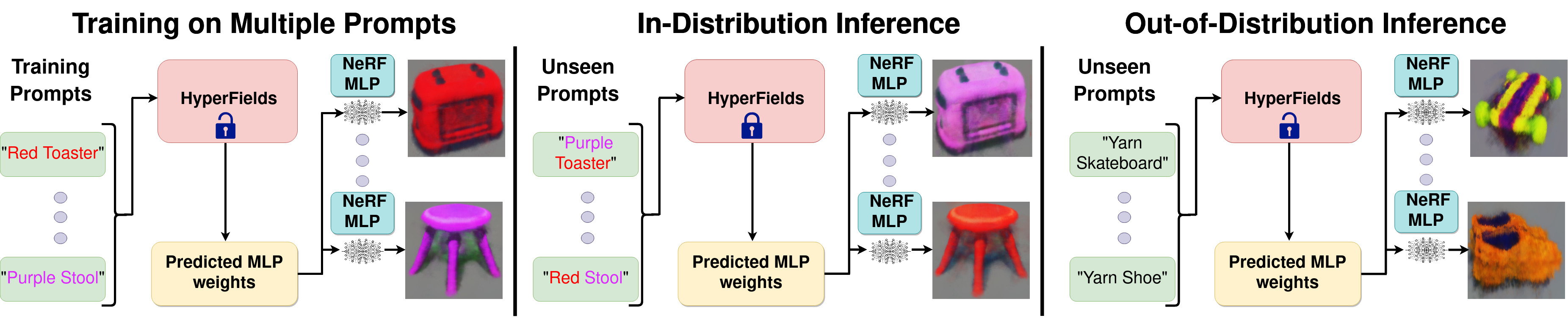} 
    \captionsetup{hypcap=false}
    \caption{\ourmethod{} is a hypernetwork that learns to map text to the space of weights of Neural Radiance Fields (first column). On learning such a mapping, HyperFields is capable of generating in-distribution scenes (unseen during training) in a feed-forward manner (second column), and for unseen out-of-distribution prompts HyperFields can be fine-tuned to yield scenes respecting prompt semantics with just a few gradient steps (third column).}
    \label{fig:teaser}
\end{figure}

%

Motivated by the hypernetwork's ability to adapt faster on novel tasks, in this chapter we extend the application of hypernetworks to generate weights of Neural Radiance Fields (NeRFs)  for the purposes of text-to-3D tasks \cite{NeRF, pixelnerf, dietnerf}. 
{\par}
Most text-conditioned 3D synthesis methods rely on either text-image latent similarity matching or diffusion denoising, both which involve computationally intensive per-prompt NeRF optimization \citep{jain2021dreamfields, dreamfusion,lin2022magic3d}. Extending these methods to bypass the need for per-prompt optimization remains a non-trivial challenge.

{\par}
We propose to solve this problem through a hypernetwork-based neural pipeline, in which a single hypernetwork \citep{hypernetworks} is trained to generate the weights of individual NeRF networks, each corresponding to an unique scene. Once trained, the hypernetwork acquires sufficient priors enabling to efficiently produce the weights of NeRFs corresponding to novel prompts, either through a single forward pass or with minimal fine-tuning. Sharing the hypernetwork across multiple training scenes enables effective transfer of knowledge to new scenes, leading to better generalization and faster convergence. However, we find that a naive hypernetwork design is hard to train. 

{\par}
Our method, \emph{\ourmethod{}}, overcomes these challenges through several design choices. We propose predicting the weights of each layer of the NeRF network in a \emph{progressive} and \emph{dynamic} manner. Specifically, we observe that the intermediate (network) activations from the hypernetwork-predicted NeRF can be leveraged to guide the prediction of subsequent NeRF weights effectively. 

{\par}
To enhance the training of our hypernetwork, we introduce an alternative distillation-based framework rather than the Score Distillation Sampling (SDS) used in \cite{dreamfusion,sjc}. We introduce NeRF distillation, in which we first train individual text-conditioned NeRF scenes (using SDS loss) that are used as teacher NeRFs to provide fine-grained supervision to our hypernetwork (see Figure~\ref{fig:overview}). The teacher NeRFs provide exact colour and geometry labels, eliminating any potentially noisy training signals. 

{\par}
Our NeRF distillation framework allows for training \ourmethod{} on a much larger set of scenes than with SDS, scaling up to 100 different scenes without any degradation in scene quality. A potential explanation for this is that SDS loss exhibits high variance in loss signals throughout different sampling steps. This instability in the loss likely contributes to the challenge of training the hypernetwork on multiple scenes.

{\par}
Once trained, our model can synthesize novel in-distribution NeRF scenes in a single forward pass (Figure~\ref{fig:teaser}, second column) and enables accelerated convergence for out-of-distribution scenes, requiring only a few fine-tuning steps (Figure~\ref{fig:teaser}, third column).  We clarify our use of the terms ``in-distribution" and ``out-of-distribution" in Sections \ref{sec:generalization} and \ref{sec:gen_ood} respectively. These results suggest that our method learns a semantically meaningful mapping. We justify our design choices through ablation experiments which show that both dynamic hypernetwork conditioning and NeRF distillation are critical to our model's expressivity. 

{\par}
Our successful application of dynamic hypernetworks to this difficult problem of generalized text-conditioned NeRF synthesis suggests a promising direction for future work on generalizing and parameterizing neural implicit functions through other neural networks.

\begin{figure}[t]
\begin{center}
  \includegraphics[width=\textwidth]{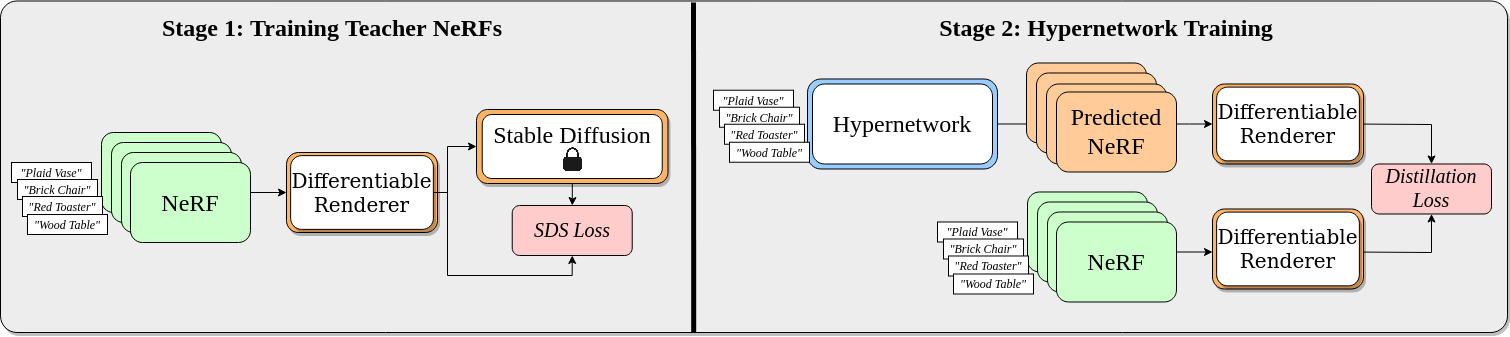}
  \caption{\textbf{Overview.} Our training pipeline proceeds in two stages. \textbf{Stage 1:} We train a set of single prompt text-conditioned teacher NeRFs using Score Distillation Sampling (SDS). \textbf{Stage 2:} We distill these single scene teacher NeRFs into the hypernetwork, through a photometric loss between the renders of the hypernetwork with the teacher network, which we dub our \textit{distillation loss.}}
  \label{fig:overview}
  \end{center}
\end{figure}

\section{Background and Related Work}
Our work combines several prominent lines of work: Neural Radiance Fields (NeRF), score-based 3D synthesis, and learning function spaces using hypernetworks. 


\subsection{3D Representation via Neural Radiance Fields}
There are many competing methods of representing 3D data  in 3D generative modeling, such as point-clouds \citep{pointe, PointVoxelDiffusion}, meshes \citep{text2mesh, AvatarCLIP, LatentNeRF, LION}, voxels \citep{CLIPForge, TextCraft}, and signed-distance fields \citep{Neus, volsdf, kiloneus}. This work explores the popular representation of 3D scenes by Neural Radiance Fields (NeRF) \citep{NeRF, neuralfields, NeRF3DVision}. NeRFs were originally introduced to handle the task of multi-view reconstruction, but have since been applied in a plethora of 3D-based tasks, such as photo-editing, 3D surface extraction, and large/city-scale
3D representation~\citep{NeRF3DVision}.

There have been many improvements on the original NeRF paper, especially concerning training speed and fidelity \citep{PureCLIPNeRF, TensoRF, instantngp, directvoxel, plenoctree}. \ourmethod{} uses the multi-resolution hash grid introduced in InstantNGP~\citep{instantngp}.


\subsection{Score-Based 3D Generation}
While many works attempt to directly learn the distribution of 3D models via 3D data, others opt to use guidance from 2D images due to the vast difference in data availability. Such approaches replace the photometric loss in NeRF's original objective with a guidance loss. The most common forms of guidance in the literature are from CLIP \citep{clip} or a frozen, text-conditioned 2D diffusion model. The former methods minimize the cosine distance between the image embeddings of the NeRF's renderings and the text embedding of the user-provided text prompt \citep{jain2021dreamfields, PureCLIPNeRF, dietnerf}.

Noteworthy 2D diffusion-guided models include DreamFusion \citep{dreamfusion} and Score Jacobian Chaining (SJC) \citep{sjc}, which feed noised versions of images rendered from a predicted NeRF into a frozen text-to-image diffusion model (Imagen \citep{imagen} and StableDiffusion \cite{rombach2021highresolution}, respectively) to obtain what can be understood as a scaled Stein Score \citep{stein}. Our work falls into this camp, as we rely on score-based gradients derived from StableDiffusion to train the NeRF models which guide our hypernetwork training.

We use the following gradient motivated in DreamFusion:
\begin{equation} 
\nabla_{\theta} \mathcal{L}(\phi,g(\theta)) \triangleq  \mathbb E_{t,c} 
\begin{bmatrix} 
	w(t) (\hat \epsilon_\phi(z_t; y, t)-\epsilon)\frac{\partial x}{\partial \theta})
\end{bmatrix}
\end{equation}
which is similar to the gradient introduced in SJC, with the key difference being SJC directly predicts the noise score whereas DreamFusion predicts its residuals. We refer to optimization using this gradient as \textit{Score Distillation Sampling} (SDS), following the DreamFusion authors. Followup work has aimed at improving 3D generation quality \citep{wang2023prolificdreamer, metzer2023latent,chen2023fantasia3d}, whereas we target an orthogonal problem of generalization and convergence of text-to-3D models.

\textbf{Connections to ATT3D:}
We note that our work is concurrent and independent of ATT3D \citep{lorraine2023att3d}. We are similar in that we both train a hypernetwork to generate NeRF weights for a set of scenes during training and generalize to novel in-distribution scenes without any test time optimization.  On top of the in-distribution generalization experiments, we also demonstrate accelerated convergence to novel out-of-distribution scenes (defined in \ref{sec:gen_ood}), which ATT3D does not.

On the technical side, we primarily differ in our novel dynamic hypernetwork architecture. Our hypernetwork generates the MLP weights of the NeRF, while ATT3D outputs the weights of the hash grid in their InstantNGP model. Importantly, our hypernetwork layers are conditioned on not just the input text prompt, but also the activations of the generated NeRF MLP (\ref{sec:dynamic}). We show through our ablations that this dynamic hypernetwork conditioning is essential to the expressivity of our network, as it enables the network to change its weights for the same scene as a function of the view that is being rendered. In contrast, in ATT3D, the generated hash grid is the same regardless of the view being rendered, potentially resulting in the loss of scene detail.

Finally, ATT3D is built on Magic3D \citep{lin2022magic3d} which is a proprietary and more powerful text-to-3D model than the publicly available stable DreamFusion model \citep{stable-dreamfusion} that we use in most of our experiments. We show that our model is capable of learning high quality and complex NeRF scenes produced by more powerful models such as ProlificDreamer without reduction in generation quality (Section~\ref{sec:prolificdreamer}).

\subsection{Hypernetworks}
Hypernetworks are networks that are used to generate weights of other networks which perform the actual task (task performing network) \citep{ha2016hypernetworks}. Many works attempt to use hypernetworks as a means to improve conditioning techniques. Among these, some works have explored applying hypernetworks to implicit 2D representations \citep{siren, FiLM, hyperstyle}, and 3D representations \citep{scenerepresentationnetworks, lightfield, StylizingNeRF}. Very few works apply hypernetworks to radiance field generation. Two notable ones are HyperDiffusion and Shape-E, which both rely on diffusion denoising for generation \citep{erkocc2023hyperdiffusion, jun2023shapegeneratingconditional3d}. HyperDiffusion trains an unconditional generative model which diffuses over sampled NeRF weights, and thus cannot do text-conditioned generation. Shap-E diffuses over latent codes which are then mapped to weights of a NeRF MLP, and requires teacher point clouds to train. Due to the memory burden of textured point clouds, scene detail is not well represented in Shap-E. Both of these methods have the same limitations of slow inference due to denoising sampling. In contrast, our method predicts NeRF weights dynamically conditioned on the 1) text prompt, 2) the sampled 3D coordinates, and 3) the previous NeRF activations.

An interesting class of hypernetworks involve models conditioned on the activations or inputs of the task-performing network \citep{chen2020dynamic}. These models take the following form: let $h,g$ be the hypernetwork and the task performing network respectively. Then $W = h(a)$, where $W$ acts as the weights of $g$ and $a$ is the activation from the previous layer of $g$ or the input to $g$. These are called dynamic hypernetworks, since the predicted weights change dynamically with respect to the layer-wise signals in $g$. Our work explores the application of dynamic hypernetworks to learning a general map between text and NeRFs. It is to be noted that our hypernetwork is a transformer \cite{vaswani2017attention} network.

\section{Method}
Our method consists of two key innovations, the dynamic hypernetwork architecture and NeRF distillation training. We discuss each of these two components in detail below. 

\begin{figure}[t]
    \centering
    \includegraphics[width=\textwidth]{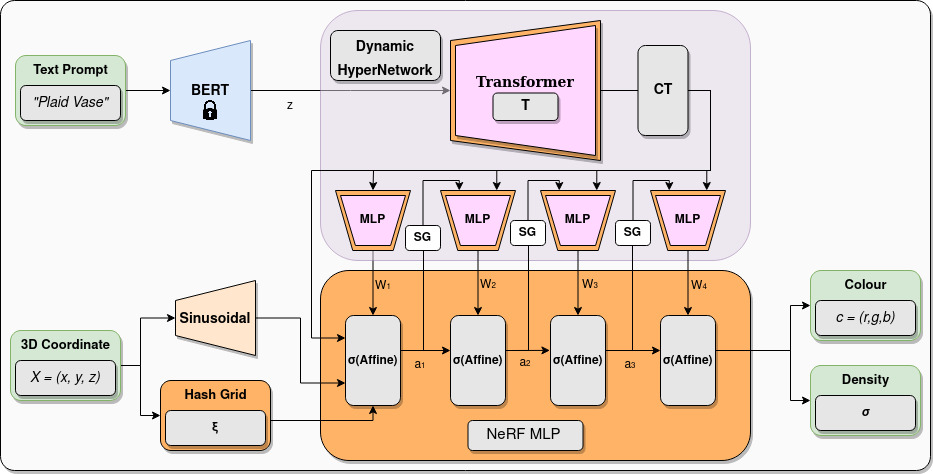}
    \caption{The input to the \ourmethod{} system is a text prompt, which is encoded by a pre-trained text encoder (frozen BERT model). The text latents are passed to a Transformer module, which outputs a conditioning token (CT). This conditioning token (which supplies scene information) is used to condition each of the MLP modules in the hypernetwork. The first hypernetwork MLP (on the left) predicts the weights $W_1$ of the first layer of the NeRF MLP. The second hypernetwork MLP then takes as input both the CT and $a_1$, which are the \textit{activations} from the first predicted NeRF MLP layer, and predicts the weights $W_2$ of the second layer of the NeRF MLP. The subsequent scene-conditioned hypernetwork MLPs follow the same pattern, taking the activations $a_{i-1}$ from the previous predicted NeRF MLP layer as input to generate weights $W_{i}$ for the $i^{th}$ layer of the NeRF MLP. We include stop gradients (SG) so stabilize training}
  \label{fig:high_level}
\end{figure}

\subsection{Dynamic Hypernetwork}
\label{sec:dynamic}
The dynamic hypernetwork consists of the Transformer  $\mathcal{T}$ and MLP modules as given in Figure~\ref{fig:high_level}. The sole input to the dynamic hypernetwork is the scene information represented as a text description. The text is then encoded by a frozen pre-trained BERT \cite{devlin2018bert} model, and the text embedding $z$ is processed by $\mathcal{T}$. Let conditioning token $CT$ = $\mathcal{T}(z)$ be the intermediate representation used to provide the current scene information to the MLP modules. Note that the text embeddings $z$ can come from any text encoder, though in our experiments we found frozen BERT embeddings to be the most performant.\\

In addition to conditioning token $CT$, each MLP module takes in the activations from the previous layer $a_{i-1}$ as input.  Given these two inputs, the MLP module is tasked with generating parameters $W_{i}$ for the $i^{th}$ layer of the NeRF MLP. For simplicity let us assume that we sample only one 3D coordinate and viewing direction per minibatch, and let $h$ be the hidden dimension of the NeRF MLP. Then $a_{i-1} \in \mathbb{R}^{1 \times h}$.  Now the weights $W_{i} \in \mathbb{R}^{h \times h}$  of the $i^{th}$ layer are given as follows: 
\begin{eqnarray}
W_{i} = \text{MLP}_{i}(CT,a_{i-1}) 
\end{eqnarray}
\vspace{-0.1cm}%
The forward pass of the $i^{th}$ layer is:
\begin{eqnarray}
a_{i} = W_{i}*a_{i-1} \label{eqn:fp}
\end{eqnarray}
where $a_{i} \in \mathbb{R}^{1\times h}$ and * is matrix multiplication. This enables the hypernetwork MLPs to generate a different set of weights for the NeRF MLP that are best suited for each given input 3D point and viewing direction pair. This results in effectively a unique NeRF MLP for each 3D point and viewing direction pair.

However training with minibatch size 1 is impractical, so during training we sample a non-trivial minibatch size and generate weights that are best suited for the given minibatch as opposed to the above setting where we generate weights unique to each 3D coordinate and viewing direction pair. 

In order to generate a unique set of weights for a given minibatch we do the following:
\begin{eqnarray}
\overline{a}_{i-1} =& \mu(a_{i-1}) \\
W_{i} =& MLP_{i}(CT,\overline{a}_{i-1})
\end{eqnarray}
Where $\mu$(.) averages over the minibatch index. So if the minibatch size is $n$, then $a_{i-1} \in R^{n\times h}$, and $\overline{a}_{i-1} \in \mathbb{R}^{1\times h}$ and the forward pass is still computed as given in equation \ref{eqn:fp}. This adaptive nature of the predicted NeRF MLP weights leads to the increased flexibility of the model. 
\subsection{NeRF Distillation}
As shown in Figure~\ref{fig:overview}, we first train individual DreamFusion NeRFs on a set of text prompts, following which we train the HyperFields architecture with supervision from these single-scene DreamFusion NeRFs. 

The training routine is outlined in Algorithm~\ref{alg:training}, in which at each iteration, we sample $n$ prompts and a camera viewpoint for each of these text prompts (lines 2 to 4). Subsequently, for the $i^{th}$ prompt and camera viewpoint pair we render image $\mathcal{I}_{i}$ using the $i^{th}$ pre-trained teacher NeRF (line 5). We then condition the HyperFields network $\phi_{hf}$ with the $i^{th}$ prompt, and render the image $I^{'}_{i}$ from the $i^{th}$ camera view point (line 6). We use the image rendered by the pre-trained teacher NeRF as the ground truth supervision to HyperFields (line 7). For the same sampled $n$ prompts and camera viewpoint pairs, let $\mathcal{I}^{'}_{1}$ to $\mathcal{I}^{'}_{n}$ be the images rendered by HyperFields and $\mathcal{I}_{1}$ to $\mathcal{I}_{n}$ be the images rendered by the respective pre-trained teacher NeRFs. The distillation loss is given as follows:
\begin{eqnarray}
\mathcal{L}_{d} = \sum_{i=1}^{n} (\mathcal{I}_{i} - \mathcal{I}^{'}_{i})^{2}
\end{eqnarray}


\begin{algorithm}
\caption{Training HyperFields with NeRF Distillation}
\begin{algorithmic}[1]
\Require $\mathcal{T} = \{\mathcal{T}_{1}, \mathcal{T}_{2}, \cdots, \mathcal{T}_{N} \}$ \Comment{Set of text prompts}
\Require $\mathcal{C}$ \Comment{Set of Camera viewpoints}
\Require $\theta_{1}, \theta_{2}, \cdots, \theta_{N}$ \Comment{Pre-trained NeRFs}
\Require $\phi_{HF}$ \Comment{Randomly initialized HyperFields}
\Require $\mathcal{R}$ \Comment{Differentiable renderer function}
\For{each step}
    \State $\mathcal{T}_{l}, \mathcal{T}_{m}, \mathcal{T}_{n} \sim \mathcal{T}$ \Comment{Sample text prompts from $\mathcal{T}$}
    \For{$\mathcal{T}_{i} \in \{\mathcal{T}_{l}, \mathcal{T}_{m}, \mathcal{T}_{n}\}$}
        \State $\mathcal{C}_{i} \sim \mathcal{C}$
        \State $\mathcal{I}_{i} = \mathcal{R}(\theta_{i}(\mathcal{C}_{i}))$ \Comment{$i^{th}$ NeRF renders image for given camera $\mathcal{C}_{i}$}
        \State $\mathcal{I}_{i}^{'} = \mathcal{R}(\phi_{HF}(\mathcal{T}_{i}, \mathcal{C}_{i}))$ \Comment{Condition $\phi_{HF}$ on $i^{th}$ prompt}
        \State $\mathcal{L}_{i} = (\mathcal{I}_{i} - \mathcal{I}_{i}^{'})^{2}$
    \EndFor
    \State $\mathcal{L}_d = \sum_{i \in \{l,m,n\}} \mathcal{L}_{i}$
\EndFor
\end{algorithmic}
\label{alg:training}
\end{algorithm}

\subsection{Implementation Details}
We use the multiresolution hash grid developed in InstantNGP \cite{instantngp} for its fast inference with low memory overhead, and sinusoidal encodings $\gamma$ to combat the known spectral bias of neural networks \citep{spectralbias}. The NeRF MLP has 6 layers (with weights predicted by the dynamic hypernetwork), with skip connections every two layers. The dynamic hypernetwork MLP modules are two-layer MLPs with ReLU non-linearities and the Transformer module has 6 self-attention layers. Furthermore, we perform adaptive instance normalization before passing the activations into the MLP modules of the dynamic hypernetwork and also put a stop gradient operator on the activations being passed into the MLP modules (as in Figure~\ref{fig:high_level}). The exact dimensions of the various components of the architecture are described in Appendix~\ref{sec:hyperfield_model}.

\section{Results}

We evaluate \ourmethod{} by demonstrating its generalization capabilities, out-of-distribution convergence, and amortization benefits. Further, through ablation experiments we justify the use of dynamic hypernetwork and NeRF distillation. In Section~\ref{sec:generalization} and Section~\ref{sec:gen_ood} we evaluate the model's ability to synthesize novel scenes, both in and out-of-distribution. We quantify the amortization benefits of having this general model compared to optimizing individual NeRFs in Section~\ref{sec:amortization}. 
\begin{figure}[!]
    \centering
    \includegraphics[width=0.8\textwidth]{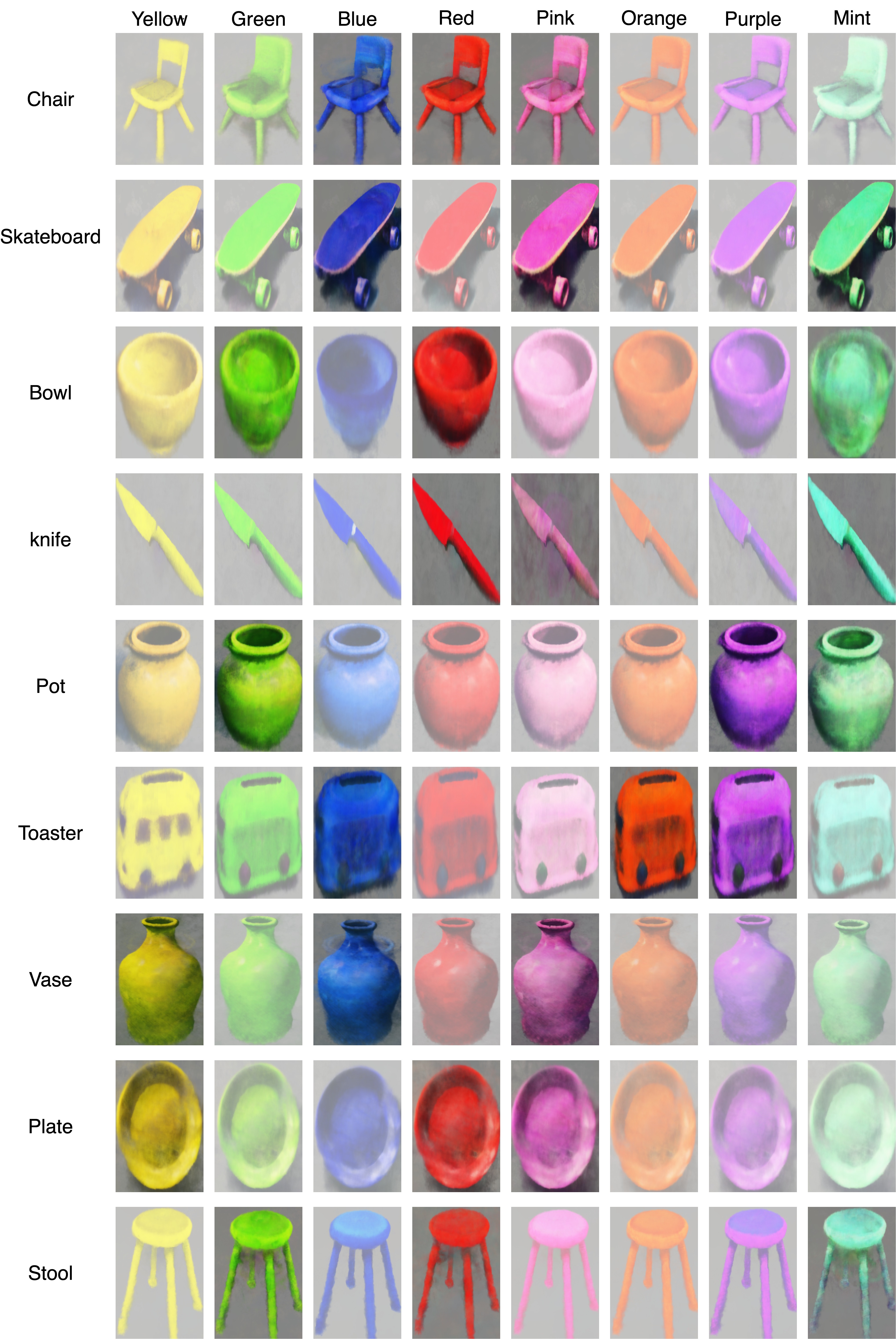}
    \caption{\textbf{Zero-Shot In-Distribution Generalization.} During training, the model observes every individual shape and color, but we hold out a subset of color/shape combinations. During inference, the model generalizes by generating scenes for the held out combinations zero-shot. For example, ``red chair'' is an unseen combination, but the model is able to generalize from individual instances of ``red'' and ``chair'' from training. The faded scenes are generated from the training set, while the bright scenes are zero-shot predictions of the held-out prompts.}
    \label{fig:colormatrix}
\end{figure}

\begin{figure*}[!]
\centering
\includegraphics[width=1.0\textwidth]{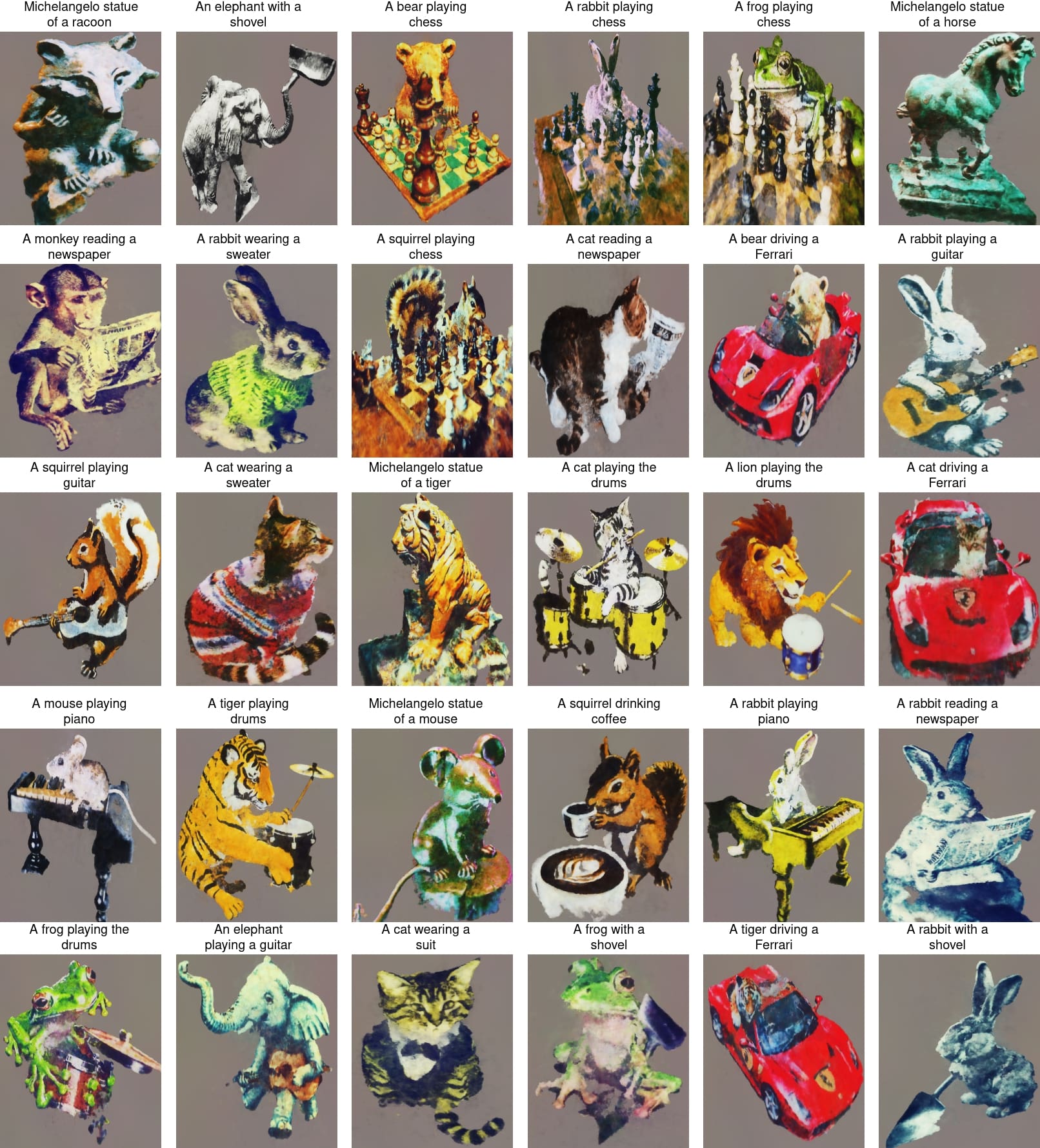}
\caption{\textbf{Prolific Dreamer scenes distilled into HyperFields:} We depict various complex poses of animals, thereby underscoring the ability of a single HyperFields model to learn multiple complex scenes. } 
\label{fig:adddreamerfields}
\end{figure*}

 \begin{figure*}
     \centering
     \includegraphics[width=1.0\textwidth]{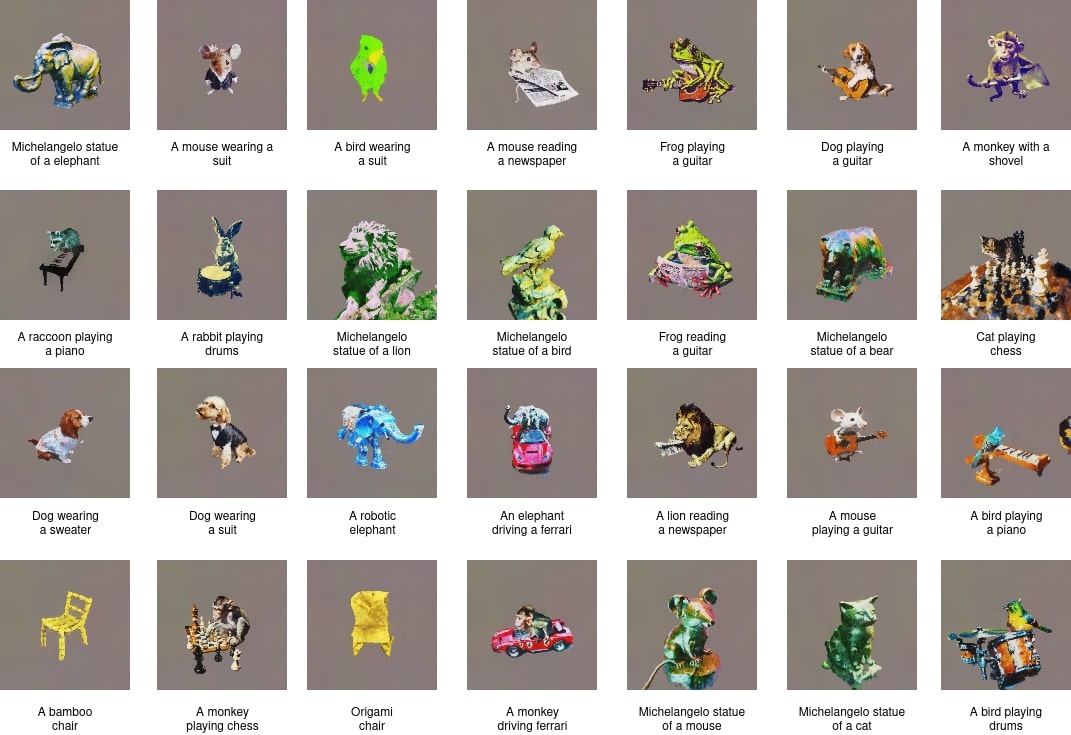}
     \caption{Additional set of Prolific Dreamer scenes distilled into HyperFields model, showcasing the ability of HyperFields to learn a significantly diverse set of geometries.}
     \label{fig:new_dreamer}
 \end{figure*}

\subsection{In-Distribution Generalization}
\label{sec:generalization}
Our method is able to train on a subset of the colour-shape combinations, and during inference predict the unseen colour-shape scenes \emph{zero-shot, without any test time optimization}. Figure~\ref{fig:colormatrix} shows the results of training on a subset of combinations of 9 shapes and 8 colours, while holding out 3 colours for each shape. Our model generates NeRFs in a zero-shot manner for the held-out prompts (opaque scenes in Figure~\ref{fig:colormatrix}) with quality nearly identical to the trained scenes. 

We call this \textit{in-distribution generalization} as both the shape and the color are seen during training but the inference scenes (opaque scenes in Figure~\ref{fig:colormatrix}) are novel because the combination of color and shape is unseen during training. Example: ``Orange toaster" is a prompt the model has not seen during training, though it has seen the color ``orange" and the shape ``toaster" in its training set. 


We quantitatively evaluate the quality of our zero-shot predictions with CLIP retrieval scores. The support set for the retrieval consists of all 72 scenes (27 unseen and 45 seen) shown in Figure~\ref{fig:colormatrix}. In Table ~\ref{tab:clip_ret} we compute the top-$k$ retrieval scores by CLIP similarity. The table reports the average scores for Top-1, 3, 5, 6, and 10 retrieval, separated by unseen (zero-shot) and seen prompts. The similarity in scores between the unseen and seen prompts demonstrates that our model's zero-shot predictions are of similar quality to the training scenes with respect to CLIP similarity.

\begin{table}
  \centering
  \begin{tabular}{lccccc}
    \toprule
    & Top-1 & Top-3 & Top-5 & Top-6 &  Top-10 \\
    \midrule
    Unseen & 57.1 & 85.7 & 85.7 & 90.4 & 95.2 \\
    Seen & 69.5 & 88.1 & 94.9 &94.9 &96.6 \\
    \bottomrule
  \end{tabular}
  \caption{CLIP Retrieval Scores: We report the average retrieval scores for the scenes shown in Figure~\ref{fig:colormatrix}. The small difference in scores between the seen and unseen scene prompts indicates that our zero-shot generations are of similar quality to the training scenes.}
  \label{tab:clip_ret}
\end{table}


\subsection{HyperFields with Proflic Dreamer Teachers}
\label{sec:prolificdreamer}

In addition to the scenes shown in Figure~\ref{fig:adddreamerfields}, we train HyperFields on a different set of ProlificDreamer teachers and the scenes generated by our single HyperFields model are shown in Figure~\ref{fig:new_dreamer}. This demonstrates the ability of HyperFields to learn another varied set of scenes with complex geometries.


NeRF distillation training means that our pipeline is agnostic to the choice of text-to-3D model, and thus can inherit high-quality generation properties from the latest open-source models. We demonstrate this in Figure~\ref{fig:adddreamerfields}, where we generate teacher NeRFs using Prolific Dreamer \cite{wang2023prolificdreamer} and distill them into a single HyperFields model. Our model generates the distilled scenes with virtually no quality degradation. We provide a visual comparison of our generations against the same scenes from ATT3D in Figure~\ref{fig:att3d_compare}.

\begin{figure}[H]
\centering
\includegraphics[width=1\textwidth]{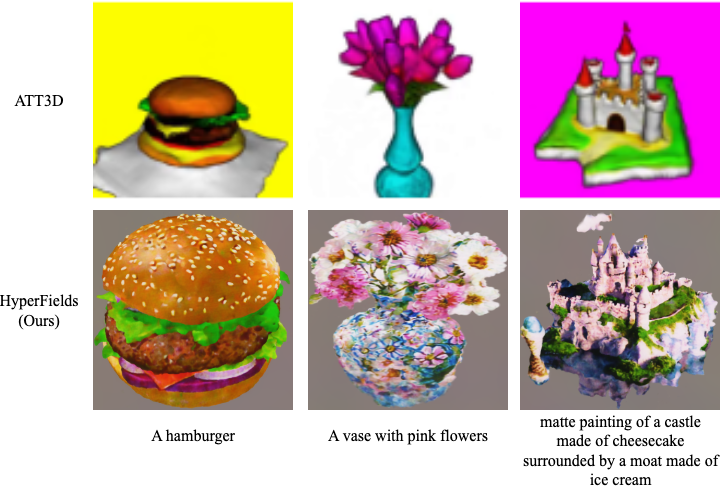}
\caption{\textbf{Visual Comparison to ATT3D.} We visually compare scenes packed into HyperFields against the same scenes shown in ATT3D. NeRF distillation allows \ourmethod{} to inherit the high generation quality of Prolific Dreamer, so the scenes we generate are of higher visual quality and complexity.} 
\label{fig:att3d_compare}
\end{figure}
\subsection{Accelerated Out-of-Distribution Convergence}
\label{sec:gen_ood}
\begin{figure}[!]
    \centering
    \includegraphics[width=1.0\textwidth]{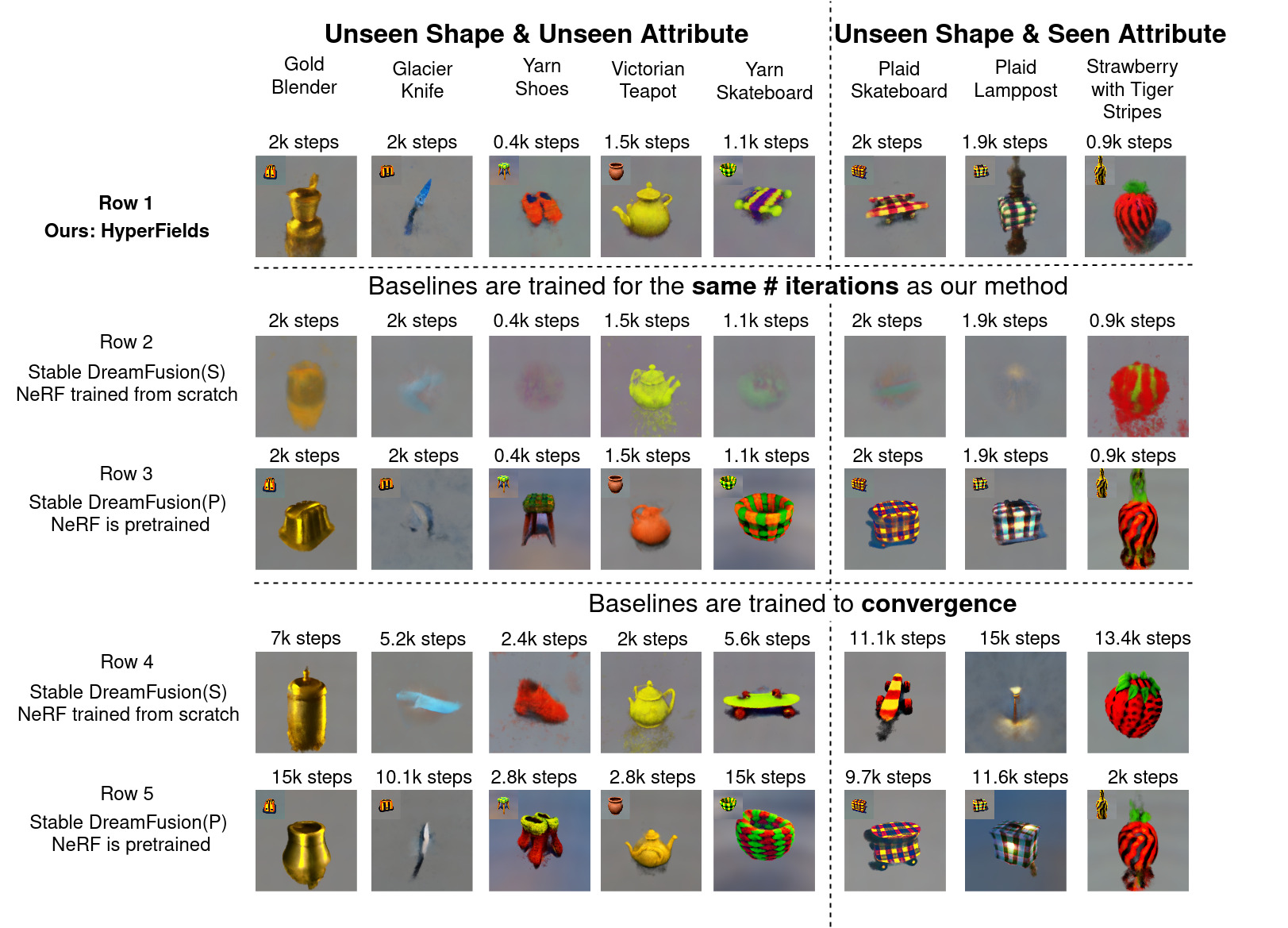}
    \caption{\textbf{Finetuning  to out-of-distribution prompts: unseen shape and or unseen attribute. } Our method generates out-of-distribution scenes in at most 2k finetuning steps (row 1), whereas the baseline models are far from the desired scene at the same number of iterations (rows 2 and 3). When allowed to fine-tune for significantly longer (rows 4 and 5) the baseline generations are at best comparable to our model's generation quality, demonstrating that our model is able to adapt better to out-of-distribution scenes.
}
    \label{fig:ood}
\end{figure}

We further test \ourmethod{}'s ability to generate shapes and attributes that it has \textit{not seen} during training. We call this \textit{out-of-distribution inference} because the specified geometry and/or attribute are not within the model's training set. 

We train our model on a rich source of prompts, across multiple semantic dimensions (material, appearance, shape). The list of prompts used is provided in  \cite{babu2023hyperfields}. Post training, we test our model on the prompts in Figure~\ref{fig:ood}. The prompts are grouped based on whether both shape and attribute are unseen (column 1, Figure~\ref{fig:ood}) or just the shape is unseen (column 2, Figure~\ref{fig:ood}). For example, in ``gold blender" both material ``gold" and shape ``blender" are unseen during training. 

Since these prompts contain geometry/attributes that are unseen during training, we do not expect high quality generation without any optimization. Instead, we demonstrate that fine-tuning the trained \ourmethod{} model on SDS loss for the given the out-of-distribution prompt can lead to accelerated convergence especially when compared to the DreamFusion baselines. 

We consider two baselines, 1) \textbf{Stable Dreamfusion (S):} Publicly  available implementation of Dreamfusion trained from \textbf{S}cratch, 2) \textbf{Stable Dreamfusion (P):} Stable Dreamfusion model \textbf{P}re-trained on a semantically close scene and finetuned to the target scene. The motivation in using Stable Dreamfusion (P) is to have a pre-trained model as a point of comparison against \ourmethod{} model. 

We show out-of-distribution generation results for 8 different scenes in Figure~\ref{fig:ood}. The inset images in the upper left of row 1 of Figure~\ref{fig:ood} are the scenes generated zero-shot by our method, \emph{with no optimization}, when provided with the out-of-distribution prompt. The model chooses the \textit{semantic nearest neighbour} from its training data as the initial guess for out-of-distribution prompts. For example, when asked for a ``golden blender" and ``glacier knife", our model generates a scene with ``tiger striped toaster", which is the only related kitchenware appliance in the model sees during training. We pre-train the Stable Dreamfusion(P) baselines to the same scenes predicted by our model zero-shot. The pre-trained scenes for Stable Dreamfusion(P) are given as insets in the upper left of row 3 and 5 in Figure~\ref{fig:ood}.   

By finetuning on a small number of epochs for each out-of-distribution target scene using score distillation sampling, our method can converge much faster to the target scene than the baseline DreamFusion models. In row 2 and 3 of Figure~\ref{fig:ood}, we see that both Dreamfusion(S) and (P), barely learn the target shape for the same amount of training budget as our method. In rows 4 and 5 of Figure~\ref{fig:ood} we let the baselines train to convergence, despite which the quality of the longer trained baseline scenes are worse or at best comparable to our model's generation quality. On average we see a 5x speedup in convergence.

Importantly, DreamFusion(P) which is pre-trained to \textbf{the same zero-shot predictions of our model} is unable to be fine-tuned to the target scene as efficiently and at times gets stuck in suboptimal local minima close to the initialization (see ``yarn skateboard" row 3 and 5 in Figure~\ref{fig:ood}). This demonstrates that HyperFields learns a semantically meaningful mapping from text to NeRFs that cannot be achieved through single scene optimization.

\subsection{User Study}

\label{sec:study}
In order to get a quantitative evaluation of our generation quality for the out-of-distribution prompts we conduct a human study where we ask participants to rank the render that best adheres to the given prompt in descending order (best render is ranked 1). We compare our method's generation with 33 different DreamFusion models. One is trained from scratch and the other 32 are finetuned from checkpoints corresponding to the prompts used in \cite{babu2023hyperfields}. Of these 33 models we pick the best model for each of the out-of-distribution prompts, so the computational budget to find the best baseline for a given prompt is almost 33x that our of method. Note each of these models, including ours, is trained for the same number of steps. We report average user-reported rank for our method and the average best baseline chosen \textit{for each prompt} in Tab.~\ref{tab:study}. We outrank the best DreamFusion baseline consistently across all our out-of-distribution prompts.  

\begin{table*}[ht]
    \centering
    \scalebox{0.6}{
        \begin{tabular}{lcccccc}
            \toprule
            Model & \centering Golden Blender & Yarn Shoes & Yarn Skateboard & Plaid Skateboard & Plaid Lamppost & Strawberry  with Tiger Stripes \\
            \midrule
            Our Method ($\downarrow$) & 1.30 & 1.07 & 1.32 & 1.29 & 1.36 & 1.12 \\
            Best DreamFusion Baseline ($\downarrow$) & 2.50 & 2.40 & 2.33 & 1.75 & 2.00 & 2.25 \\
            \bottomrule
        \end{tabular}
    }
    \caption{\textbf{Average User-Reported Ranks (N=450):} We report the average rank submitted by all users for our method and compute the average rank for all 33 baselines. We report the average rank of the best-performing baseline for each prompt. Our method is consistently preferred over the best baseline, despite the best baseline consuming 33× more computational resources than our method.}
    \label{tab:study}
\end{table*}

\subsection{Quantitative Evaluation}
We compute additional quantitative results, and report CLIP Precision and Recall (retrieving the 3 nearest-neighbors), along with KID, SSIM scores compared with the Stable DreamFusion baselines trained for the same computational budget for the out-of-distribution scenes from Figure~\ref{fig:ood}. This is provided in Table \ref{tab:model_comparison}

We compute these metrics for Stable DreamFusion using multiple views of the scenes rendered in row 2 of Figure~\ref{fig:ood}. For our model, we compute these metrics using multiple views of the scenes rendered in row 1 of Figure~\ref{fig:ood}. Note we use same set of camera poses to render views across all scenes for both the models. Our model beats the baselines across all these metrics.

\begin{table}[h]
\centering
\begin{tabular}{lcccc}
\hline
\textbf{Model/Metric} & \textbf{CLIP Precision Top-3} & \textbf{CLIP Recall Top-3} & \textbf{KID ↓} & \textbf{SSIM ↑} \\ \hline
Stable DreamFusion & 0.50 & 0.37 & 0.17 & 0.55 \\
Our Model          & 0.77 & 0.63 & 0.13 & 0.62 \\ \hline
\end{tabular}
\caption{\textbf{Quantitative Evaluation.} As suggested by the various metrics our generations are of better quality and adhere better to the given text prompt. }
\label{tab:model_comparison}
\end{table}


\begin{figure}[ht]
    \centering
    \includegraphics[width=1\textwidth]{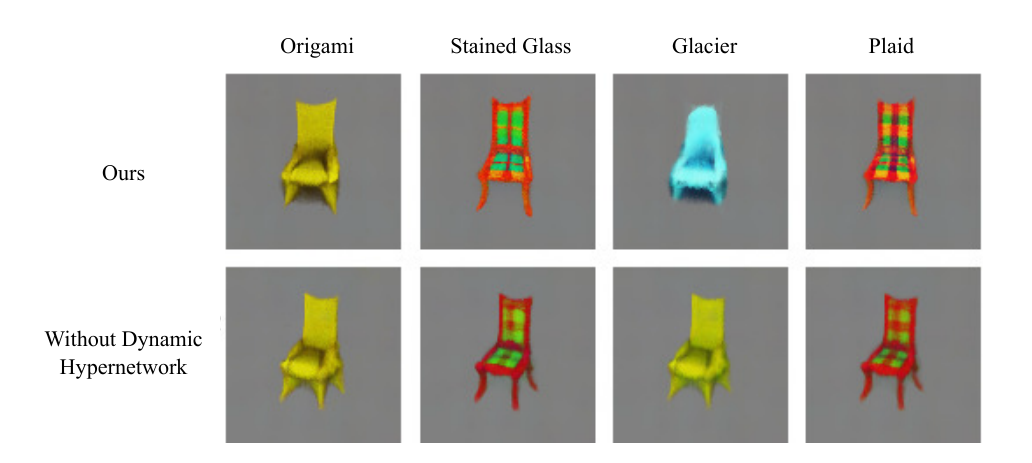}
    \caption{\textbf{Dynamic Hypernet Packing.} Without dynamic conditioning, the network collapses the origami/glacier attributes and stained glass/plaid attributes.}
    \label{abl:packing}
\end{figure}

\begin{figure}[ht]
    \centering
    \includegraphics[width=0.70\textwidth]{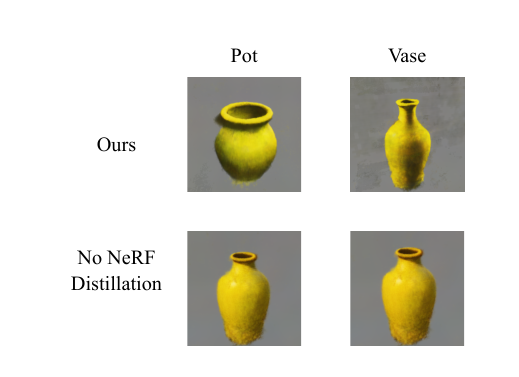}
    \caption{\textbf{NeRF Distillation.} We compare packing results when training with score distillation (``No NeRF Distillation") versus our NeRF distillation method (``Ours"). The iterative optimization of score distillation causes mode collapse in geometry.}
    \label{abl:distillation}
\end{figure}
\subsection{Ablations}
\label{sec:ablations}


We ablate on the activation conditioning in our dynamic hypernetwork (``without dynamic hypernetwork'') in Figure~\ref{abl:packing}. Row 2 shows that even in the simple case of 4 scenes the static hypernetwork collapses the ``glacier" and ``origami" styles, and the ``plaid" and ``stained glass" styles.

If we attempt to pack the dynamic hypernetwork using just Score Distillation Sampling (SDS) from DreamFusion, we experience a type of mode collapse in which the SDS optimization guides similar shapes towards the same common geometry. See Figure~\ref{abl:distillation} for an example of this mode collapse.

\subsection{Amortization Benefits} 

\label{sec:amortization}
The cost of pre-training \ourmethod{} and individual teacher NeRFs is easily amortized in both in-distribution and out-of-distribution prompts. Training the teacher NeRFs is not an additional overhead; it's the cost of training a DreamFusion model on each of those prompts. The only overhead incurred by our method is the NeRF distillation training in stage 2 (Figure~\ref{fig:overview}), which takes roughly two hours. This overhead is offset by our ability to generate unseen combinations in a feed-forward manner.

For comparison, the DreamFusion baseline takes approximately 30 minutes to generate each test scene in Figure~\ref{fig:colormatrix}, totaling $\sim$14 hours for all 27 test scenes. In contrast, after the 2 hour distillation period, our model can generate all 27 test scenes in less than a minute, making it an order of magnitude faster than DreamFusion, even with the distillation overhead. 

Our method's ability to converge faster to new out-of-distribution prompts leads to linear time-saving for each new prompt. This implies a practical use case of our model for rapid out-of-distribution scene generation in a real world setting. As shown in Figure~\ref{fig:ood}, the baseline's quality only begins to match ours after 3-5x the amount of training time.

\subsection{Extension: Learning Feature Fields via HyperFields}
3D segmentation is of significant interest to both the vision and graphics community \cite{qi2017pointnet, qi2017pointnet++,zhao2021point,cciccek20163d,huang2021supervoxel,zhu2023less}. It is crucial for a variety of applications such as autonomous driving, medical diagnosis, robotics, augmented reality \cite{sanchez2023domain,alalwan2021efficient,guerry2017snapnet,tchapmi2017segcloud,tanzi2021real}. 

Recently, with the advent of neural implicit models capable of storing radiance fields of various complex scenes \cite{tancik2022block,barron2021mip,mildenhall2021nerf,barron2022mip,chabra2020deep}, several works have demonstrated that semantic fields corresponding to 3D scenes can be learnt \cite{zhang2024open,kerr2023lerf,kobayashi2022decomposing,siddiqui2023panoptic}. This is achieved by distilling the knowledge of a foundational model into the neural implicit model. 

These distilled feature fields often provide view consistent semantic features leading to better 3D view consistent segmentation when compared to even the foundational models that are used for distillation. While this is impressive, the distillation process is time consuming and is repeated each time a new scene is required to be segmented. This is undesirable, as this leads to wastage of time and compute. 

We seek to circumvent this problem by learning an efficient prior via a hypernetwork. We extend the HyperFields architecture to generate NeRF MLPs that also generate a feature field along with a radiance field. By training a hypernetwork over multiple scenes we acquire a prior that enables generation of parameters of the NeRF MLP in a forward pass (or minimal fine-tuning), when the hypernetwork is presented with a few views of a novel scene. This thereby enables synthesizing novel 3D views of a given scene along with a feature field which can be used to perform semantic segmentation. 

\subsubsection{ShapeNet Results}
We present the results of our proposed HyperSegmentationFields model on the ShapeNet dataset, a widely used benchmark for 3D object recognition and segmentation tasks \cite{chang2015shapenetinformationrich3dmodel}. ShapeNet provides a diverse collection of 3D models across multiple categories, making it an ideal testbed for evaluating the effectiveness of our approach in generating semantic feature fields.

Figure~\ref{fig:hyper_seg_arch_res} showcases the capability of our model to produce high-quality feature fields for various 3D scenes using only a handful of training views. The results demonstrate that our hypernetwork can efficiently generate NeRF MLP parameters (in a single forward pass) that accurately reconstruct both the geometry and the semantic features of the scenes. The generated feature fields are consistent across different views, highlighting the model's ability to capture the intricate details and semantics of the 3D scenes.

\begin{figure}[H]
    \centering
    \includegraphics[width=1.0\textwidth]{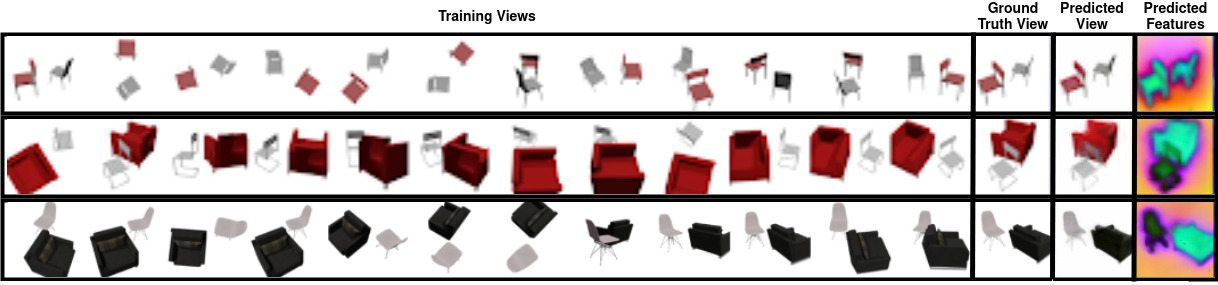}
    \caption{\textbf{HyperSegmentationFields}: Feedforward generation of feature field for various scenes when conditioned on a handful of training views.  }
    \label{fig:hyper_seg_arch_res}
\end{figure}
\section{Discussion}
We present \ourmethod{}, a novel framework for generalized text-to-NeRF synthesis, which can produce individual NeRF networks in a single feed-forward pass. Our results highlight a promising step in learning a general representation of semantic scenes.  Our novel dynamic hypernetwork architecture coupled with NeRF distillation learns an efficient mapping of text token inputs into a smooth and semantically meaningful NeRF latent space. Our experiments show that with this architecture we are able to fit over 100 different scenes in one model, and predict high quality unseen NeRFs either zero-shot or with a few finetuning steps. Comparing to existing work, our ability to train on multiple scenes greatly accelerates convergence of novel scenes.  In future work we would like to explore the possibility of generalizing the training and architecture to achieving zero-shot open vocabulary synthesis of NeRFs and other implicit 3D representations. 
\clearpage

\chapter{Learning Geometry Aware Molecule Representations via Generation}
\section{Introduction}
Representation learning is foundational in providing significant improvements in various downstream tasks spanning multiple modalities such as vision, text, and speech \cite{ dalle2,alayrac2022flamingo,zhang2022dino, lu2019vilbert,radford2021learning,tan2019lxmert,radford2022robust}. These representation learning approaches involve training models on pretext tasks, which enables them to learn rich and useful representations of data without relying on explicit manual annotations. In vision, tasks such as contrastive learning, image colorization, patch position prediction, and image inpainting help models grasp spatial and contextual nuances \cite{chen2020simple, zhang2016colorful, larsson2017colorization,doersch2015unsupervised,he2022masked}. Textual understanding is deepened through pretext tasks like masked language modeling, next sentence prediction, and sentence ordering \cite{devlin2018bert,radford2018improving}. For speech, audio frame prediction and contrastive predictive coding are instrumental in teaching models the subtleties of acoustic patterns and temporal dynamics \cite{oord2018representation,9414567,liu2022audio}. 
{\par}
Given the transformative impact of representation learning across vision, text, and speech domains, there's a compelling opportunity to extend its benefits to computational drug design. Central to many challenges in drug design is the accurate modeling of molecular structures and consequently their properties. This underscores the critical need for developing representation learning models tailored for learning molecule representations. Thereby enabling prediction of wide range of molecular attributes essential for enhancing the efficiency and effectiveness of identifying and optimizing novel drug candidates. 

Building on the remarkable achievements of generative models as potent representation learning systems \cite{bhattad2024stylegan,zhan2023does,Yang_2023_ICCV, radford2018improving,zhang2023deciphering,zhang2023structural}, we repurpose a diffusion-based \cite{ho2020denoising, song2020denoising, dhariwal2021diffusion} framework for molecule generation to serve as an effective system for learning molecular representations. By leveraging generative models, we aim to learn geometry-aware molecular representations that are effective for downstream tasks, even when training data is limited. 

Beyond their ability to generate diverse and high-fidelity molecular structures, diffusion models possess inherent denoising capabilities that enable the extraction of robust and informative representations \cite{zhang2023structural, zhang2023deciphering,mittal2023diffusion}. This makes them particularly well-suited for low-data regimes, where learning stable and generalizable features from noisy limited samples is critical.

This approach holds significant practical value in drug discovery, where data scarcity often limits the effectiveness of predictive models. By leveraging diffusion-based representations, we enhance the ability to identify potential drug-target interactions, as demonstrated in our results, where diffusion-derived features improve binding prediction performance. More broadly, the ability to extract robust molecular features from generative models can accelerate hit discovery, ultimately improving the efficiency of drug development pipelines.

\section{Related Works}

In the field of molecule generation, the primary objective is to generate a diverse set of novel, stable, and valid molecules that possess desirable properties, such as low toxicity. Recent advancements have explored various approaches to achieve this goal, particularly through the use of equivariant models and diffusion processes.

Hoogeboom et al.~\cite{hoogeboom2022equivariant} propose a novel approach that leverages equivariant Graph Neural Networks (GNNs) in conjunction with a modified SE(3) equivariant diffusion process. The method ensures SE(3) equivariance by making the center of mass of the molecule zero at each diffusion step, using isotropic Gaussian noise to guarantee rotational equivariance, and relying on an equivariant GNN to maintain the overall consistency of the diffusion process. Xu et al.~\cite{xu2023geometric} build upon this by implementing the diffusion process in latent space while keeping the model architecture largely the same.

Xu et al.~\cite{xu2024geometric} address a significant limitation in previous models that only generate pairwise distances and rely on heuristics for bond prediction. Their model introduces the use of additional triplet bond angle information to better predict bond nature and incorporates a loss function that enforces atomic valency constraints during training. This refinement biases the generated molecules toward greater stability. The model architecture integrates self-attention layers for processing rotation-invariant features, while equivariant GNNs are used to model coordinate features. By updating atomic, pairwise, and triplet features collectively and refining geometry information at each layer, this approach aims to produce more accurate and stable molecular structures during the diffusion process.

\section{Methods}

\subsection{Diffusion Model as a Feature Extractor}
\begin{figure}[htbp]
    \centering
    \includegraphics[width=1\textwidth]{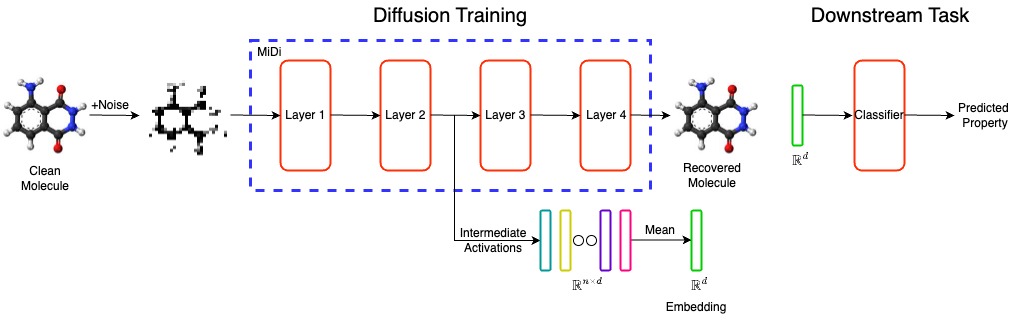} 
    \caption{\textbf{Feature Extraction Pipeline}: A noised molecule is passed through a diffusion model and intermediate features are extracted as activations for downstream tasks.}
    \label{fig:midi_extract}
\end{figure}

We leverage a pre-trained diffusion model MiDi, as a feature extractor to obtain geometry-aware molecular representations \cite{vignac2023midi}. Our approach involves processing molecular conformers with the diffusion model at a fixed noise level and extracting representations from intermediate activations. These extracted features are then utilized for downstream prediction tasks via a multi-layer perceptron (MLP) classifier. A schematic overview of this process is depicted in Figure~\ref{fig:midi_extract}. We follow MiDi's input representation and noise injection and they are explained below.

\subsection{Input Representation and Noise Injection}
A molecule is represented by atom coordinates $ X \in \mathbb{R}^{n \times 3} $, where $n$ is the number of atoms, and atom types which is a discrete $C \in \mathbb{R}^c$ matrix, where $c$ is the number of atom types. 
Now to obtain molecular embeddings, we pass molecular conformers through the pre-trained diffusion model after corrupting at a low noise level. We follow the same noising process as in MiDi, the noising process varies depending on if its \textit{continuous} atom coordinates or \textit{discrete} atom types. 

For \textit{continous} coordinates $X$, we add \textit{zero center-of-mass Gaussian noise} at step $t$:
\begin{eqnarray}
X_t &=& \alpha_t X_{t-1} + \sigma_t\, \epsilon,\\
\epsilon &\sim& \mathcal{N}_{\text{CoM}}(0, I),
\end{eqnarray}
where $\alpha_t$ controls how much of the signal is preserved and $\sigma_t$ controls the magnitude of noise at time step $t$, further, $\sum_i \epsilon_i = 0$ enforces a zero center-of-mass constraint. For \textit{discrete} atom types $C$, we employ a \textit{categorical diffusion} process:
\begin{eqnarray}
C_t &\sim& \mathrm{Categorical}\bigl(C_{t-1} Q^t\bigr),
\end{eqnarray}
where the transition matrix $Q^t$ is defined as:
\begin{eqnarray}
Q^t &=& \alpha_t I + \beta_t (1 a)\, m',
\end{eqnarray}
ensuring realistic transitions between atom types based on an empirical prior $m'$ and $\beta_{t}$ controls the atom type transition rate. The resulting noised inputs $(X_t, C_t)$ are then fed into the diffusion model to extract geometry- and type-aware features for downstream tasks.

\subsection{Feature Extraction}

As illustrated in Figure~\ref{fig:midi_extract}, we obtain an intermediate-layer feature matrix 
$
F \in \mathbb{R}^{n \times d}
$
from the diffusion model, where \(n\) is the number of atoms and \(d\) is the feature dimension.
To derive a single molecular embedding, we aggregate \(F\) over the atom dimension by taking the mean,
thereby preserving overall molecular context while reducing the influence of atom-specific noise.
The final representation thus resides in 
$
\mathbb{R}^{d}
$.
\subsection{Downstream Task Prediction}
The aggregated feature representation  serves as input to an MLP, which is trained for downstream prediction tasks. The MLP consists of two fully connected layers with non-linear activations and a final task-specific output layer. Given an input feature representation, the MLP is trained to predict molecular property or classification label. The MLP is trained using a standard supervised loss, such as mean squared error (MSE).

By leveraging the generative prior of diffusion models, this method enables learning of robust molecular representations suitable for various downstream tasks, even in data-scarce scenarios.

\section{Results}
We evaluate the effectiveness of diffusion-based molecular representations for a drug discovery task. In particular, the goal is to identify candidate drugs that bind effectively with a target protein (Claudine). The binding accuracy of our predictions is tested experimentally in the wet lab. Due to proprietary constraints we do not disclose the names of the drugs. We have a few hundred experimentally verified binding affinities which we fine-tune the classifier head on. We measure model performance using the Area Under the Receiver Operating Characteristic Curve (AUROC).
\begin{table}[t]
    \centering
    \renewcommand{\arraystretch}{1.2} 
    \setlength{\tabcolsep}{12pt} 
    \begin{tabular}{l c}
        \toprule
        \textbf{Model} & \textbf{ AUROC} \\
        \midrule
        ChemBERTa& 0.850 \\
        ChemBERTa + Midi Diffusion features & \textbf{0.877} \\
        \bottomrule
    \end{tabular}
    \caption{AUROC scores for different molecular representation models. Augmenting ChemBERTa with diffusion-based features (Midi) achieves the best performance.}
    \label{tab:auroc_results}
\end{table}
We use the ChemBERTa model as a baseline, which is trained on 77 million unique SMILES sequences. In contrast, the MiDi diffusion model is trained on approximately 450 thousand unique molecular structures \cite{chithrananda2020chembertalargescaleselfsupervisedpretraining, vignac2023midi}. Despite the disparity in number of training samples, ChemBERTa when augmented with MiDi features improves the performance significantly, marking a 2.7 absolute improvement, as shown in Table~\ref{tab:auroc_results}. This improvement underscores the value of geometry-aware molecular representations obtained from the diffusion model. 

We hypothesize that the enhanced performance can be attributed to the following factors. The MiDi diffusion model effectively captures spatial relationships between atoms, which are crucial for modeling molecular interactions such as drug binding. Additionally, the model's inherent denoising capability, used as a pretext task, produces robust embeddings, improving generalization to new molecules. Moreover, although not quantitatively presented, wet lab collaborators observed that the model augmented with MiDi features selected a more diverse set of candidate molecules for experimental testing. This suggests that the diffusion-derived features complement ChemBERTa’s text-based molecular embeddings (e.g., SMILES) with rich geometric context, resulting in a broader and more chemically varied selection of molecules as effective binders. 

These results demonstrate that integrating diffusion-based geometry-aware features with large-scale language models like ChemBERTa can enhance the identification of promising drug candidates, even in low-data regimes, providing valuable guidance for experimental validation in drug discovery pipelines.

\chapter{Future Work: Applying Meta-Learning Tools to Computational Immunology}
In this section we discuss potential application of tools developed in this thesis to overcome data scarcity and thereby deepen our understanding of critical and yet unresolved issues in immunology.

\section{Modeling of T-cell-Antigen Interactions}
 T cells are pivotal to immune defense, with the T-cell receptor (TCR) serving as a unique biological ‘lock and key,’ recognizing and binding to specific antigens, which can be anything from pathogen fragments to mutated proteins in cancer cells. These interactions trigger the immune response. Nonetheless, predicting TCR-antigen interactions remains difficult, due to both the immense variety of T cells, each with its distinct TCR, and the absence of extensive datasets cataloging the intricacies of these interactions \cite{davis1988t}. This data scarcity impedes our understanding of how the immune system recognizes a vast array of antigens and, consequently, the development of targeted immunotherapies. Addressing these challenges of data scarcity and ensuring efficient transfer learning in domains with limited pre-training data is imperative.

 \textbf{Potential of Few-Shot Learning.}
Few-shot learning methods offer a promising approach to making more generalizable predictions about TCR-antigen interactions from limited data. Few-shot learning aims to learn from a few examples and extrapolate to new, unseen data. This technique is particularly well-suited for TCR profiling, where it's impractical to collect extensive data for every TCR-antigen interaction. By analyzing small datasets, networks can identify and apply fundamental patterns to predict new TCR interactions with various antigens, despite the immune system's complexity and variability.

\textbf{Hypernetworks for TCR-Specific Models.}
In this thesis, we develop hypernetworks as a strategy for transferring knowledge to new predictive conditions. Techniques such as yeast display and single-cell dextramer sorting are enabling the generation of large TCR-specific datasets against a wide array of antigens or antigen-specific data against an array of TCRs~\cite{zhang2018high,adams2016structural}. 
This leads us to explore whether hypernetworks when trained on exisisting TCRs or antigens can robustly generate MLPs for novel TCRs or antigens with just limited amount of data. The representational learning of TCRs and antigens through hypernetworks could offer a more adaptable and effective framework. Unlike traditional models that pool data—often resulting in uneven distribution and biased learning due to variability across antigens and TCRs—hypernetworks offer a solution to learn TCR-specific models of antigen binding, and quickly adapt to new TCRs. This approach could provide a more efficient and inclusive solution that accommodates the inherent heterogeneity of immunological data.

\section{Feature Fields for Modelling Protein-Ligand Binding}
Efficient modeling of protein-ligand binding is central to many drug discovery pipelines. Due to the limited availability of protein-ligand interaction data and the localized nature of binding, we propose using HyperFields to learn a feature field on the ligand/drug surface. This approach allows for the capture of local features crucial for modeling localized binding interactions. Additionally, hypernetworks facilitate the efficient transfer of priors across different molecules while enabling the learning of a feature field for novel ligands.
\chapter{Meta-Learning Beyond Adaptation: Towards Pre-Training Strategies}

Our findings demonstrate that hypernetworks and other meta-learning approaches significantly improve generalization, particularly in out-of-distribution (OOD) test scenarios. This suggests a compelling direction for rethinking how foundational models are trained. Rather than relying solely on traditional large-scale pre-training followed by fine-tuning, we propose that integrating hypernetworks or meta-learning techniques into pre-training could lead to models that generalize better and adapt more effectively to unseen conditions, which is crucial in fields like computational immunology and chemistry. In such domains, acquiring large-scale annotated datasets is often expensive and slow, meaning that models frequently encounter OOD scenarios at inference time—for instance, novel proteins or molecular scaffolds in bioinformatics, or unseen cell states in single-cell analysis \citep{stark2022equibind}. Similarly, in computer vision, foundational models may face generalization challenges when deployed on new modalities such as medical imaging (e.g., X-rays, MRI, CT) \citep{wang2023medfmc} or emerging techniques like cryo-electron tomography (Cryo-ET), where data characteristics diverge significantly from natural images \citep{zeng2021survey}. These challenges underscore the need for pre-training strategies that are inherently robust to distribution shifts.

One promising approach could be episodic pre-training, where each mini-batch is structured as a small dataset containing both support and query samples. Within each mini-batch, the model is trained on the support set, and the loss is computed on the query set, thereby baking generalization directly into the optimization process. By explicitly structuring the learning process around diverse mini-batch distributions, episodic training encourages models to rapidly adapt to novel distributions—a capability that traditional optimization strategies often lack. Techniques such as MAML and memory-based meta-learning exemplify how this approach can enhance a model’s ability to generalize beyond its training distribution and could serve as strong foundations for designing meta-learning-based pre-training routines.

Another potential direction is to extend the HyperFields framework (Chapter \ref{chap:hf}) beyond its current scope in 3D vision to broader 3D object domains, such as molecular structures. Hypernetworks, which generate neural implicit representations, are well-suited for processing any type of 3D information. When trained at scale on large 3D datasets, such models could exhibit even better OOD generalization compared to the results presented in Chapter \ref{chap:hf}, where training was limited to relatively small datasets. With large-scale training, hypernetworks could emerge as a default architecture for processing diverse range of 3D objects by generating neural implicits, potentially leading to a general 3D foundational model. This approach holds the potential to produce richer feature representations, stronger OOD generalization capabilities, and more robust predictions for 3D structures—all of which are critical for applications spanning from molecular modeling to scene reconstruction.

As foundational models take center stage in a wide range of applications, it becomes necessary to imbue them with ability to generalize better to OOD tasks. This raises the question: should the principles of meta-learning be applied not just for adaptation, but as a core strategy for pre-training itself?

\chapter{Conclusion}
In this thesis, we have explored the potential of meta-learning architectures, particularly memory-augmented networks and hypernetworks, to enhance the efficiency and adaptability of neural networks in scenarios where traditional transfer learning approaches fall short. Our contributions span multiple modalities, from 2D and 3D vision tasks to molecular representation learning, demonstrating the broad applicability and robustness of our proposed methods.

Our investigation into memory-based meta-learning reveals that distributing memory cells across network layers significantly improves task adaptation, particularly in challenging online learning scenarios. This approach offers a compelling alternative to existing architectures by simplifying the adaptation process and allowing for more flexible and efficient learning across tasks.

In the realm of hypernetworks, we have shown that these networks can effectively acquire and transfer priors across multiple tasks, enabling rapid adaptation with limited data. Our enhancements to the design and training of hypernetworks, including the use of MAML and progressive, dynamic weight prediction, have proven effective in overcoming the challenges associated with training hypernetworks in complex settings. The application of hypernetworks to text-to-3D tasks, and feed-forward generation of feature fields as demonstrated by our HyperFields model, underscores their potential to impact fields requiring fast, efficient generation of high-quality 3D outputs.

Furthermore, repurposing a diffusion-based framework for molecule generation as a proxy for representation learning highlights the promise of extending generative models as representation learning methods for computational drug design. 

Looking forward, the methods and architectures developed in this thesis hold significant promise for addressing pressing challenges in domains characterized by data scarcity and the need for rapid adaptation. The potential applications in immunology, particularly in modeling T-cell-antigen interactions and protein-ligand binding, suggest that our contributions could pave the way for more efficient and effective solutions in critical data scarce fields such as computational chemistry and immunology.



\bibliographystyle{plain}
\bibliography{egbib}
\appendix

\chapter{Online Adaptation via Distributed Neural Memory}

\section{Details of Online Few-Shot Learning Experiments}
\label{app:of}
\subsection{Datasets}
\textbf{CIFAR-FS:} \cite{bertinetto2018meta} adapt the CIFAR-100 dataset for few-shot learning by performing a class-wise partition, yielding a training, validation, and testing set composed of 60, 16, and 20 classes, respectively. We sample 10-shot 5-way online few-shot tasks by sampling 5 classes and 10 samples per class (a total of 50 images), which are fed to the model at a rate of one image per time step. We report validation performance for all models.

\textbf{Omniglot:} This dataset has 1623 characters from 50 different alphabets, with 20 images per character \cite{lake2015human}. Following \cite{vinyals2016matching}, we resize images to 28 $\times$ 28 and augment the dataset by rotating each character by multiples of 90 degrees. We use the same split as \cite{protorepo}. We report validation performance for all models.

\subsection{Task Details :}

Figure~\ref{fig:omni-task} shows a sample online 10-shot 5-way task. As mentioned in Section~\ref{sec:task}, at time step $i$, we pass the $i_{th}$ sample and the label associated with the sample at the previous time step $i-1$. Further, at each time step the model predicts the label for the given sample; the model's efficacy is the accuracy of the predicted labels. Subsequently, in the next time step when the correct label is presented, we expect the model to update itself in order to achieve better generalization for future examples of the given task.

\begin{figure*}[ht]
\centering
\includegraphics[scale=0.3]{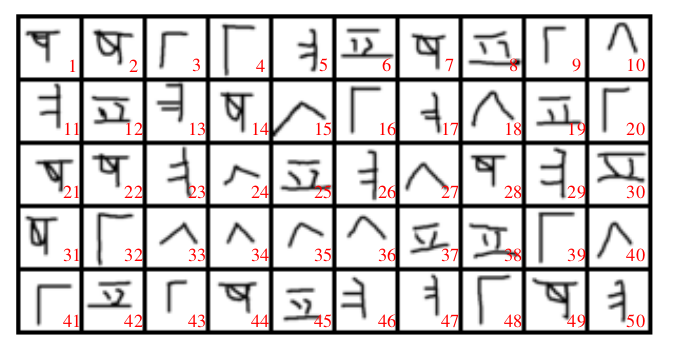}
 \caption{\textbf{A sample Omniglot-based online 10-shot 5-way task.} Each class consists of 10 examples and there are 5 classes in total.  The red number at the bottom right of each character is the time-step at which that character is presented to the model. }
 \label{fig:omni-task}
\end{figure*}

\subsection{Computing k-shot 5-way accuracy:}
We follow the same protocol as in \cite{santoro2016one}.  k-shot results are calculated based on the prediction accuracy of ${k^{th}+1}^{th}$ instance of the class– that is, when the model sees an image of the class for the $k$ times.
\subsection{Architecture Details}
The LSTM baseline and the controller in NTM have a hidden size of 400, followed by a 5-way classifier. The non-parametric memory in NTM is a matrix in $\mathbb{R}^{120 \times 40}$, with 0.95 as the decay rate. In order to make comparisons fair, we use a 4-layer ConvLSTM(CL) with a 64 channels in each layer, followed by a LSTM of hidden size 400 and a 5-way classifier.

For both OPN and CPM, we use the same features extractor as in \cite{ren2020wandering}: a 4-layer CNN with 64 channels in each layer. The CPM has, in addition to the feature extractor, an LSTM of hidden size 400, followed by a gating a mechanism for updating prototypes. We adopt the same feature extractor when training APL, along with a LSTM decoder with hidden size 400 and a total of 5 neighbors read from the memory for each sample. Note: In the APL paper they use a ResNet as the feature extractor.

All models are trained with a learning rate of $10^{-3}$ using Adam \cite{kingma2014adam} for 100k iterations with a meta batch size of 16. 

\section{Task Length Dependent Optimization Issues}
\label{app:opt}
In initial experiments, we observe that for tasks with 50 time steps these models did not train well. We hypothesize that this could be due to the same network being repeated 50 times, thereby inducing an effective depth of the network to be 6 $\times$ 50 (since our model has 6 layers). We resolve training difficulties by adding a skip connection, in the style of residual networks \cite{he2015deep}, between the second layer and the fourth layer (omitted in Figure~\ref{fig:method}). This makes a significant impact in convergence and generalization.

From Table~\ref{tab:effective_learn}, we see that regardless of actual depth, effective depth seems to be critical in determining if the model can learn an efficient task adaptation strategy. In these experiments, we use a task protocol that is slightly different from the protocol in Section~\ref{sec:task}. Here, we have a separate support and query set, and the model is evaluated only on its performance on the query set. The support set is presented to the model in an online manner, one character at a time step. For models with depth 4, we have the following architecture: 2 ConvLSTM layers followed by a LSTM and a classifier. For models with depth 6, we use 4 ConvLSTM layers followed by a LSTM and a classifier. As mentioned in the paper, we ameliorate training issues by adding a skip connection to these models.

\begin{table}[ht]
\centering
\begin{tabular}[t]{ccccc}
\toprule
Network Depth & Task & Task Length & Effective Depth & Efficient Task Adaptation  \\
\midrule
4 & 5-shot 6-way &  30 & 120 & \checkmark \\
4 & 5-shot 7-way &  35 & 140 & \xmark \\
6 & 5-shot 4-way &  20 & 120 & \checkmark \\
6 & 5-shot 5-way &  25 & 150 & \xmark \\
\bottomrule
\end{tabular}

\caption{\textbf{Effect of task length on learning efficient task adaptation strategies.} Effective depth = network depth $\times$ task length, seems to determine if the network can learn efficient adaptation rules. If effective depth is greater than 140, the model fails to adapt efficiently to a given task.}
\label{tab:effective_learn}
\end{table}

\section{Details of Continual Learning Experiments}
\label{app:ocl}
\subsection{Training}
 We always start by exposing the model to a 5-way continual learning problem, after which for every 5K updates we increase the length of the tasks exposed to the model by 1 till the exposed task length is equal to the desired task length. Example: Suppose 7-way 5-shot is the desired task we want our model to solve, we start with a 5-way task for the first 5K updates, then 6-way for the next 5K updates, and then finally we train the model on 7-way tasks until convergence.  We train CL+LSTM with a meta-batch of 16 using Adam and a learning rate of $10^{-3}$.
\subsection{Details of OML Experiments}
We use publicly released code\footnote{\url{https://github.com/khurramjaved96/mrcl}} with the suggested hyperparameters (Adam/SGD with learning rates $10^{-4}$/$0.03$ for meta-training/adaptation), while adopting a 4-layer CNN with a parameter count comparable to ours. Unlike \cite{javed2019meta}, we use $28 \times 28$ images instead of $84 \times 84$ and train for a total of 150k episodes. We use SGD for adaptation instead of Adam in meta-testing, which resulted in significantly better results for our setting where trajectories are short, likely due to the lack of momentum.

\begin{figure}[ht]
\centering
\includegraphics[width=\linewidth]{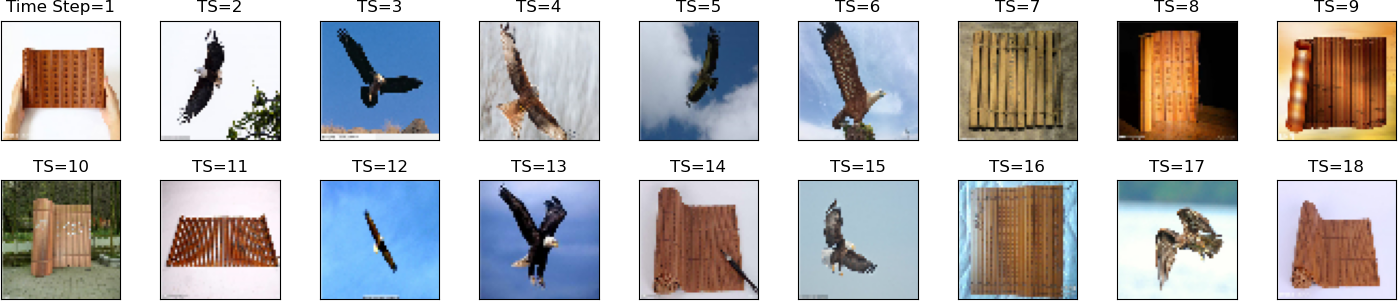}
\caption{ \textbf{A sample online segmentation task with distractors:} A sequence of parchment and eagle images are presented to the model (one image at a time step). Here parchments are the distractor images and the eagles are the images to be segmented.}
\label{fig:semsega}
\end{figure}
\begin{figure}[t]
\centering
\includegraphics[width=\linewidth]{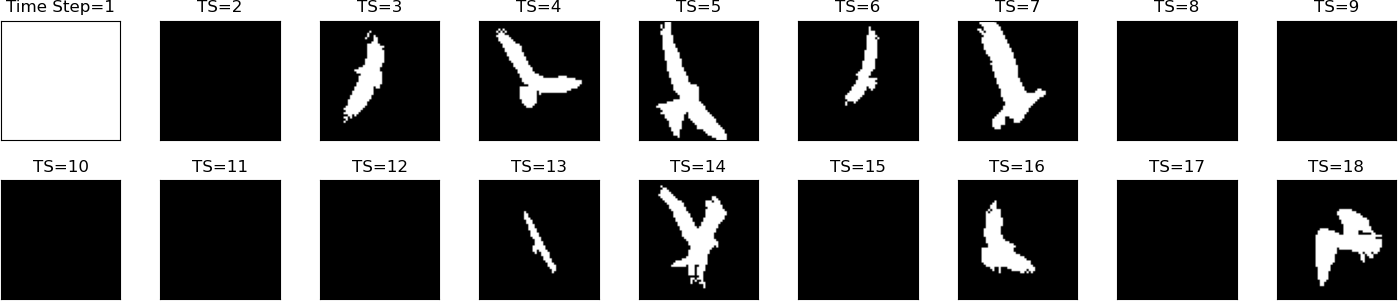}
\caption{ \textbf{Ground truth segmentation labels for the above task.} The first time step we present a offset null label as mentioned in the Section~\ref{sec:semseg}. In the subsequent time steps, we provide the model with ground truth label for the previous time step's input image. The ground truth corresponding to the parchment image is the all zero tensor since it has to be masked out.}
\label{fig:semsegb}
\end{figure}

\subsection{Details of ANML Experiments}
We use publicly released code\footnote{\url{https://github.com/uvm-neurobotics-lab/ANML}} with the suggested the hyperparameters: meta-learning rate $10^{-3}$, Adam outer loop optimizer, and gradient descent as the inner loop optimizer. We use a 4-layer CNN with a 256 filters and a 3-layer hypernetwork  yielding a model with parameter count 3 times that of our CL+LSTM model. Similar to OML, ANML suffered in generalizing to the last subtask due to momentum issues and using SGD as in the case of OML did not help alleviate the issue. Hence, we used Adam for adaptation as in the original paper.

\section{Segmentation Experiments}
\label{app:seg}
\subsection{Dataset}
\label{app:fssdata}
In our segmentation experiments we use the FSS1000 dataset \cite{li2020fss}, which contains 1000 semantic labels and each label containing 10 images. The dataset is split into 700 training, 60 validation, and 240 test classes. We resize these images into 56 $\times$ 56 spatial resolution. 
\subsection{Task Details}
A sample online few-shot segmentation task is presented in Figure~\ref{fig:semsega} and its corresponding ground truth is presented in Figure~\ref{fig:semsegb}. In this task adaptation on the input images is essential as the  model has to (on the fly) learn the distractor concept and mask it out, while segmenting the other images. 

\subsection{Training Details}
\label{app:segtrain}
Both the CL U-Net models are trained with a learning rate of $10^{-3}$ using Adam optimizer and a batch size of 4 throughout the entire curriculum mentioned in Section~\ref{sec:semseg}. Further following \cite{li2020fss} we train our models with mean squared error loss. 

The fine tuning baseline is MAML pre-trained on online few-shot segmentation with no distractors similar to \cite{banerjee2020meta}. We train the model with an outer loop learning rate of $10^{-3}$ using Adam and an inner loop learning rate of $10^{-2}$ using gradient descent. The results in Table \ref{tab:seg} for the fine tuning baseline is obtained by taking such a MAML pre-trained model and fine tuning it on online few-shot segmentation tasks with distractors. When we tried meta-training this model on few-shot segmentation with distractors, the model converged to a trivial solution of masking out both the distractor images and the images that were to be segmented. Hence we only present the fine tuning results.
\subsection{Model Details}
\label{app:segmod}

\begin{figure*}[h]
\centering
\includegraphics[scale=0.4]{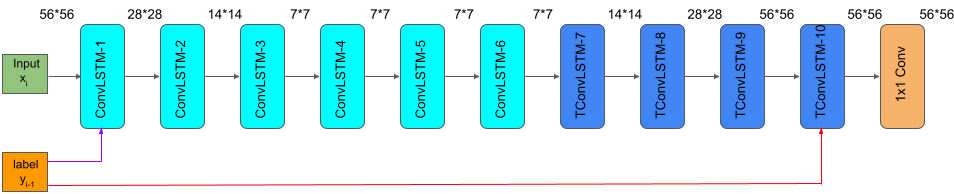}
 \caption{\textbf{CL U-Net: Augmenting memory cells to layers of a U-Net variant.} The cyan layers are ConvLSTM layers, and the reduction in spatial resolution is brought about via pooling layers (not in figure). The dark blue layers (TConvLSTM) are transpose convolutional LSTM layers and are used to upsample along the spatial dimension. Finally, a 1$\times$1 conv layer is used to generate logits at a pixel level. The spatial resolution of the activation at each layer is denoted at the top of each layer. We have two modes of label injection: 1) the ground truth label $y_{i-1}$ corresponding to the image at the previous time step is fed to the first layer (purple line); 2) the ground truth label of previous time step image $y_{i-1}$ is presented to the $10^{th}$ layer (red line). We observe that the first mode (purple) is more effective than the second mode (red). Skip connections are not shown for sake of clarity: skip connections exist between layers 2 and 4; layers 4 and 6; layers 6 and 8.  }
 \label{fig:clunet}
\end{figure*}
In Figure~\ref{fig:clunet}, we use a variant of U-Net wherein each layer is augmented with memory cells. Unlike U-Nets, we do not have skip connections from the first layer to the last layer; we employ skip connections similar to ResNets--once every two layers between the $2^{nd}$ layer and the $8^{th}$. In U-Nets, it is conceivable that label information propagation is easier even to the last layer of the network, since  there is a direct connection between the first and the last layer. However, the effectiveness of label injection in a CL U-Net (as evidenced by Table \ref{tab:seg}) demonstrates that effective label information propagation can take place in standard networks like ResNets as is, without having to employ any additional pathways or tricks. This underlines that label injection can be used for a vast variety of standard networks.

We replace the ConvLSTM cells with standard CNN layers and use ReLUs as nonlinearities to get the architecture for the CNN based finetuning baseline.

\chapter{HyperFields: Towards Zero-Shot Generation of NeRFs from Text}
\section{Model Details}
\label{sec:hyperfield_model}
\textbf{Baselines:} Our baseline is  6 layer MLP with skip connections every two layers. The hidden dimension is 64. We use an open-source \href{https://github.com/ashawkey/stable-dreamfusion}{re-implementation} \cite{stable-dreamfusion} of DreamFusion as both our baseline model and architecture predicted by \ourmethod{}, because the original DreamFusion works relies on Google's Imagen model which is not open-source. Unlike the original DreamFusion, the re-implementation uses Stable Diffusion (instead of Imagen). We use Adam with a learning rate of 1e-4, with an epoch defined by 100 gradient descent steps.

\textbf{HyperFields:} The architecture is as described in Figure~\ref{fig:high_level}. The dynamic hypernetwork generates weights for a 6 layer MLP of hidden dimension 64. The transformer portion of the hypernetwork has 6 self-attention blocks, each with 12 heads with a head dimension of 16. We condition our model with BERT tokens, though we experiment with T5 and CLIP embeddings as well with similar but marginally worse success. Similar to the baseline we use Stable Diffusion for guidance, and optimize our model using Adam  with the a learning rate of 1e-4. 

We use the multiresolution hash grid developed in InstantNGP \cite{instantngp} for its fast inference with low memory overhead, and sinusoidal encodings $\gamma$ to combat the known spectral bias of neural networks \citep{spectralbias}. The NeRF MLP has 6 layers (with weights predicted by the dynamic hypernetwork), with skip connections every two layers. The dynamic hypernetwork MLP modules are two-layer MLPs with ReLU non-linearities and the Transformer module has 6 self-attention layers. Furthermore, we perform adaptive instance normalization before passing the activations into the MLP modules of the dynamic hypernetwork and also put a stop gradient operator on the activations being passed into the MLP modules (as in Figure~\ref{fig:high_level}).

\begin{figure*}[H]
    \centering
    
    \includegraphics[width=0.8\textwidth]{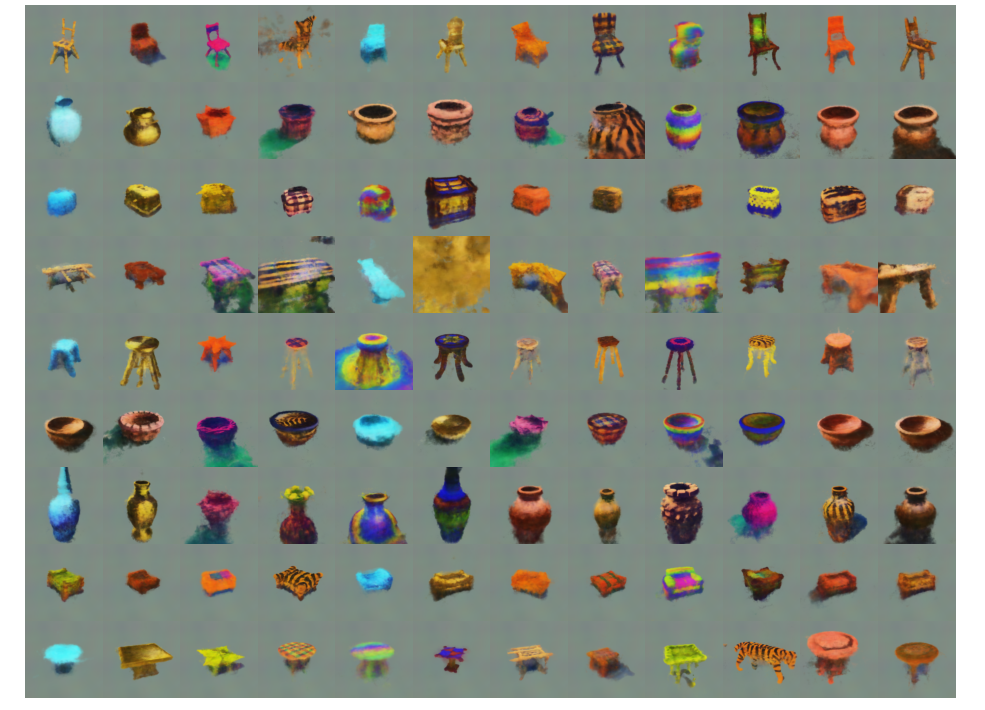}
    \caption{\textbf{ Prompt Packing.} Our dynamic hypernetwork is able to pack 9 different objects across 12 different prompts for a total of 108 scenes. Dynamic hypetnetwork coupled with NeRF distillation enables packing these scenes into one network.}
    \label{fig:complexpack}
\end{figure*}

\section{In-Distribution Generalization with Complex Prompts}

For additional attributes (``plaid", ``Terracotta" etc.), our model produces reasonable zero-shot predictions, and after fewer than 1000 steps of finetuning with SDS is able to produce unseen scenes of high quality. We show these results in Figure~\ref{fig:stylematrix} with 8 objects and 4 styles, where 8 shape-style combinational scenes are masked out during training (opaque scenes in Figure~\ref{fig:stylematrix}).

\begin{figure*}
\centering
\includegraphics[width=0.8\textwidth]{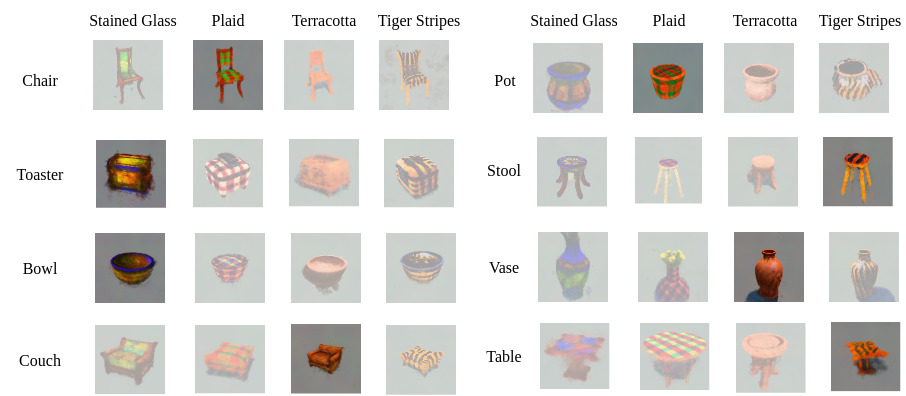}
\caption{\textbf{ Fine-Tuning In-Distribution: seen shape, seen attribute, unseen combination.} During training, the model observes every shape and color, but some combinations of shape and attribute remain unseen.
During inference, the model generalizes by generating scenes that match prompts with previously unseen combinations of shape and attribute, with small amount of finetuning (atmost 1k steps).}
\label{fig:stylematrix}
\end{figure*}

\begin{figure}
\centering
\includegraphics[width=0.8\columnwidth]{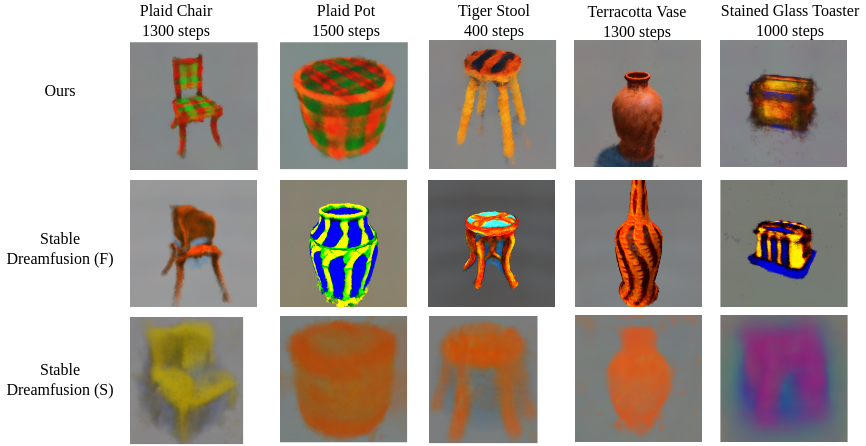}
\caption{\textbf{Generalization Comparison.} We train a single \ourmethod{} model and compare Stable DreamFusion. ``Stable DreamFusion (F)" indicates finetuning from an initialized DreamFusion model. ``Stable DreamFusion (S)" indicates the DreamFusion model trained from scratch. Zero-shot results and initializations are shown in the upper left of ``Ours" and ``Stable DreamFusion (F)", respectively. Above each column indicates the number of training epochs for each method add figures in the upper left. }
\label{fig:stylebaseline}
\end{figure}
\section{Out-of-Distribution Convergence}
\label{sec:ood_supp}
In Figure~\ref{fig:ood} we show the inference time prompts and the corresponding results. Here we provide the list of prompts used to train the model: ``Stained glass chair",  ``Terracotta chair", ``Tiger stripes chair", ``Plaid toaster", ``Terracotta toaster", ``Tiger stripes toaster", ``Plaid bowl", ``Terracotta bowl", ``Tiger stripes bowl", ``Stained glass couch", ``Plaid couch", ``Tiger stripes couch", ``Stained glass pot", ``Terracotta pot", ``Tiger stripes pot", ``Stained glass vase", ``Plaid vase", ``Tiger stripes vase", ``Stained glass table", ``Plaid table", ``Terracotta table".

Since the training prompts dont contain shapes such as ``Blender", ``Knife", ``Skateboard",  ``Shoes", ``Strawberry", ``Lamppost", ``Teapot" and attributes such as ``Gold", ``Glacier", ``Yarn", ``Victorian", we term the prompts used in Figure~\ref{fig:ood} as out-of-distribution prompts--as the model does not see these shapes and attributes during training.

\section{User Study Renders}

We link to the images (including baselines) used in the user study described in Section~\ref{sec:study} \href{https://drive.google.com/file/d/1CTbGXMLVnqTsupslpD6XK0fuOlmwynbv/view?usp=sharing}{here}. All renders are taken from the same camera angle and the baseline scenes are finetuned with the same number of iterations as our model.

\section{Multi-View Consistency of Generated Scenes }
In Figure~\ref{fig:multiview}, we show multiple scenes generated  by a single HyperFields model from various camera poses. Across multiple views we see that geometry is well formed and consistent with the geometry in other views. 
 \begin{figure*}
     \centering
     \includegraphics[width=1\textwidth]{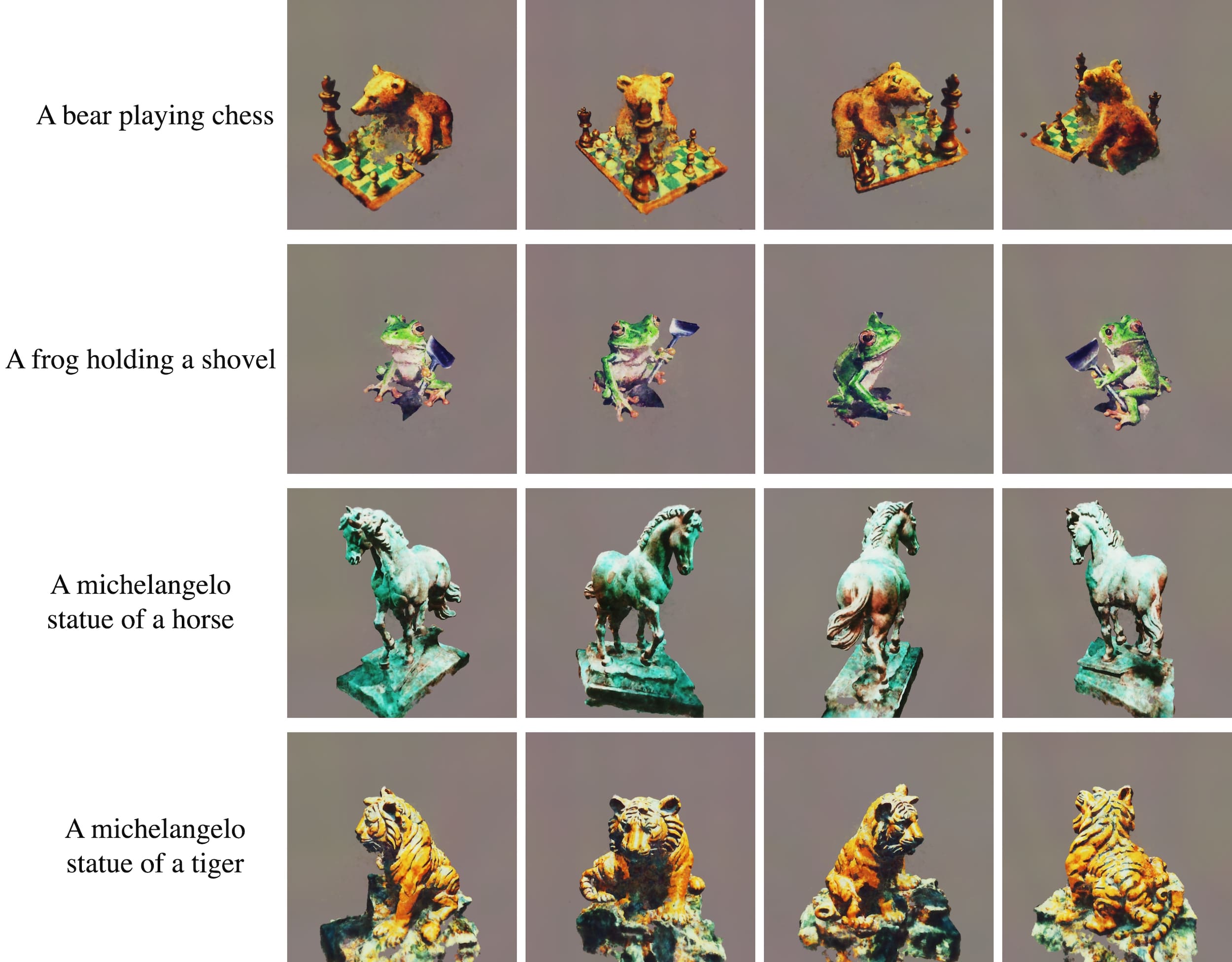}
     \caption{Renders of various scenes generated by HyperFields from various camera poses. The geometry is well formed and consistent across multiple views.}
     \label{fig:multiview}
 \end{figure*}

\end{document}